\documentclass[twoside,11pt]{article}
\usepackage[preprint]{jmlr2e}
\usepackage[english]{babel} 		

\firstpageno{1}                     \usepackage{xspace}
\usepackage{textcomp}				\usepackage{enumitem}
\setitemize{itemsep=0pt}
\usepackage[toc,page,titletoc]{appendix}		\usepackage[nottoc]{tocbibind}					\settocbibname{References}						

\usepackage{caption}
\usepackage{subcaption}
\usepackage{tikz}
\usetikzlibrary{shapes}

\usepackage{amsfonts}				\usepackage{amsmath}				\usepackage{siunitx}  				\usepackage{xfrac}					\usepackage{bm}						\usepackage{mathabx}				

\usepackage{booktabs}               \usepackage{multirow}               \usepackage{soul}                   \PassOptionsToPackage{table}{xcolor} \usepackage{changepage,threeparttable} \usepackage{tabularx}               \usepackage{multirow}               \usepackage{etoolbox}               
\newrobustcmd{\B}{\bfseries}

\usepackage[nameinlink]{cleveref}

\AtBeginEnvironment{appendices}{
	\crefalias{section}{appendix}
	\crefalias{subsection}{appendix}
	\crefalias{subsubsection}{appendix}
}

\hypersetup{ 
	pdfborder={0 0 0}, breaklinks=true,
	colorlinks=true,
	citecolor=green,
	linkcolor=blue,
	urlcolor=magenta,
}

\DeclareRobustCommand{\colorline}[1]{\begin{tikzpicture}
		\raisebox{1.5pt}{
			\draw[#1,solid,line width=1.5pt] (0,0) -- (1em,0);
		}
	\end{tikzpicture}}

\DeclareRobustCommand{\dashedcolorline}[1]{\begin{tikzpicture}
		\raisebox{1.5pt}{
			\draw[#1,dashed,line width=1.5pt] (0,0) -- (1em,0);
		}
	\end{tikzpicture}}

\DeclareRobustCommand{\colorsquare}[1]{\begin{tikzpicture}[baseline=(a.south)]
		\node[rectangle, scale=0.9,color=white, fill=#1] (a) {};
	\end{tikzpicture}}

\DeclareRobustCommand{\colorstar}[1]{\begin{tikzpicture}[baseline=(a.south)]
		\node[star, star points=5, star point ratio=2.25, scale=0.4,color=white, fill=#1] (a) {};
	\end{tikzpicture}}

\newcommand{\colorcross}[1]{\textcolor{#1}{\ding{54}}}
\newcommand{\thincolorcross}[1]{\textcolor{#1}{\ding{53}}}

\usepackage{pifont}

\newcommand{\inlinecode}[1]{\begin{tikzpicture}[baseline=0ex]\node[anchor=base,text height=1em,text depth=1ex,inner ysep=0pt,draw=lightgray!50,fill=lightgray!50,rounded corners=2pt] at (0,0) {\footnotesize\texttt{#1}};\end{tikzpicture}}

\newcommand{\mlcommons}{\textsc{MLCommons}\textsuperscript{\textregistered}\xspace}
\newcommand{\algoperf}{\mbox{\textsc{AlgoPerf}}\xspace}
\newcommand{\benchmarkname}{\textsc{\mbox{AlgoPerf:} Training Algorithms}\xspace}
\newcommand{\benchmarkinghardware}{8$\times$NVIDIA V100 GPUs\xspace}              \newcommand{\nworkloads}{$8$\xspace}                                \newcommand{\resnettesttarget}{$34.6\,\%$\xspace}                                     \newcommand{\lib}{$\downarrow$}                                     \newcommand{\gib}{$\uparrow$}

\newcommand*{\eg}{e.g.\@\xspace}
\newcommand*{\ie}{i.e.\@\xspace}

\usepackage[super]{nth}

\newcommand{\ml}{machine learning\xspace}

\newcommand{\dl}{deep learning\xspace}

\newcommand{\nn}{neural network\xspace}
\newcommand{\nns}{neural networks\xspace}

\newcommand{\dataset}{dataset\xspace}
\newcommand{\datasets}{datasets\xspace}
\newcommand{\runtime}{runtime\xspace}
\newcommand{\runtimes}{runtimes\xspace}
\newcommand{\Runtime}{Runtime\xspace}
\newcommand{\heldout}{held-out\xspace}

\newcommand{\ta}{training algorithm\xspace}
\newcommand{\tas}{training algorithms\xspace}

\newcommand{\wl}{workload\xspace}
\newcommand{\wls}{workloads\xspace}

\newcommand{\Wls}{Workloads\xspace}
\newcommand{\hp}{hyperparameter\xspace}
\newcommand{\Hp}{Hyperparameter\xspace}
\newcommand{\hps}{hyperparameters\xspace}

\newcommand{\ruleset}{ruleset\xspace}
\newcommand{\rulesets}{rulesets\xspace}
\newcommand{\quasirandom}{quasirandom\xspace}

\newcommand{\adam}{\textsc{\mbox{Adam}}\xspace}
\newcommand{\adamw}{\textsc{\mbox{AdamW}}\xspace}
\newcommand{\sgd}{\textsc{\mbox{SGD}}\xspace}
\newcommand{\heavyball}{\textsc{\mbox{Heavy Ball}}\xspace}
\newcommand{\nesterov}{\textsc{\mbox{Nesterov}}\xspace}
\newcommand{\momentum}{\textsc{\mbox{Momentum}}\xspace}
\newcommand{\nadam}{\textsc{\mbox{Nadam}}\xspace}
\newcommand{\nadamw}{\textsc{\mbox{NadamW}}\xspace}
\newcommand{\shampoo}{\textsc{\mbox{Shampoo}}\xspace}
\newcommand{\distshampoo}{\textsc{Distributed \mbox{Shampoo}}\xspace}
\newcommand{\kfac}{\textsc{\mbox{K-FAC}}\xspace}
\newcommand{\adagrad}{\textsc{\mbox{AdaGrad}}\xspace}
\newcommand{\radam}{\textsc{\mbox{RAdam}}\xspace}
\newcommand{\lars}{\textsc{\mbox{LARS}}\xspace}
\newcommand{\lamb}{\textsc{\mbox{LAMB}}\xspace}
\newcommand{\rmsprop}{\textsc{\mbox{RMSProp}}\xspace}
\newcommand{\adafactor}{\textsc{\mbox{Adafactor}}\xspace}
\newcommand{\sam}{\textsc{\mbox{SAM}}\xspace}
\newcommand{\samadam}{\textsc{\mbox{SAM}(w.~\adam)}\xspace}
\newcommand{\betaone}{$\beta_1$}
\newcommand{\betatwo}{$\beta_2$}
\newcommand{\optlisttext}{\textsc{\mbox{OptList}}\xspace}

\newcommand{\cosinedecay}{cosine decay\xspace}
\newcommand{\cosinelrdecay}{cosine learning rate decay\xspace}
\newcommand{\warmup}{warmup\xspace}
\newcommand{\Warmup}{Warmup\xspace}
\newcommand{\lineardecay}{linear decay\xspace}

\newcommand{\constantschedule}{constant\xspace}

\newcommand{\wcd}{\warmup$\!+\!$ \cosinedecay}
\newcommand{\wldc}{\warmup$\!+\!$ \lineardecay$\!+\!$ \constantschedule}

\newcommand{\imagenetresnet}{\textsc{ImageNet ResNet-50}\xspace}

\newcommand{\wmttransformer}{\textsc{WMT Transformer}\xspace}

\newcommand{\criteodlrm}{\textsc{Criteo 1TB DLRM small}\xspace}
\newcommand{\ogbggnn}{\textsc{OGBG GNN}\xspace}

\newcommand{\imagenet}{\textsc{ImageNet}\xspace}
\newcommand{\wmt}{\textsc{WMT}\xspace}
\newcommand{\librispeech}{\textsc{LibriSpeech}\xspace}
\newcommand{\criteo}{\textsc{Criteo 1TB}\xspace}
\newcommand{\ogbg}{\textsc{OGBG}\xspace}
\newcommand{\fastmri}{\textsc{fastMRI}\xspace}

\newcommand{\cifar}{\textsc{CIFAR}\xspace}
\newcommand{\cifarten}{\textsc{CIFAR-10}\xspace}

\newcommand{\svhn}{\textsc{SVHN}\xspace}
\newcommand{\mnist}{\textsc{MNIST}\xspace}

\newcommand{\resnetfifty}{\textsc{ResNet-50}\xspace}
\newcommand{\resnet}{\textsc{ResNet}\xspace}
\newcommand{\resnets}{\textsc{ResNet}s\xspace}
\newcommand{\wideresnet}{\textsc{Wide ResNet}\xspace}
\newcommand{\vit}{\textsc{ViT}\xspace}
\newcommand{\vitfull}{\textsc{Vision Transformer}\xspace}
\newcommand{\transformer}{\textsc{Transformer}\xspace}
\newcommand{\deepspeech}{\textsc{DeepSpeech}\xspace}
\newcommand{\conformer}{\textsc{Conformer}\xspace}
\newcommand{\dlrmsmall}{\textsc{DLRMsmall}\xspace}
\newcommand{\gnn}{\textsc{GNN}\xspace}
\newcommand{\unet}{\textsc{U-Net}\xspace}

\newcommand{\resnetvtwo}{\textsc{ResNetV2}\xspace}
\newcommand{\resnettwohun}{\textsc{ResNet-200}\xspace}
\newcommand{\dlrm}{\textsc{DLRM}\xspace}

\newcommand{\preln}{\textsc{Pre-LN}\xspace}
\newcommand{\prelnfull}{\textsc{Pre-Layer Norm}\xspace}
\newcommand{\postln}{\textsc{Post-LN}\xspace}
\newcommand{\postlnfull}{\textsc{Post-Layer Norm}\xspace}

\newcommand{\batchnorm}{batch normalization\xspace}
\newcommand{\layernorm}{layer normalization\xspace}
\newcommand{\Layernorm}{Layer normalization\xspace}
\newcommand{\instancenorm}{instance normalization\xspace}
\newcommand{\relu}{ReLU\xspace}
\newcommand{\silu}{SiLU\xspace}
\newcommand{\gelu}{GELU\xspace}
\newcommand{\Tanh}{TanH\xspace}
\newcommand{\leakyrelu}{Leaky ReLU\xspace}
\newcommand{\dropout}{dropout\xspace}

\newcommand{\bleu}{BLEU\xspace}
\newcommand{\sacrebleu}{sacreBLEU\xspace}

\newcommand{\rulesurl}{\href{https://github.com/mlcommons/algorithmic-efficiency/blob/main/RULES.md}{github.com/mlcommons/algorithmic-efficiency/blob/main/RULES.md}\xspace}
\newcommand{\mlccodebase}{https://github.com/mlcommons/algorithmic-efficiency}
\newcommand{\initcodebase}{https://github.com/google/init2winit}

\newcommand{\deepobs}{\textsc{DeepOBS}\xspace}
\newcommand{\backpack}{\textsc{BackPACK}\xspace}
\newcommand{\jraph}{\textsc{Jraph}\xspace}

\newcommand{\pytorch}{\textsc{PyTorch}\xspace}
\newcommand{\jax}{\textsc{JAX}\xspace}
\newcommand{\flax}{\textsc{Flax}\xspace}
\newcommand{\mlperf}{\textsc{MLPerf}\texttrademark\xspace}
\newcommand{\mlperftraining}{\textsc{MLPerf\texttrademark\xspace Training}\xspace}
\newcommand{\dawnbench}{\textsc{DAWNBench}\xspace}
\newcommand{\dlbs}{\textsc{Deep Learning Benchmark Suite}\xspace}
\newcommand{\deepbench}{\textsc{DeepBench}\xspace}
\newcommand{\tbdnn}{\textsc{Training Benchmark for DNNs}\xspace}
\newcommand{\benchopt}{\textsc{Benchopt}\xspace}

\newcommand{\specaug}{\textsc{SpecAugment}\xspace}

\newcommand{\val}{\mathrm{val}}

\newcommand{\normltwo}{L^2}

\DeclareMathAlphabet{\mathsfit}{\encodingdefault}{\sfdefault}{m}{sl}
\SetMathAlphabet{\mathsfit}{bold}{\encodingdefault}{\sfdefault}{bx}{n}

\definecolor{SNSblue}{rgb}{0.1216, 0.4666, 0.7059}
\definecolor{SNSorange}{rgb}{1.0, 0.4980, 0.0549}
\definecolor{SNSgreen}{rgb}{0.1725, 0.6274, 0.1725}
\definecolor{SNSred}{rgb}{0.84, 0.15, 0.16}
\definecolor{SNSpurple}{rgb}{0.58, 0.40, 0.74}

\definecolor{SNSorange_shaded}{HTML}{ffcea3}
\definecolor{SNSblue_shaded}{HTML}{8ebad9}
\definecolor{SNSgreen_shaded}{HTML}{cae7ca}
\definecolor{SNSred_shaded}{HTML}{ea9293}
  
\begin{document}

\title{Benchmarking Neural Network Training Algorithms}

\author{\name George E.~Dahl$^1$\thanks{Corresponding authors.} \email gdahl@google.com \\
    \name Frank Schneider$^2$\footnotemark[\value{footnote}] \email f.schneider@uni-tuebingen.de \\
    \name Zachary Nado$^1$\footnotemark[\value{footnote}] \email znado@google.com \\
    \name Naman Agarwal$^1$\footnotemark[\value{footnote}] \email namanagarwal@google.com \\
    \name Chandramouli Shama Sastry$^{3,4}$\thanks{These authors are listed in random order.} \email chandramouli.sastry@gmail.com \\
    \name Philipp Hennig$^2$\footnotemark[\value{footnote}] \email philipp.hennig@uni-tuebingen.de \\
    \name Sourabh Medapati$^1$\footnotemark[\value{footnote}] \email smedapati@google.com \\
    \name Runa Eschenhagen$^2$\footnotemark[\value{footnote}] \email re393@cam.ac.uk \\
    \name Priya Kasimbeg$^1$\footnotemark[\value{footnote}] \email kasimbeg@google.com \\
    \name Daniel Suo$^1$\footnotemark[\value{footnote}] \email dsuo@google.com \\
    \name Juhan Bae$^{3,5}$\footnotemark[\value{footnote}] \email jbae@cs.toronto.edu \\
    \name Justin Gilmer$^1$\footnotemark[\value{footnote}] \email gilmer@google.com \\
    \name Abel L.~Peirson$^6$\footnotemark[\value{footnote}] \email alpv95@stanford.edu \\
    \name Bilal Khan$^1$\footnotemark[\value{footnote}] \email kbilal@google.com \\
    \name Rohan Anil$^1$\footnotemark[\value{footnote}] \email rohananil@google.com \\
    \name Mike Rabbat$^7$\footnotemark[\value{footnote}] \email mikerabbat@meta.com \\
    \name Shankar Krishnan$^1$\footnotemark[\value{footnote}] \email skrishnan@google.com \\
    \name Daniel Snider$^{3,5}$\thanks{These authors are listed in random order.} \email dans@cs.toronto.edu \\
    \name Ehsan Amid$^1$\footnotemark[\value{footnote}] \email eamid@google.com \\
    \name Kongtao Chen$^1$\footnotemark[\value{footnote}] \email kongtao@google.com \\
    \name Chris J.~Maddison$^{3,5}$\footnotemark[\value{footnote}] \email cmaddis@cs.toronto.edu \\
    \name Rakshith Vasudev$^8$\footnotemark[\value{footnote}] \email rakshith.vasudev@dell.com \\
    \name Michal Badura$^1$\footnotemark[\value{footnote}] \email mbadura@google.com \\
    \name Ankush Garg$^1$\footnotemark[\value{footnote}] \email ankugarg@google.com \\
    \name Peter Mattson$^1$\footnotemark[\value{footnote}] \email petermattson@google.com\AND
    \addr $^1$Google;
    \addr $^2$University of Tübingen;
    \addr $^3$Vector Institute;
    \addr $^4$Dalhousie University;
    \addr $^5$University of Toronto;
    \addr $^6$Stanford University;
    \addr $^7$Meta AI (FAIR);
    \addr $^8$Dell Technologies
}

\editor{tbd.}
 \maketitle

\newpage  \begin{abstract}Training algorithms, broadly construed, are an essential part of every \dl pipeline. Training algorithm improvements that speed up training across a wide variety of \wls (e.g., better update rules, tuning protocols, learning rate schedules, or data selection schemes) could save time, save computational resources, and lead to better, more accurate, models. Unfortunately, as a community, we are currently unable to reliably identify \ta improvements, or even determine the state-of-the-art \ta. In this work, using concrete experiments, we argue that real progress in speeding up training requires new benchmarks that resolve three basic challenges faced by empirical comparisons of \tas: (1) how to decide when training is complete and precisely measure training time, (2) how to handle the sensitivity of measurements to exact \wl details, and (3) how to fairly compare algorithms that require \hp tuning. In order to address these challenges, we introduce a new, competitive, time-to-result benchmark using multiple \wls running on fixed hardware, the \benchmarkname benchmark. Our benchmark includes a set of \wl variants that make it possible to detect benchmark submissions that are more robust to \wl changes than current widely-used methods. Finally, we evaluate baseline submissions constructed using various optimizers that represent current practice, as well as other optimizers that have recently received attention in the literature. These baseline results collectively demonstrate the feasibility of our benchmark, show that non-trivial gaps between methods exist, and set a provisional state-of-the-art for future benchmark submissions to try and surpass.
\end{abstract}
 

\jmlrheading{1}{2000}{1-48}{4/00}{10/00}{meila00a}{\algoperf Authors}
\ShortHeadings{Benchmarking Neural Network Training Algorithms}{\algoperf Authors}

\tableofcontents
\newpage  

\newpage

\section{Introduction}
\label{sec:intro}

Although artificial \nns are extremely useful models, training them remains quite expensive. Moreover, investing more time and computational resources in training produces better, more accurate models. For example, training larger models, training longer, training on more data, or performing more exploratory experiments can all improve results. 
Training more efficiently would directly reduce costs and/or indirectly produce more accurate models.
Although there are many ways to make training more efficient, in this work, we restrict our attention to improvements in training \emph{algorithms}.

Unfortunately, as a community, we are currently unable to identify which existing \tas are best, let alone understand what novel methods are the most promising. A multitude of \tas have been proposed, and more are proposed every year. \citet{Schmidt2021} lists well over a hundred methods, mostly published in the last seven years. 
Naturally, each of these papers claims that their algorithm has significant benefits, but the vast majority of the \dl community never uses any of these techniques. At present, most \tas are assumed to not be useful until they have been widely adopted, creating a chicken-and-egg problem. This state of affairs should be troubling for anyone trying to train \nns or develop new \tas.

To provide actionable guidance to practitioners, we need to understand how existing and future methods perform in practice.
Additionally, the community needs to incentivize research that makes actual progress on \tas. 
Currently, the paper that introduces a new technique is also the primary source of experimental evidence for its merits, which causes a conflict of interest if the researchers proposing the techniques also have control over the baselines included for comparison.

Although comparative studies and meta-analyses are effective ways to assess the merits of competing techniques in principle, so far, they have not been able to provide definitive answers.
One reason may be that such studies are themselves often subject to criticism and accusations of bias: For any concrete empirical comparison, the proponents of any particular technique can always find fault with the details of the setup. 
Without broad agreement on delicate issues such as \hp tuning, in many cases, such criticism may indeed be justified. 
Another issue is that the studies' retrospective nature shifts too much of the burden of proof away from the original inventors of \tas. Although a useful tool, empirical studies ultimately face an uphill battle to generate strong enough evidence to actually convince practitioners, especially since the truth can be quite complicated. 

Instead of relying entirely on retrospective comparative studies, why not have researchers submit their methods to a common competitive evaluation?
Such competitive benchmarks have historically worked well.
For example, the \dl community tried for years to convince the computer vision community that \nns were better models for image classification by publishing papers, but \dl was only widely adopted for computer vision tasks after \nns showed success in high-profile competitive benchmarks \citep[\eg,][]{Russakovsky2015}.
Similarly, if we can construct benchmarks that capture useful notions of training speed in sufficiently realistic conditions, we could determine whether widely-used \tas are indeed the best currently available. If not, we could demonstrate that current incumbent methods should be replaced with better alternatives.

The mission of the \mlcommons\footnote{\mlcommons and \mlperf are registered and unregistered trademarks, respectively, of the MLCommons Association in the United States and other countries. All rights reserved. Unauthorized use strictly prohibited.} Algorithms Working Group is to create a set of rigorous and relevant benchmarks to measure \nn training speedups due to algorithmic improvements. The best benchmarks will capture what is needed to drive progress at a particular time in the field and continue evolving along with the needs of the community. This paper describes the working group's first attempt to benchmark \tas for \nns, and we intend to continue releasing improved versions periodically. We hope that anyone interested in improving upon the benchmark we propose here will consider joining the working group to help shape future versions.

\subsection{Contributions of This Work}

\begin{enumerate}
	\item We precisely articulate the challenges of benchmarking training algorithms, provide concrete experiments demonstrating the methodological issues we identify, and explain how they hold back training algorithms research (\Cref{sec:challenges}).

    \item We introduce the \benchmarkname benchmark --- a competitive, time-to-result benchmark on multiple workloads running on fixed hardware for systematically comparing training algorithms (\Cref{sec:rules}).
    \begin{enumerate}
        \item Our benchmark defines a complete and workable procedure for setting (validation and test error) targets and measuring training time to reach them. Furthermore, this procedure produces targets that are reasonably competitive with results in the literature, given the resource constraints of the benchmark (\Cref{sec:exp_target_setting}).
        
        \item Our benchmark incentivizes generally useful training algorithms by computing a joint score across all workloads and by including randomized workloads to simulate novel problems.

        \item Our benchmark requires submissions to explicitly account for the hyperparameter tuning they need to achieve their results across workloads, giving submissions that are easier to tune an advantage. Submissions can either compete using limited, parallel tuning resources, or enter as a completely self-tuning and hyperparameter-free algorithm.
        
        \item We specify a set of benchmark workloads covering image classification, speech recognition, machine translation, MRI reconstruction, click-through rate prediction, and chemical property prediction tasks, and we design workload variants to challenge \tas to be more robust to natural workload modifications.
        
        \item We provide open-source \jax and \pytorch implementations of all workloads, and a \ta API that supports submissions in both frameworks. By providing code along with details of the benchmark system, we make it easy to independently reproduce results on the benchmark.
    \end{enumerate}
    
    \item We construct baselines by defining search spaces for eight popular optimizers (\adamw, \nadamw, \heavyball, \nesterov, \lamb, \adafactor, \samadam, \distshampoo) that includes both popular optimizers that represent current practice and methods that have received attention in the recent literature (\Cref{sec:exp_baselines}).
    \begin{enumerate}
        \item Collectively, these baselines demonstrate that our benchmark is feasible and that there is a non-trivial gap between different methods.
        
        \item We show that our benchmark score successfully favors algorithms that are easier to tune and that successful \tas require good search spaces that are tailored to the tuning budget. Specifically, baselines using adaptive methods (\adamw and \nadamw) score more highly than baselines using non-adaptive methods (\heavyball and \nesterov), in large part because of the difficulty of constructing good search spaces for the latter.
        
        \item We set a provisional state of the art on the benchmark under our external tuning \ruleset based on extensive experiments including tens of thousands of tuning trials.
        \nadamw with our search space performs well on every (fixed) \wl.
    \end{enumerate}
    
    \item We construct a set of 24 workload variants (three for each fixed \wl) that make it possible to detect improvements in robustness over current popular methods, and demonstrate the need for additional research on \hp transfer (\Cref{sec:exp_randomized_workloads}).
\end{enumerate}

\paragraph{How to read this paper.} This paper serves a dual purpose: it acts both as a research report on benchmarking \nn \tas, and as a form of technical documentation of the \benchmarkname benchmark and its rules.
Readers who plan to participate in the competition may wish to start by reading the sections that describe the rules (\Cref{sec:rules}), using them as a higher-level companion to the latest version of the complete rules that can be found online, and then move to the experiments performed on this benchmark (\Cref{sec:exp_target_setting,sec:exp_baselines,sec:exp_randomized_workloads}).
Readers who are primarily interested in the research aspects of the paper may wish to skip, or only briefly skim, the rules section (\Cref{sec:rules}) during an initial read, with the exception of \Cref{sec:rls_scoring} which is essential to understanding the scores reported in \Cref{sec:exp_baselines}.

\subsection{Training Algorithm Benchmark Goals and Scope}
\label{sec:intr_scope}

In order to construct a benchmark for general \nn \tas that measures training speed, we need to decide what we mean by a \emph{general \ta}, and how we measure training \emph{speed}.

\paragraph{Design goal.} We aim to encourage \emph{general-purpose} \tas that are easy to apply across different data modalities and model architectures. It is legitimate to wonder what the best \ta is for a particular \wl, \eg ``\resnetfifty on \imagenet," or even a single \wl family, \eg ``convolutional neural net image classifiers." But given the still rapid rate of change among architectures in the field, it is more promising for the community to develop relatively general methods first. A downside of this design choice is that it implicitly assumes a clean separation between model and algorithm.

\paragraph{Separating ``algorithm'' and ``model''.} We adopt the paradigm of \ml as optimization; \ie, we view the optimizer and the model as two separate modules and focus on the first. Nevertheless, we must consider more pieces of the training pipeline than just the optimizer because the separation can be subtle. For example, should \batchnorm \citep{Ioffe2015} be considered an aspect of the \ta or the modeling? An ideal benchmark should allow competitors a maximum of innovation and creativity while still providing enough constraints to make fair comparisons and glean useful insights from the results. \Cref{sec:rls_submission} explains how we delineate what constitutes a \ta and, thus, the design domain of competition entries. In general, we aim to offer flexibility while enforcing generality: We allow competitors to choose optimization algorithms, update rules, and control regularization. We also allow algorithms that reorder or re-sample the training data. But we restrict or prohibit methods that only apply to specific model architectures or data modalities.

\paragraph{Measuring training performance.} Although this work, like the majority of the community, uses the metaphor of optimization, optimization and \emph{learning} are, of course, not technically the same. We ultimately care about how quickly a method reaches a satisfactory out-of-sample error, as estimated by a validation or test set error rate. To incentivize practically useful techniques, success in our competition must be based on this end goal, not on how quickly an algorithm can actually minimize the training objective. Unfortunately, this forces us to confront the complexities of regularization and overfitting in the benchmark (which are also present, of course, in any real application). \Cref{sec:rls_timetoresults} details how we define ``time to result". In short, we define a target out-of-sample error rate for every \wl  and score submissions based on how quickly they reach these targets.

\paragraph{Internal and external \hp tuning.} Comparing generally-applicable \nn \tas requires careful attention to the rules around \hp tuning for benchmark submissions. In particular, should certain types of tuning be viewed as integral parts of the \ta or external to the \ta? At the moment, choosing \hps is still largely the job of the user, so algorithms should perform well under an external tuning schedule. However, methods that do not require external tuning at all are, of course, desirable, and internal parameter tuning is arguably among the biggest opportunities for improvement in resource use. So we also need an opportunity for methods that effectively tune themselves to shine. \Cref{sec:rls_tuning} describes our approach to \hp tuning in detail.

\section{The Challenges of Empirical Comparisons of Training Algorithms}
\label{sec:challenges}

An ideal empirical comparison of \tas should be (as much as possible) convincing, informative, and practically relevant. A convincing comparison generates high quality empirical evidence, makes ``fair'' comparisons to strong baselines, and is not misleading. An informative comparison disentangles the causes of any measured improvements and provides insight into the observed phenomena. A practically relevant comparison measures situations that are likely to arise in important applications and studies conditions as similar as possible to current practice in applied work. Unfortunately, several basic challenges make achieving these desiderata far from simple.

\subsection{Precisely Defining and Measuring Training Speed}
\label{sec:chal_training_speed}

Currently, papers proposing \tas for deep \nns tend to shy away from making quantitative, \emph{empirical} claims, although some will make theoretical convergence rate claims that depend on assumptions that preclude them from being directly applicable to \nn training (\eg, assuming a convex loss function). For example, in abstracts we see phrases such as  ``frequently delivers faster convergence'' \citep{Lucas2019},  ``outperforms other methods with fast convergence and high accuracy'' \citep{Zhuang2020}, or ``works well in practice and compares favorably to other stochastic optimization methods'' \citep{Kingma2015}. In contrast, \dl modeling papers usually make precise quantitative claims. For example, in abstracts we see language such as ``achieves 28.4 BLEU on the WMT 2014 English-to-German translation task, improving over the existing best results'' \citep{Vaswani2017}, ``achieves state-of-the-art 84.3\,\% top-1 accuracy on \imagenet, while being 8.4$\times$ smaller and 6.1$\times$ faster on inference'' \citep{Tan2019}, or ``achieves 86.2\,\% top-1 \imagenet accuracy, while being 4.7$\times$ faster'' \citep{Bello2021}.\footnote{Ironically, \citet{Bello2021} showed that the gains observed in \citet{Tan2019} were largely due to an improved training procedure, even though \citet{Tan2019} was presented as a modeling paper.} Some examples of recent optimizer papers that make quantitative claims are \citet{chen2023symbolic}, \citet{xie2022adan}, and \citet{Tian2022}, but these claims are still using different metrics or are referring to different benchmarks or workload versions.

To the detriment of progress on \tas, a lack of quantitative claims means there is no clear notion of the ``state of the art'', only what is popular or topical. Most \ta  papers will display the training curve (loss vs time) for their method and some baseline, but without a shared understanding of how such curves should be converted into quantitative measurements of training speed, they will only provide an illusory sense of precision. Even valiant attempts to make quantitative sense of such a plot will produce claims tortured with caveats. For example, \citet{Tian2022} claims their method's loss is ``always significantly lower beyond [the first] 30\,\% of the training procedure.''

\begin{figure}[!ht]
    \includegraphics[width=\textwidth]{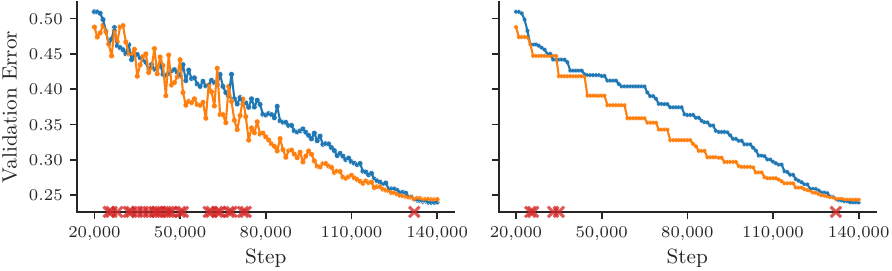}
    \caption{\textbf{A direct comparison of training curves is ill-posed if they intersect.}
    	\textit{Left:} The validation error for two different runs (\colorline{SNSorange}, \colorline{SNSblue}) of \adam on \resnetfifty on \imagenet. \textit{Right:} The best validation error obtained so far by each curve (\ie the running minimum of the left subplot). Even after removing the early part of training and looking at the best result so far (\textit{right}), the curves intersect multiple times (highlighted by red crosses (\colorcross{SNSred}) at the bottom of each plot). Through much of the plot, the orange curve is winning, but the blue curve overtakes it near the end of the runs.}
    \label{fig:crossing_curves}
\end{figure}

The root cause of the preponderance of vague training speed claims in the literature is that a direct comparison of two training curves is ill-posed. Certainly, if one curve is strictly below the other, we can say it is ``better'', but even then, it is not clear by how \emph{much}. Real-world training curves are noisy and tend to cross each other, often multiple times. \Cref{fig:crossing_curves} (\textit{left}) shows two sample validation error curves for \resnetfifty trained on \imagenet, achieving a final validation error of 24.0\,\% (\colorline{SNSblue}) and 24.4\,\% (\colorline{SNSorange}), respectively (see \Cref{app:exp_details_chal_crossing_curves} for exact experiment details). These curves intersect multiple times, even after ignoring the early part of training. Even more importantly, the curves of the best validation error seen so far (\Cref{fig:crossing_curves}, \textit{right}) also intersect multiple times, showing that which run is leading the race swaps back and forth. The run that trains \resnetfifty the fastest depends on what it means for training to be complete.

Consequently, to measure the training time, we need a clear criterion that indicates when training is complete. There can be many possible criteria, and it is not obvious which is the correct one. As a result, many papers avoid the issue altogether, relying instead on more vague descriptions of training results. Straight-forward approaches, such as computing the area under the validation loss curve, fail to capture what is relevant in practical applications. For example, improvements that affect only the initial stage of training are  irrelevant in most cases. Despite these difficulties, a precise definition of training speed is essential. If the community can standardize how training time should be measured, we can shift the literature towards clear quantitative comparisons between \tas.

The simplest approach is to set a target error rate and measure how long it takes to first achieve this target. Specifically, we propose measuring training time based on how long it takes a run to reach a \wl-specific, out-of-sample error target that is selected to be a competitive, near-state-of-the-art error rate. For any given comparison between algorithms, standardizing hardware is also a prerequisite for measuring wall-clock time, although perhaps it would be sufficient to standardize a cost model for primitive operations. Determining an appropriate target error rate for a particular \wl is not trivial. \Cref{sec:rls_timetoresults} describes our approach to setting error rate targets. Targets for a given \wl may need to be revised as new results emerge, but they should be more stable than the state-of-the-art result on the corresponding \dataset since a \wl freezes a specific model as well. Ultimately, targets will always be somewhat arbitrary, but that is all the more reason they should be standardized \emph{before} comparing \tas. Our general philosophy is to try and pick the best error rates we can consistently and reliably achieve with standard methods.

\subsection{Dependence on the Workload}
\label{sec:chal_wl}

The performance of a training algorithm is necessarily a function of the workload it is timed on. Ultimately, in applications, we are interested in finding the method best suited for the particular workload in front of us. In general, this will be a novel workload without many previous results. The more we can generalize from previous results on related workloads, the better we can prioritize what to try. For generic methods research, on the other hand, we are interested in characterizing how methods perform across a diverse set of workloads representative of current important applications. Once again, we cannot measure every possible method even on a single workload, let alone on all possible workloads, so we are forced to extrapolate. When building generic methods, we need to detect when a method provides a generally useful improvement and thus we also need to find a meaningful way to aggregate performance across multiple workloads.

\subsubsection{Sensitivity of Optimizer Ranking to the Model Architecture}
\label{sec:chal_model_changes}

Although it is obvious that the model and \dataset affect how well different algorithms perform, the situation is worse than it might initially appear because even seemingly small details about the workload can have a large effect on our results. As a recent example, \citet{Nado2021} describes several small implementation details that make a large difference in matching state of the art \mlperf\!\resnetfifty \imagenet training speed results. For example, they mention needing to set the initial value of the \batchnorm scale parameters in the final layer of each residual block to a value less than one, match the exact v1.5 version of \resnetfifty, match the virtual batch size of \batchnorm, and (perhaps least surprisingly) avoid applying $\normltwo$ regularization to bias variables.

To further illustrate how brittle training results can be, we ran three different experiments, each making a minor change to a different workload: changing the stride in the final residual block of a \wideresnet on \cifarten, adding extra \batchnorm layers to \resnets, and switching between two different published \transformer models that only differ by the placement of the \layernorm block.

\begin{figure}[!ht]
	\centering
    \includegraphics[width=\textwidth]{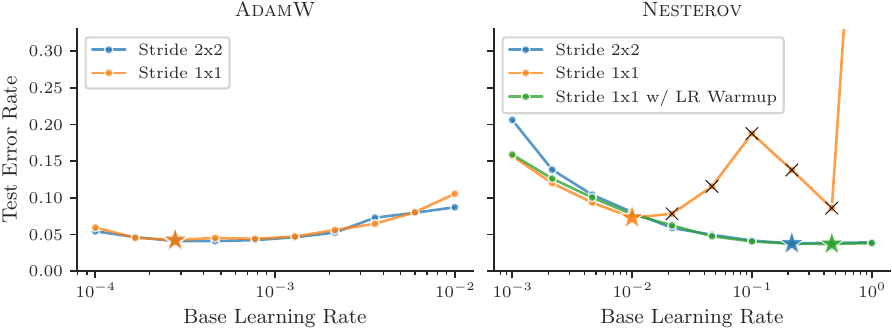} 
    \caption{{\bf Small architectural changes can result in different rankings between \tas.} In the above plot, we compare a learning rate sweep with \adamw (\textit{right}) and \nesterov (\textit{left}) on two variants of the \cifarten \wideresnet. The first variant is the standard architecture, which uses a 2x2 convolutional stride in each residual block, labeled ``Stride 2x2'' (\colorline{SNSblue}). The second variant changes the strides in the final residual block to 1x1, labeled ``Stride 1x1'' (\colorline{SNSorange}). On the standard architecture, \nesterov outperforms \adamw (the best-performing \hp setting per architecture is highlighted with a colored star, \eg (\colorstar{SNSblue})). However, the stride change results in significant training instability at high learning rates with \nesterov (highlighted with a black cross (\thincolorcross{black})). This instability causes \nesterov to under perform \adamw. If a learning rate \warmup is applied (\colorline{SNSgreen}) then we recover the performance of the original model. \adamw on the other hand is unaffected by the training instability caused by this architectural change.}
    \label{fig:wrn_swap}
\end{figure}

\paragraph{Wide ResNet with stride changes}
In the first experiment, we start with a standard \wideresnet \citep{Zagoruyko2016} architecture on \cifarten.
For this workload, we compare the \nesterov optimizer with \adamw on a simple learning rate sweep with \cosinedecay (full experimental details provided in \Cref{app:exp_details_chal_wideresnet}).
After selecting the best learning rate from the sweep, \nesterov achieves a better test error than \adamw, see \Cref{tab:wrn_swap} and \Cref{fig:wrn_swap}.
The superiority of \nesterov in this experiment is largely due to \adamw overfitting in these conditions; \ie, \adamw achieves a better training loss than \nesterov.
However, if we change the convolutional strides in the final residual block from the standard 2$\times$2 stride to a 1$\times$1 stride, the performance of \nesterov shows a large $3.5\%$ increase in test error.
\adamw, on the other hand, is largely unaffected by this architectural change.
\nesterov struggles to train the 1$\times$1 stride model due to an early training instability caused by the large loss curvature at initialization---the largest eigenvalue of the loss Hessian changes from $32$ to $1052$ with this architectural modification and initial weight distribution.
This kind of instability can be dealt with using learning rate \warmup \citep{Gilmer2021}. Adding 1000 steps of linear \warmup makes \nesterov stable at large learning rates despite the high initial loss curvature (see \Cref{fig:wrn_losscurve} in the Appendix).\begin{table}[!ht]
    \centering
    \begin{tabular}{lS[table-format=1.4]S[table-format=1.4]}\toprule
 \textbf{Training Algorithm} & \textbf{Stride} $\pmb{2\times2}$ & \textbf{Stride} $\pmb{1\times1}$ \\
 & \textit{(standard)} & \\
 \midrule
\nesterov & \B 0.0376 & 0.0726 \\
\adamw & 0.0407 & 0.0420 \\
\nesterov + Warmup & 0.0380 & \B 0.0378  \\
\bottomrule
\end{tabular}
     \caption{\textbf{Performance of different \tas on \wideresnet with stride changes.} After changing from the standard $2\times2$ stride to a $1\times1$ stride, the performance of \nesterov drops significantly, while \adamw is largely unaffected by this architectural change. Adding a learning rate \warmup to \nesterov (\ie \nesterov + Warmup) allows it to recapture its original performance.}
    \label{tab:wrn_swap}
\end{table}

\paragraph{Additional \batchnorm layers}
A similar inversion of rank ordering between \nesterov and \adamw occurs when we add in additional \batchnorm \citep{Ioffe2015} layers to the $200$-layer \resnetvtwo architecture \citep{He2016a}.
This \batchnorm layer is added after every residual connection, similar to the structure of the original \postlnfull \transformer \citep{Vaswani2017}.
This architectural change actually benefits \nesterov with the 50 layer model, offering slight validation error and stability improvements.
However, when this \resnet with extra \batchnorm layers is scaled to $200$ layers, it becomes completely untrainable with \nesterov at any learning rate in the sweep (see \Cref{tab:resnet200_swap}).
\adamw is also affected by this architectural change, but is able to train successfully with a $3\%$ drop in final performance.
Similar to the \wideresnet stride change, this instability results from a dramatic increase in the initial loss curvature---with the extra \batchnorm layers the initial loss curvature exceeds $10^{10}$!
Prepending a linear learning rate \warmup in this case is insufficient to resolve the training instability. However, with the addition of gradient clipping, we are able to recover the performance of the original architecture.
\begin{table}[!ht]
    \centering
    \begin{tabular}{lS[table-format=1.4]S[table-format=1.4]}\toprule
\textbf{Training Algorithm} & \textbf{\resnettwohun} & \textbf{\resnettwohun}  \\
 & \textit{(standard)} & \textit{Extra-BN} \\
 \midrule
\nesterov &\B 0.2090 & {No Feasible Trials} \\
\adamw & 0.2626  & 0.2722 \\
\nesterov + Grad Clip & 0.2091 & \B 0.2094 \\
\bottomrule
\end{tabular}
     \caption{\textbf{Performance of different \tas after adding an extra \batchnorm layer to a \resnettwohun.} While \adamw is only slightly affected by adding extra \batchnorm layers, vanilla \nesterov becomes untrainable. Adding gradient clipping (\ie \nesterov + Grad Clip), however, allows it to recover the original architecture's performance.}
    \label{tab:resnet200_swap}
\end{table}

\begin{figure}[!ht]
    \centering
    \includegraphics[width=\textwidth]{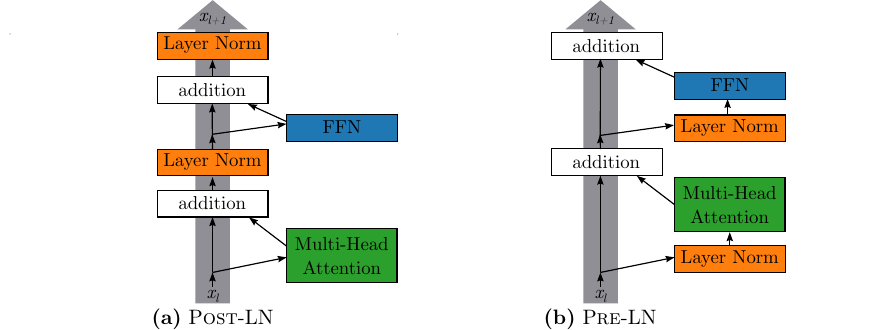}
    \caption{\textbf{Comparison between (a) \postlnfull \transformer layer (\postln) and (b) \prelnfull \transformer layer (\preln).} Figure recreated from \citet{Xiong2020} with permission.}
    \label{fig:pre_post_ln}
\end{figure}

\paragraph{Architectural modifications to \transformer models}
The \transformer model is the most widely used architecture for natural language processing tasks.
There are two popular versions of this model. First, the original version of the model from \citet{Vaswani2017}, the \postlnfull (\postln) \transformer, which places the \layernorm between the residual blocks. Second, the \prelnfull (\preln) \transformer \citep{Xiong2020}, which places the \layernorm inside the residual blocks.
\Cref{fig:pre_post_ln} shows the differences between the two architectures.
\begin{table}[!ht]
    \centering
    \begin{tabular}{lcc@{\hskip 0.03\textwidth}ccS[table-format=1.4]}
    \toprule
   \textbf{Training} & \multicolumn{2}{c}{\textbf{\preln}} & \multicolumn{2}{c}{\textbf{\postln}} & \textbf{Difference}\\
   \cmidrule(lr){2-3} \cmidrule(lr){4-5}
    \textbf{Algorithm} & \textit{Best} & \textit{Confidence interval} & \textit{Best} & \textit{Confidence interval} & \\
    \midrule
    \adamw & $31.3306$ & $31.0386 \pm 0.2811$ & $30.8098$ & $29.6819 \pm 1.1374$ & 0.5208\\
    \nadamw & $31.3381$ & $31.1153 \pm 0.2681$ & $30.8894$ & $29.6122 \pm 1.2608$ & 0.4487\\
    \nesterov & $29.2951$ & $28.9091 \pm 0.8864$ & $26.3211$ & $20.0433 \pm 7.0693$ & 2.9740\\
    \shampoo & $30.2051$ & $29.9168 \pm 0.2816$ & $29.9482$ & $29.0940 \pm 1.1800$ & 0.2569\\
    \bottomrule
\end{tabular}
     \caption{\textbf{Test set BLEU score for \prelnfull (\preln) and \postlnfull (\postln) architectures for different training algorithms.} The architectural modification of \preln vs. \postln affects \adamw, \nadamw, \nesterov, and \shampoo very differently as highlighted by the performance difference between the best-performing trials for \preln and \postln. \Cref{tab:pre_post_ln_ce} in the Appendix shows the analogous table with the cross-entropy loss instead of the BLEU score.}
    \label{tab:pre_post_ln_bleu}
\end{table}

This architectural modification affects \tas differently (\Cref{tab:pre_post_ln_bleu}). The experiment to produce \Cref{tab:pre_post_ln_bleu} was conducted as follows: we designed a customized search space for each optimizer and ran 100 trials using random samples from this search space for each of the architectural modifications. Each of the trials were run for a fixed number of steps (in this case 133K steps). The best BLEU score from each trial was chosen for the analysis. We performed Bootstrap analysis~\citep{EfroTibs93} on this data using a bootstrap sample size of 20 (corresponding to a plausible submission) and calculated the mean and standard deviation of the resulting distribution.

\subsubsection{Implications of Workload Sensitivity}
\label{sec:chal_workload_implications}

As we have seen above, seemingly minor changes to the training pipeline, such as model architecture tweaks or the exact placement of normalization layers, can have a large effect on the results different training algorithms achieve. 
As a consequence, empirical comparisons can fall into several failure modes and fail to achieve our desiderata of being convincing, informative, and practically relevant. First, it is very tempting to claim that a particular method provides a general improvement even though it is only useful on a very specific workload. Second, comparisons to previously published results can easily fail to exactly match the training pipeline details and end up with better or worse results for reasons unrelated to the methods being evaluated. Achieving a true apples-to-apples comparison usually requires reproducing earlier results in the same codebase and matching other details of the implementation and setup, an extremely labor-intensive prospect. Sometimes this issue might manifest as improvements improperly attributed to the training algorithm. Third, common benchmark workloads may not be very representative of the applications that are most important to the community, especially since the importance of various applications naturally waxes and wanes. Popular benchmark tasks also suffer from a selection bias that leaves us with unusually well-behaved training problems (\eg, unusually stable) or with training problems that co-evolved with the training algorithms used by whoever created them. For example, it shouldn't surprise us that \adam works especially well for model architectures built to train using \adam.

Our benchmark takes several steps to handle workload sensitivity. First and foremost, we fix a set of workloads (see \Cref{sec:rls_workloads}) and prevent pipeline changes that are not part of the training algorithm (see \Cref{sec:rls_submission}). By separating changes to the training algorithm and other parts of the pipeline, we can avoid the case where training speedups are incorrectly attributed, at the cost of having to occasionally split philosophical hairs on what parts of the training pipeline are off limits for submissions to alter. We also try to include a set of workloads that is relatively diverse, while still being representative of the most popular deep learning applications. However, workload diversity and relevance could always be better. Also, adding workloads comes at a cost both in engineering labor and in terms of our ability to run comprehensive experiments, sharply limiting what was feasible for the first version of our benchmark. Long term, we will need an even larger community effort to standardize workloads for training algorithm research and contribute measurements of new methods in a standardized way. Workloads should be implemented in open source code (ideally in multiple frameworks) and any variations or deviations should come with new names, in order to avoid confusion and misleading comparisons.

\subsection{We Cannot Compare Families of Training Algorithms, Only Specific Instances}
\label{sec:chal_comparing_algorithms}

Most \tas in \dl are not actually procedures we can run so much as algorithm templates we can instantiate with particular \hps.
Unfortunately, we cannot eliminate these \hps because they exist for a reason: using different \hp values for different \wls can lead to much better results (\Cref{sec:chal_comparing_hyperparams}).
To compare \tas with free \hps, we must either set the \hps to specific values or choose a specific procedure to tune them.
If we choose a tuning protocol, we should view the protocol, including the \hp search space, as part of the \ta definition, since the same algorithm template combined with a different \hp tuning protocol (or even just a different search space \Cref{sec:chal_comparing_search_spaces}) can produce significantly different results.
For a fair comparison, it is necessary to ensure that all \tas are tuned for the same tuning \emph{goal} (\Cref{sec:chal_comparing_tuning_goals}).

\subsubsection{Training Algorithms With Different Hyperparameters}
\label{sec:chal_comparing_hyperparams}

To illustrate the issues with using a single \hp setting for all workloads, we fixed a \ta and studied the impact of per-workload \hp tuning on the validation performance after a fixed number of training steps.
Assuming the \ta definition specifies a \hp search space $\mathcal{H}$, we can use \quasirandom search to sample a finite set $H\!\subseteq\!\mathcal{H}$ of candidate \hps.
Let $\val(w,h)$ be the validation metric value achieved on \wl $w$ by running the \ta with \hp setting $h$ for a fixed number of steps.
Furthermore, let $\val_H(w)\!=\!\min_{h \in H} \val(w,h)$ be the best validation score achieved among all \hp settings $h\!\in\!H$.\footnote{To simplify the notation, we assume that the validation metric should be minimized, \eg, as is the case for an error rate. For cases were a higher validation metric is better, \eg the \bleu score, we would instead simply define $\val_H(w) = \max_{h \in H} \val(w,h)$.}

Suppose now that we used the same \hp setting $h$ for every workload instead of tuning the \hps separately for each workload. Then we can consider the following measure of the worst-case relative performance degradation over all workloads
\begin{align*}
    \varphi(h, H) = \max_w \left| \frac{\val(w, h) - \val_H(w)}{\val_H(w)} \right| \; .
\end{align*}
Note that if $\val(w,h)\!=\!\val_H(w)$ for all workloads (\ie, if the same $h$ achieves the best validation score across all workloads), then $\varphi(h, H)$ will be zero. Generally the higher the value of $\varphi(h, H)$, the larger the relative performance gap between employing $h$ versus tuning over $H$. We can now find the $h\!\in\mathcal{H}$ with the lowest such degradation,\ie
\begin{align*}
    \Phi(H) = \min_{h \in H} \left( \varphi(h, H) \right) = \min_{h \in H} \left( \max_w \left| \frac{\val(w, h) - \val_H(w)}{\val_H(w)} \right| \right) \,.
\end{align*}
$\Phi(H)$ denotes the \emph{best} worst-case relative performance degradation achieved among the \hps $h\!\in\!H$.
For a set of \hps $H$, the quantity $\Phi(H)$ thus measures how much validation performance degrades from using any single setting of the \hps in $H$ that is shared across all workloads instead of tuning within $H$ to find the best per-workload setting.
Thus, we can view it as a measure of how much the \ta benefits from per-workload \hp tuning on a particular set of workloads, tuning over a particular set of \hp search points $H$.
By definition, $\Phi(H)\!\geq\!0$ and if it is zero then there is a \hp setting $h^\star\!\in\!H$ that achieves the best performance on every workload among all settings in $H$.
A larger value of $\Phi(H)$ suggests that the \ta benefits more from workload-specific \hp tuning.

We computed $\Phi(H)$ for four \tas on eight workloads (for the details of these \wls, see \Cref{tab:workloads} and \Cref{app:workload_details}). We constructed $H$ by sampling $100$ \hp settings with \quasirandom search from the tuning search spaces defined in \Cref{tab:target-setting-search-spaces}, for each \ta.
\Cref{tab:hparams_workloads_joint} shows the $\Phi$-values achieved by different methods using these specific search spaces. Note that $\Phi(H)$ is a random variable that depends on the random seeds in training and is a function of the set of \hps $H$ which were sampled from the search space $\mathcal{H}$.
In these experiments, we used a single training run per \hp, \ie, we used one sample to estimate the mean of $\val(w,h)$.

The results illustrate the performance gains that can be achieved by using \wl-specific tuning of \hps. We see that for every algorithm the $\Phi$-value is least $0.169$ which implies that for every \hp setting there is a workload for which the performance of that setting is at least 16.9\% worse than the performance of the optimal \hp setting for that particular workload. 
\Cref{tab:hparams_workloads_adam,tab:hparams_workloads_nadam,tab:hparams_workloads_nesterov,tab:hparams_workloads_heavyball} in the appendix show the performance of the \textit{optimal per-workload} \hps (\ie, $\val_H(w)$), the performance of the \textit{optimal overall} \hps per workload (\ie, $\val(w,h^\ast)$ where $h^\ast$ is the \hp that minimizes $\Phi(H)$), and their associated relative performance degradation for each \ta.
\begin{table}[!htp]
	\centering
	
\begin{tabular}{lS[table-format=1.6]}
	\toprule
    \textbf{Training Algorithm} & $\bm{\Phi(H)}$ \\
    \midrule
    \adamw & 0.195425 \\
    \nadamw & 0.169197 \\
    \nesterov & 0.230001 \\
    \heavyball & 0.239372 \\
    \bottomrule
\end{tabular}
 	\caption{\textbf{Using \wl-specific \hps can significantly improve the performance of \tas.} The $\Phi$-values obtained by different \tas with search spaces as defined in \Cref{tab:target-setting-search-spaces}. Higher values indicate the necessity of tuning to achieve good performance across multiple \wls.}
	\label{tab:hparams_workloads_joint}
\end{table}

\subsubsection{Training Algorithms With Different Hyperparameter Search Spaces}
\label{sec:chal_comparing_search_spaces}

Differences in tuning protocols can wreak havoc on empirical work. \citet{Choi2019} argue that they are the single most important factor explaining contradicting results from empirical optimizer comparisons. Learning rate schedules are an especially pernicious free parameter to tune because different algorithms result in different implicit schedules. Ideally, these implicit schedules would be separated from the other properties of the algorithm \citep{Agarwal2020}. There is an entire literature on semi-automated tuning algorithms for \hp tuning that includes Bayesian optimization and other black-box optimization techniques. However, even if a particular Bayesian optimization tool became standard, such tools still require search spaces as input and any procedure for constructing search spaces would need to be aware of the tuning budget \citep{Ariafar2022}.

Even if we fix everything about the tuning protocol except the search space, changes to the search space alone suffice to change training results. To illustrate this, we defined two search spaces for \adamw (\Cref{tab:search_space_importance_search_spaces}) and compared the best \hp points found from each search space for each workload in our benchmark. The first search space (\adamw \textsc{Narrow}) is completely contained within the second search space (\adamw \textsc{Broad}). In principle, with a large enough tuning budget, the broader search space can be no worse than the narrower one. However, at any particular tuning budget, either search space might end up performing better. Both search spaces seem reasonable \emph{a priori}. Indeed, they both contain good points, although without the benefit of hindsight one might be concerned that the narrow search space does not allow small enough values of weight decay.
\begin{table}[!htp]
    \centering
    \begin{tabular}{llcccc}
\toprule
\textbf{Search Space} & & \textbf{Learning Rate} & \textbf{Weight Decay} & $\bm{1-\beta_1}$ & $\bm{\beta_2}$ \\
\midrule
\adamw \textsc{Narrow} & &  [2e-4, 5e-3] & [2e-2, 0.5] & 0.1 & 0.999 \\
\adamw \textsc{Broad} & &   [5e-6, 2e-2] & [5e-6, 2.0] & [1e-3, 1.0] & 0.999 \\ 
\bottomrule
\end{tabular}
     \caption{\textbf{\Hp search spaces for two \tas using \adamw.} All \hps are sampled using a log-uniform distribution with the lower and upper bounds as shown in the table.}
    \label{tab:search_space_importance_search_spaces}
\end{table}

From each of the search spaces, we sampled $100$ points using \quasirandom search. Using these 100 points, we simulated tuning with a budget of $T$ trials by repeatedly sampling groups of $T$ trials, with replacement, from the 100 real results and taking the best trial within the group based on its performance on the validation set.
\Cref{tab:search_space_importance_results} shows the results of simulated tuning with a budget of $20$ trials for the two search spaces. The validation performance is the median over $1000$ simulations.
At this budget, the narrow search space achieves markedly better validation metrics across all workloads in our benchmark.
\begin{table}[!htp]
    \centering
    \scriptsize
    \begin{tabular}{
		l l S[table-format=2.6] S[table-format=2.6] S[table-format=2.6] S[table-format=2.6] S[table-format=2.6] S[table-format=2.6] }
\toprule
\textbf{Workload}         &              & \multicolumn{3}{c}{\textbf{\textsc{AdamW Narrow}}} & \multicolumn{3}{c}{\textbf{\textsc{AdamW Broad}}} \\ \cmidrule(lr){3-5} \cmidrule(lr){6-8}
&              & {Median}      & {$Q_1$} & {$Q_3$} & {Median}   & {$Q_1$} & {$Q_3$} \\
\midrule
\criteo            & \dlrmsmall   & \B 0.12401  & 0.123967       & 0.124025       & 0.124087 & 0.12396        & 0.124025       \\
\addlinespace
\fastmri     & \unet        & \B 0.734746    & 0.734590   & 0.734936   & 0.734311    & 0.734054    & 0.734522    \\
\addlinespace
\imagenet    & \resnetfifty & \B 0.23256     & 0.23094    & 0.23330    & 0.24334     & 0.23904     & 0.24708     \\
& \vit         & \B 0.21992  & 0.22118        & 0.22118        & 0.23616  & 0.22694        & 0.24038        \\
\addlinespace
\librispeech & \conformer   & \B 0.075989    & 0.075962   & 0.076817   & 0.080673    & 0.078963    & 0.087340    \\
& \deepspeech  & \B 0.112706 & 0.112353       & 0.113485       & 0.120674 & 0.116902       & 0.127974       \\
\addlinespace
\ogbg & \gnn         & \B 0.28214  & 0.281595       & 0.284034       & 0.276307 & 0.275642       & 0.279285       \\
\addlinespace
\wmt  & \transformer & \B 31.3523  & 31.2824        & 31.3946        & 30.9950  & 30.9129        & 31.1748 \\
\bottomrule
\end{tabular}
     \caption{\textbf{Performance across multiple \wls for \adamw with two different \hp search spaces.} Shown are the median, as well as the lower and upper quartiles ($Q_1$ and $Q_3$) of the best observed validation metric. The results are for a budget of $T\!=\!20$ trials (see \Cref{tab:search_space_importance_25p_t5} for results with $T\!=\!5$) across $1000$ simulations. At this budget, \textsc{AdamW Narrow} performs significantly better across all test \wls.}
    \label{tab:search_space_importance_results}
\end{table}

Although it might seem obvious that the results would depend on the search space, it is still not standard within the literature to report exact tuning protocols or even supply search space details. The narrow and broad search spaces we considered here could have instead had less---or even no---overlap and produced even more dramatic differences in results. It is extremely easy to select a particular search space for tuning a baseline and then claim that another algorithm outperforms it without including appropriate caveats about the search space. Our experiment here shows that even if we consider two very reasonable search spaces, merely neglecting to select a search space that matches the tuning budget could subtly weaken a baseline and befuddle a comparison between \tas.

\subsubsection{Training Algorithms With Different Tuning Goals}
\label{sec:chal_comparing_tuning_goals}

Even if we fix everything about the tuning procedure \emph{including} the search space, differences in tuning \emph{goals} can lead to unfair comparisons. Specifically, tuning to achieve the best validation error within a fixed training time budget is \emph{not} the same as tuning to achieve a fixed validation error as fast as possible. Suppose some previously published result achieves a particular validation error rate on a particular workload after training for a certain amount of time using Algorithm X. Now suppose that we demonstrate that Algorithm Y can achieve the same validation error in substantially less time (on the same hardware). In many cases, this kind of comparison, although seemingly innocuous, will be unfair to Algorithm X because the result with Algorithm X was from a paper that was not trying to minimize training time, but was instead engaged in an implicit competition to get the best possible validation error, within their available budget. 

\begin{figure}[!ht]
    \includegraphics[width=\textwidth]{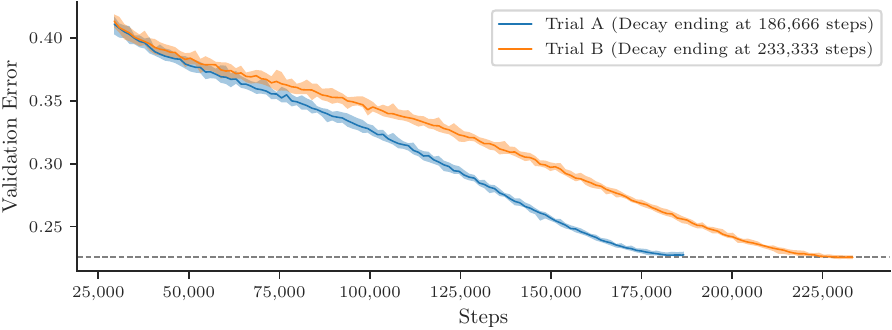}
    \caption{\textbf{For a fair comparison, training curves need to be tuned for the same criterion.}
    	Two \adamw training runs (\colorline{SNSblue}, \colorline{SNSorange}) for \resnetfifty on \imagenet using \hps tuned within the same search space, but using different step budgets. Since the \cosinelrdecay schedule stretches with a longer step budget, we see ``slower'' training caused by the larger step budget (\colorline{SNSorange}). For each of the \hp settings, we ran $20$ different random seeds to create min/max error bounds around a median trajectory (\colorsquare{SNSblue_shaded}, \colorsquare{SNSorange_shaded}). The dashed gray line (\dashedcolorline{gray}) denotes the best median validation error achieved by both training runs. See \Cref{app:exp_details_chal_comparing_tuning_goals} for experimental details.
    }
    \label{fig:different_tuning_goal}
\end{figure}

We tuned the \hps for \adamw for \resnetfifty on \imagenet for two training step budgets, $186,666$ and $233,333$. Both studies use the same search space and, by using the same seed, the same $100$ \hp samples. They also used the same \cosinedecay learning rate schedule with a linear learning rate \warmup. However, the \cosinedecay schedule depends on the maximum number of training steps and therefore differs between the two studies (for the complete search space see \Cref{tab:target-setting-search-spaces}, \adamw). In both studies, the \hp setting achieving the best validation error happened to be the same (see \Cref{tab:different-step-budget-opt-hparams} in \Cref{app:exp_details_chal_comparing_tuning_goals}). We then retrained this best trial for both studies using $20$ different random seeds. \Cref{fig:different_tuning_goal} shows the best validation error achieved so far versus the training step for Trial A (\colorline{SNSblue}, step budget of $186,666$ steps), and Trial B (\colorline{SNSorange}, step budget of $233,333$ steps).
Both trials achieve nearly identical validation errors ($22.4\,\%-23.0\,\%$ depending on the seed), but Trial A is clearly (by design) much faster. Training longer tends to very slightly improve the validation error (median of $22.6\,\%$ vs a median of $22.7\,\%$), but this improvement is hard to detect with the variance across seeds. Furthermore, if we compare larger and larger pairs of training step budgets, any difference between the budgets will be swamped by the variance across different runs. This experiment shows that we can get (roughly) the same validation error result faster, \emph{simply by setting a lower training step budget}.

Researchers trying to achieve the best headline validation error number to publish will naturally give their experiments a generous step budget to make their lives easier, and not try to find the minimum step budget that can still reproduce their result. Thus we might imagine published results more like Trial B when they are being tuned for the best validation error using a somewhat arbitrary, but generous, step budget. In general, learning rate decay schedules are necessary for the very best results, and we should expect the best validation error rates to be achieved near the end of training when the learning rate has already been reduced. However, common decay schedules make it extremely dangerous to assume that some state-of-the-art result, in terms of out-of-sample error, was achieved with a near-minimal number of training steps. Unfortunately, that is precisely what we are implicitly assuming when we compare results tuned to minimize training time to reach a particular error rate with results tuned to achieve the best possible error rate given a particular time (or step) budget.

In light of the strong effect tuning protocol details will have on the results of any empirical comparison between training algorithms, we have no choice but to view the tuning protocol as part of the algorithm. Additionally, we can only compare two results that are tuned to optimize the same criterion, \ie, two results that are both tuned to minimize time-to-result or are both tuned to minimize validation error at a fixed time budget. Once we fold the tuning protocol into the definition of the method, we can study one algorithm with two different tuning protocols as if it was two separate methods. Algorithm designers could then also provide guidance on how the methods they introduce should be tuned. Currently, new training algorithms rarely come with sufficient guidance on how they should be tuned in various budget scenarios. Despite not needing much (or any) tuning being a common selling point, we are not aware of any popular training algorithm for neural networks where this property has been precisely defined and convincingly demonstrated, although this would certainly be a valuable contribution.Unfortunately, the tuning protocol is necessarily a function of the tuning budget, which makes it difficult for algorithm designers to specify how their method is best tuned in all possible budget scenarios. Nevertheless, even providing a short list of budget scenarios along with tuning guidance would improve the situation dramatically, especially since the largest tuning budget scenarios are probably less critical to address for practitioners. Ultimately, methods that are easier to tune---or even fully self-tuning---would be extremely valuable if they could be supported by compelling experimental evidence. \Cref{sec:rls_tuning} describes our approach to tuning in detail.

Even if we view results as conditional on a tuning protocol, publication bias and other similar selection bias effects can, in effect, amount to implicit tuning. Ultimately, the root of the problem is that quantifying tuning effort when developing a training algorithm is nearly impossible. If we try 1000 different methods on a small set of workloads and find one that seems to work across this small set ``without much tuning'', we might just have obfuscated the tuning process. However, ``off the books'' implicit tuning should not bother us if the resulting training algorithms generalize to new workloads. Therefore, the solution to this issue is a combination of various generalization incentives and safeguards: preventing workload cherry-picking by standardizing sets of benchmark workloads, sharply limiting per-workload adaptation and tuning in empirical comparisons, using larger sets of more diverse benchmark workloads while measuring performance aggregated across workloads, and perhaps even generating novel workload variations randomly to create ``held-out'' workloads.

\subsection{Strong Baselines Are Far Too Hard to Obtain}

Researchers proposing new training algorithms have total control over the baselines they compare with their algorithms. Even with everyone acting in good faith, it is very easy to accidentally make an experimental design choice that gives a new algorithm an unfair advantage. For example, we could select a learning rate schedule family that works well with a new technique and not even know how appropriate it is for our baselines. Or we might select workloads where controlling overfitting is essential for good results, but not apply sufficient regularization to a baseline that minimizes training loss \emph{faster} than the novel method we are studying. There are essentially limitless possible choices we can make when designing experiments that might inadvertently hamstring our baselines and show overoptimistic improvements for some novel method. Researchers just have far too many choices to make when comparing training algorithms. Even worse, the least careful experiments not only take the least effort to conduct, but also tend to produce the most impressive-seeming results.

Although the problem of weak baselines exists in other parts of deep learning (and in machine learning as a whole), training algorithm comparisons seem particularly fraught compared to, say, model comparisons on a single task (\eg, what is the best neural network architecture for large vocabulary speech recognition). There are many factors at play, including the challenges described above, but experimental methodologies for studying training algorithms seem less mature as well. 
There is a long tradition of competitions surrounding particular datasets, \eg, the \imagenet competition \citep{Russakovsky2015}, that have at least produced a level playing field within the narrow boundaries of the original rules. We can apply the same approach to training algorithm comparisons. Direct competition under a shared set of rules can cut through the Gordian knot of tangled incentives and inconsistent experimental protocols.

\section{Related Work}
\label{sec:intro_relatedworks}

Our goal is to create a benchmark for \nn \tas.
Related work can be grouped into three broad categories:
Earlier efforts to benchmark \tas for \dl (\Cref{sec:rw_benchmarks});
previous work identifying issues with existing approaches for evaluating \dl \tas (\Cref{sec:rw_issues}); and, finally, relevant prior work that makes a case for, or against, particular \tas (\Cref{sec:rw_benefits}), with a special focus on disagreements in the literature.

\subsection{Existing Benchmarks}
\label{sec:rw_benchmarks}

Domain-specific benchmarks and challenges have driven advances in \ml for decades \citep{Garofolo1993,Krizhevsky2009,Deng2009,Lin2014,Russakovsky2015,Panayotov2015,Bojar2015,Zbontar2018,wu2018,Hu2020,dwivedi2020,dong2020}. Such benchmarks generally provide a curated \dataset---usually with pre-defined training, validation, and test splits---and specify what performance metrics should be used to measure progress. The research community uses benchmarks like these to demonstrate progress on specific tasks (\eg, image classification, speech recognition, machine translation). Ultimately, domain-specific benchmark \datasets are a component of almost all \tas benchmarks, including ours. Progress in domain-specific benchmarks often involves a combination of new models, regularization techniques, and training algorithms.
In many cases, however, it is difficult to disentangle how much improvement is due to, \eg, changes in the model versus the training procedure or tuning effort \citep{Bello2021}.

A separate set of benchmarks thus aims to measure specifically the performance of \tas on different systems, or in different frameworks. The \mlperftraining benchmark \citep{Mattson2020},\footnote{\url{https://mlperf.org/training-overview}} which grew out of the \dawnbench benchmark~\citep{Coleman2017}, has the goal of evaluating system performance for \nn training. The \emph{Closed Division} of the benchmark aims to measure the performance of \ml hardware and systems by requiring mathematical equivalence to a reference implementation, while still allowing submissions on different hardware. Similar to the \mlperftraining benchmark, the \dlbs,\footnote{\url{https://github.com/HewlettPackard/dlcookbook-dlbs}} \deepbench,\footnote{\url{https://github.com/baidu-research/DeepBench}} and the \tbdnn\citep{Zhy2018} also focus on evaluating \ml \emph{systems} (accelerators, operating systems, and frameworks) using fixed, well-established workloads and \tas.

Unlike the Closed Division of the \mlperftraining benchmark and other similar systems benchmarks, the \emph{Open Division} of \mlperftraining, has slightly more overlap with our goals. In this division, in addition to running on different systems, submissions are allowed to modify the \ta and model. However, allowing arbitrary hardware, potentially at radically different scales, makes it impossible to distinguish improvements due to algorithms from those due to extra computation; in fact, an algorithm may not perform the same at scales. Additionally, by allowing model changes, there is also no way to isolate improvements due to \ta modifications. Generally, since submissions are viewed as independent per-\wl entries, there is no incentive to avoid hyper-specific changes that only help one particular benchmark \wl. Even if a participant provides submissions for all \wls, they are free to use completely different \tas or models for each \wl and perform unlimited \wl-specific tuning. 

\deepobs~\citep{Schneider2019}, a software package and set of test problems for benchmarking \dl optimization methods, aims to simplify research and development of \tas. \deepobs recommends reporting three key performance indicators: the final predictive error after a fixed number of steps, the wall-clock runtime per optimization step, and the ease of \hp tuning. Since it positions itself as a research tool, the package does not provide guidance on how these three performance indicators should be weighted against each other or aggregated across the provided test problems. While \deepobs fixes the \wls (addressing the challenges mentioned in \Cref{sec:chal_wl}), it contains only one large-scale \dl \wl (training \resnetfifty on \imagenet). The other workloads in \deepobs are mainly smaller-scale \dl tasks (\eg, image classification on \mnist, \cifar, and \svhn). Synthetic problems and smaller \wls are useful for rapid experimentation, and thus helpful to researchers. However, conclusions drawn from these problems do not always generalize to larger \wls (\eg, larger models and/or datasets).  Furthermore, \deepobs leaves the user in control of many parts of the training protocol, such as the tuning budget, and thus it is ultimately up to the user to ensure a fair comparison.

In follow-up work, \citet{Schmidt2021} leverage the \deepobs framework to perform an empirical comparison of fifteen popular \tas. For this purpose, they fix the benchmarking protocol by comparing the methods at four different tuning budgets. This empirical comparison is not a competitive benchmark. Instead,  \citet{Schmidt2021} determined the hyperparameter search spaces of the \tas themselves, aiming for a ``fair representation'' of each method based on statements made in the original publication, and not necessarily trying to make each method perform at its best (as the original authors may have done in a competitive setting). In addition, some of the issues of \deepobs described above carry over to the study by \citet{Schmidt2021}. Notably, their empirical comparison uses mainly small-scale \dl problems, and there is no aggregated score across test problems.

\citet{Moreau2022} introduce the \benchopt software suite for evaluating \ml optimization methods. They focus on features that a software platform should support (\eg, modularity, extensibility) to enable fair and reproducible comparisons. However, at the time of writing, although \benchopt includes 16 different optimization problems, only one is a deep \nn training \wl.

\subsection{Methodological Critiques of Training Algorithm Comparisons}
\label{sec:rw_issues}

\citet{Bartz-Beielstein2020} present a series of recommendations for defining benchmarks for optimization methods for several families of optimization problems, including \dl optimizers. Although these recommendations generally align with the benchmark described in this paper, as we discussed in \Cref{sec:challenges} and again in \Cref{sec:rules}, below, a deep \nn \ta benchmark should not be restricted solely to optimization methods, and should also encompass aspects like regularization and data sampling.

Several recent works emphasize the importance of \hp tuning in making fair comparisons of different \tas. \citet{Schneider2019} recommend finding and reporting the best-performing \hps for each test problem in \textsc{DeepOBS}. \citet{Sivaprasad2020} advocate that researchers introducing a new \ta should always provide a \wl-agnostic prior distribution over \hps that can be used in conjunction with random search. They illustrate how the relative ranking of different optimization methods may change on a given \wl depending on the \hp tuning budget. \citet{Choi2019} illustrate how rankings can be sensitive to which \hps are tuned, \eg, showing that tuning \adam's $\beta_2$ and $\epsilon$ can significantly improve performance compared to simply using their default values. They also illustrate that the choice of search space parameterization can influence the effectiveness of \hp tuning, for example, explaining why tuning the relative initial learning rate $\alpha_0 / \epsilon$ of \adam is more efficient than tuning $\alpha_0$ and $\epsilon$ independently.

\subsection{Disagreement over Training Algorithms: The Case for Clear Benchmarks}
\label{sec:rw_benefits}

Several of the studies discussed in \Cref{sec:rw_issues} are  motivated by earlier work making contentious claims about certain \tas. Below, we review several examples of disagreements about \tas in the literature, since they demonstrate the opportunity for trusted, competitive benchmarks to promote joint progress.

As one prominent example, \citet{Wilson2017} showed that \sgd converges to the maximum margin solution and provide a toy example where \adagrad finds a solution which generalizes poorly while \sgd finds a solution that generalizes well. The same work also presented empirical results where adaptive methods such as \adagrad, \rmsprop, and \adam generalized much worse than \sgd with \momentum. \citet{Choi2019} subsequently argued that \citeauthor{Wilson2017}'s empirical conclusion is only valid under what they consider restrictive conditions, namely fixing \adam's $\beta_2$ and $\epsilon$ parameters to their default values, and only tuning a constant learning rate. In contrast, tuning all of \adam's \hps and incorporating a learning rate decay schedule allows \adam to perform comparably or even slightly better than \sgd with \momentum. Additionally, \citet{Agarwal2020} argued for the importance of separating the effects of search direction and per-layer learning rate scaling when comparing optimizers, showing that conflating the two also accounts for some of the findings of \citet{Wilson2017}.

As another example, \citet{Liu2020} observed that the effective per-coordinate scaling factors can have high variance early in training, which can lead to unstable training dynamics. They proposed the rectified \adam optimizer (\radam) to compensate for this, and reported several experiments showing that \radam outperforms standard \adam with learning rate warmup on image classification and machine translation tasks. Subsequently, \citet{Ma2021} showed that using \adam with an appropriately tuned learning rate warmup performs comparably or slightly better to \radam on similar tasks.

There are also disagreements about whether optimizers that use off-diagonal curvature information are useful when training neural networks. \citet{hinton2006} proposed a deep autoencoder model for reducing the dimensionality of the input data. \citet{martens2010} showed that the Non-Linear Conjugate Gradient method \citep{nocedal1999} is quite ineffective in minimizing the training loss in this problem. In contrast, non-diagonal methods such as truncated-Newton \citep{nocedal1999} provide a significant advantage. Since \citet{martens2010}’s work, the deep autoencoder problem has served as a standard benchmark for comparing diagonal and non-diagonal optimization techniques where the gap between the two methods is expected to be fairly substantial \citep{martens2015, goldfarb2020,anil2020,ren2021,ren2021b,zhao2022,Bae2022amortized}. However, \citet{amid2022} found that with sufficient hyperparameter tuning, diagonal preconditioning based methods such as \rmsprop worked reasonably well on this problem compared to non-diagonal preconditioning based methods such as \kfac \citep{martens2015} and \shampoo \citep{gupta2018,anil2020}. \citet{amid2022} thus argued that algorithmic improvements to diagonal methods can practically eliminate the gap between diagonal and non-diagonal methods.

Finally, there has been substantial disagreement in the literature about whether new training algorithms are necessary as the batch size increases. Although some of this disagreement is due to imprecise claims and experimental setups that implicitly demand perfect batch size scaling \citep{Shallue2019}, various training algorithms have been proposed to handle the supposed problem of ``large batch training.'' For example, \citet{You2017} and \citet{You2019} introduced the \lars and \lamb optimizers, respectively, and argued they were necessary at larger batch sizes. Subsequently, \mlperftraining adopted \lars for its \resnetfifty on \imagenet workload and various submitters have demonstrated impressive training speed results. Nevertheless, \citet{Nado2021} have since shown that, with appropriate \hp tuning, it is possible to obtain similar training speed results using \sgd with \nesterov \momentum with batch sizes up to $\num{32768}$. They also showed that stronger \adam baselines could outperform \lamb results on BERT \citep{Devlin2019BERT} pre-training, and argued that there was no convincing evidence that \lars and \lamb should be used over standard techniques.

The examples above illustrate how the challenges in benchmarking \tas discussed in \Cref{sec:challenges} directly affect the training algorithms community. Perhaps most critically, they emphasize the importance of tuning hyperparameters in a fair and consistent way to give each algorithm the best chance to perform well~\citep{Choi2019,Sivaprasad2020}. Although this may sound straightforward, substantial care must be taken when defining the \hp search space for each algorithm.
Framing \ta comparisons as a competition has the crucial advantage that each participant individually strives to make their method work best under the constraints of the contest, with one participant's method becoming another's baseline.
In contrast, the status quo in the literature is for researchers to make uncontrolled changes and depend on the vicissitudes of the (noisy) peer review process to enforce some notion of a ``fair'' comparison with previous work, resulting in confusing comparisons with baselines that tend to be much too weak.  
Working from a common, open codebase enables researchers to independently reproduce and verify the claims of others, and also makes it easier for entrants to the competitive benchmark to share their submissions with the community for future comparisons and studies.

\section{Rules}
\label{sec:rules}

The goal of our benchmark is to identify \textbf{general-purpose \nn \tas} that can speed up training on new \wls.
Our benchmark measures time-to-result (\Cref{sec:rls_timetoresults}) on a fixed hardware configuration.
To ensure that the benchmark results have real-world relevance, we define goal error rates based on the validation and test sets instead of the training loss.
In order to isolate the effect of the \ta, submissions must adhere to a specific API (\Cref{sec:rls_submission}) and cannot make alterations outside a limited number of functions.
To incentivize generally useful algorithms, we require that a single \ta simultaneously performs well across multiple \wls (\Cref{sec:rls_workloads}) without manual \wl-specific adaptation.
Instead, any adaptation to the \wls should either be possible with generic tuning methods (\Cref{sec:rls_external_tuning}) or be performed as part of the timed training process (\Cref{sec:rls_self_tuning}).
All \wls are considered when calculating an aggregate benchmark score of the \ta (\Cref{sec:rls_scoring}).
The resulting benchmark score is intended to serve as an estimate of the performance of a \ta on unknown \wls.

\sloppypar In the following sections, we describe the essential elements of our benchmark rules and explain the reasoning behind them. This section is based on the rules at the time of writing. As the benchmark evolves, the rules may also change, and the most up-to-date, complete rules can be found at \rulesurl.

\subsection{A Time-to-Result Benchmark}
\label{sec:rls_timetoresults}

A submission's score is based on the time to reach the target validation and test scores on each \wl. Training is considered complete, and the clock stops, as soon as a submission achieves the validation and test targets. For practical reasons, submissions are limited by a maximum allowed \runtime on each \wl. If a submission fails to achieve the targets within this maximum \runtime it will receive an infinite training time on this trial.\footnote{Depending on the tuning \ruleset, a submission may get several trials or only one trial per workload; see \Cref{sec:rls_tuning} for more details.}
Although setting these targets will always be contentious to some degree, we need a systematic procedure to determine target validation and test scores that are competitive (ideally near the state of the art), while being achievable within a reasonable time budget.

For a given target-setting \runtime budget, we defined the validation and test targets of a \wl based on what can be reliably achieved using standard methods. Specifically, we used four popular \tas (\heavyball, \nesterov, \adamw, and \nadamw) tuned with \quasirandom search \citep{Bousquet2017} on hand-engineered, \wl-agnostic search spaces. For each \ta and \wl, we ran $200$ trials for the full target-setting \runtime budget\footnote{For simplicity, we converted these target-setting \runtime budgets into step budgets since all four \tas used for target-setting happen to have nearly identical average step times.} and determined the best combination of \ta and \hp settings by finding the trial that achieved the best validation error.
We re-ran this combination of \ta and \hps $20$ times with different random seeds, taking the median of the best achieved validation errors across seeds to obtain a \emph{validation} target. Out of the $10$ repeated runs that achieved this validation target, we took the worst achieved test error across seeds as our \emph{test} target. Taking the median validation performance after rerunning the best \hp point prevents our procedure from selecting a lucky outlier. Our protocol defines both \emph{validation} set targets and \emph{test} set targets in order to implement the external tuning \ruleset, as described in \Cref{sec:rls_external_tuning}.
Exact tuning details and results can be found in \Cref{sec:exp_target_setting}.
It is important to note that the target-setting procedure does not constitute a valid submission since it
uses $200$ trials instead of the $20$ trials available in the external tuning \ruleset.

The procedure outlined above requires us to determine two \runtime budgets for each \wl. First, we need to set the \emph{maximum allowed \runtime} for the submissions. Ideally, this \runtime budget would be infinite, as it would allow us to accurately gauge the time required for a submission to successfully train each \wl. However, for practical reasons, we must limit this budget.
Second, we need to set a \emph{target-setting \runtime budget} for the target-setting procedure described in the previous paragraph.
We decided to use different \runtime budgets for the submissions and target-setting. Specifically, the target-setting \runtime budget is set to $0.75 \times$ the maximum allowed \runtime for submissions. By allowing submissions a more generous \runtime budget, they have some leeway to spend extra time on certain \wls, if they can compensate for it on other \wls.

When setting maximum \runtimes for a \wl, it is important to find a balance between challenging and achievable targets. Increasing the maximum \runtime may improve the performance but it also increases the overall \runtime of the benchmark. To balance the dual objectives of stringent targets, near the state of the art in the literature, and making benchmark submissions practical to evaluate on the full suite of \wls, we aimed to limit the combined \runtime of all fixed \wls to $100$ hours on the benchmarking hardware. We used both preliminary experiments and published results on the \datasets and models of our \wls as a guide in allocating this combined \runtime budget to the individual \wls.
The allowed maximum \runtimes for the submissions on each \wl are shown in \Cref{tab:workloads}. \Cref{sec:exp_target_setting} discusses how the benchmark targets (which were set using $0.75 \times$ the \runtimes presented in \Cref{tab:workloads}) compare to results from the literature.

\subsubsection{Measuring Runtime}
\label{sec:rls_wallclock}

We selected elapsed real time, or wall-clock time, as our measure of runtime for training. We made this choice to maximize the practical relevance of our timing measurements and to avoid imposing new restrictions on how training algorithms could operate. To make meaningful comparisons of different algorithms in terms of wall-clock time, all algorithms must be run on a standardized hardware system (discussed further in \Cref{sec:rls_hardware}). 

The research literature contains examples of several alternatives to directly measuring wall-clock time. In some cases, researchers count training steps, number of forward passes, gradients, or some other abstract notion of iterations instead of an all-encompassing time measurement. Counting steps is convenient when iterations have the same, consistent cost during a single run and we can measure the average time-per-step for different algorithms easily. More generally, abstracting away the hardware and system conditions is appealing when it is possible, but such abstract runtime proxies are sadly not an option for a general training algorithms benchmark. Counting iterations is meaningless since submissions are free to redefine what a single step means. Counting gradient computations is similarly meaningless since submissions can vary in what derivatives (if any) they compute, or even use radically different batching schemes that include intelligent data selection.

Abstract notions of steps completed or examples processed do not necessarily reward algorithms that are the most useful in practice. Some algorithms might cleverly reclaim idle accelerator time while waiting for new data (\eg, by applying Data Echoing \citep{choi2019_data_echoing}). Some algorithms might be especially memory efficient and thus able to use larger batch sizes and better exploit hardware parallelism. Conversely, some algorithms might be impractical because they require too much extra memory.

Measuring wall-clock time has some disadvantages. Although we care about implementation quality to some extent, our intention is not to make a software benchmark since that goal is better served by comparing mathematically equivalent programs. Furthermore, variation in network congestion when reading data or writing results is of little interest, nor do we care about benchmarking unrelated processes running on the system and how they interfere with the training program. We expect that by using a standardized hardware system described below (namely, a single server with multiple GPUs), and by having the training program be the only significant (\ie, computationally intensive) program running on the system, that these additional factors will not substantially affect the measured runtimes.

\subsubsection{Standardizing Benchmarking Hardware}
\label{sec:rls_hardware}

To fairly compare wall-clock \runtimes of different algorithms, the times should be measured on a standardized training system (\eg, hardware accelerators, memory, CPUs, interconnect) using a standardized execution environment, including consistent software versions.
For the initial version of the benchmark, we selected a system with \benchmarkinghardware with 16GB of VRAM per card since it is widely available on major cloud computing providers and offers a good compromise between performance and affordability.
This official benchmarking system only needs to be used for final timing runs. Tuning a submission only requires a comparison between different \hp settings of a single \ta, so it is fine to use a different but consistent system for tuning experiments.

Inevitably, accelerators and the systems available on the market will change in the future, so future iterations of the benchmark may adopt new benchmarking systems or even support multiple ``weight classes'' of systems. As long as future benchmarking systems are strictly more capable, especially in terms of accelerator memory capacity, it should be relatively straightforward to rerun the baselines and the top-performing previous submissions on new systems, at least using the old batch sizes to provide a pessimistic bound.

\subsection{Specifying a Training Algorithm}
\label{sec:rls_submission}

The rules define a precise API to be used when specifying a training algorithm submission. A submission to the competitive benchmark must define four \emph{submission functions}. The submission functions allow submitters to define how the submitted \ta updates the model parameters (\inlinecode{update\_params}) and how data are selected for training (\inlinecode{data\_selection}).
Furthermore, submitters can initialize the \ta's state (\inlinecode{init\_optimizer\_state}) and must provide a batch size for each workload (\inlinecode{get\_batch\_size}).
An implementation of the submission functions may make use of a limited API to get some basic information about the workload.
A detailed description of the submission functions signatures' and the benchmark API can be found in the rules.\footnote{See the ``Valid Submission'' Section in \url{https://github.com/mlcommons/algorithmic-efficiency/blob/main/RULES.md\#valid-submissions}.}

In addition to implementing the four submission functions, a \ta may have hyperparameters that will be tuned following one of the rule sets described in \Cref{sec:rls_tuning}. Submissions to the external tuning ruleset described in \Cref{sec:rls_external_tuning} must also specify a hyperparameter search space.

Apart from defining the submission functions and hyperparameter search space, a submission may not modify any other parts of the training pipeline. The rest of the training program (\eg, implementations of data pipelines, model architecture, training loop, calculation of validation and test metrics, and measurement of runtime) are implemented by the benchmark.
We intentionally restrict which parts of the training program a submission may modify for two main reasons.
First, we prohibit certain types of changes in order to isolate speedups due to \ta changes (\Cref{sec:rls_isolate_ta}). Second, we hope to deter submissions that are over-specialized to particular benchmark workloads and are unlikely to be generally useful on new problems (\Cref{sec:rls_general_submissions}).

The benchmark exposes a limited API of functions that a submission may call, within the four submission functions, to get information about a training workload at execution time. The workload-specific information available to submissions is restricted, since the goal of this benchmark is to identify training algorithms that are generally useful across many workloads.
One of the ways in which a submission may be workload-specific is by providing a different batch size to be used for each workload. This is allowed since the benchmark hardware, including accelerator memory capacity, is fixed, and different \tas may involve storing and updating different state.

It is impossible to design an API and comprehensive rules that would prohibit all possible cases of submissions circumventing the spirit of the benchmark.
Instead, we will prohibit these submissions in the abstract and will defer rulings about specific submissions to a ``spirit [of the rules] jury."
Similarly, there may be modifications that we would allow in principle, but which are currently not practically feasible within the provided API.
Since this is a practical benchmark, we must accept that we cannot guarantee a perfect overlap between what is allowed and what is possible.
However, we may modify the API in future iterations of this benchmark to accommodate a larger set of allowed modifications.

\subsubsection{Isolate the Training Algorithm}
\label{sec:rls_isolate_ta}

A \tas benchmark should prohibit modifications outside of the \ta in order to disentangle improvements due to the \ta from other beneficial pipeline changes (see \Cref{sec:chal_workload_implications}).
However, exactly what constitutes part of the \ta is not always clear.
The distinction between the model and the \ta can be quite subtle, as well as the distinction between improving the implementation quality of the submission versus improving its software dependencies in a way that could apply to all submissions.

\paragraph{Model vs.~training algorithm.}
There is not always an obvious separation between the model and the \ta.
In our benchmark, we use the basic rule of thumb that anything that can be applied to a generic \wl is part of the \ta.
On the other hand, if something only applies to some \wls or is otherwise inherently \wl-specific, it should not be considered part of the \ta.

One delicate area is regularization.
Because our benchmark focuses on validation and test set performance, we need to include regularization as part of the \ta in our benchmark and allow submitters at least some control over it since some \tas may implicitly have a regularizing effect.
However, submissions can only be allowed complete control over model-agnostic regularization.
Therefore, we allow submissions to tune the \emph{strength} of regularization methods predefined by the workloads, and we do not allow submissions to introduce new regularization \emph{strategies} that require modifications to the data preprocessing or model architecture.
For example, submissions can set or tune the dropout rates of models that already have dropout layers.
However, they cannot introduce additional dropout layers, as this would require knowledge of the model architecture.
Similarly, the submitted \tas receive raw loss values and can add any \wl-agnostic regularization term, \eg, $\normltwo$ regularization of the desired strength.

Data augmentation strategies often improve generalization performance.
However, most data augmentations strongly depend on the input data modality and are therefore \wl-dependent.
Submissions cannot introduce new data augmentation strategies to the benchmark \wls. However, they do have control over batching and could potentially filter, reorder, or otherwise prioritize training data.

Another tricky technique to handle is \batchnorm, which can be seen either as a training algorithm component or---as it is commonly implemented---as a model layer and thus part of the model architecture.
In our benchmark, submissions cannot introduce additional \batchnorm layers to the \wls or change the location of existing normalization layers.
However, they do have control over whether the \batchnorm statistics should be updated.
This is important, for example, for line search approaches that search over multiple candidate updates before applying the most appropriate one.
In this case, such a submission might not want to update the \batchnorm statistics after each candidate evaluation, but only once an acceptable point has been selected.

\paragraph{Introducing software dependencies.}
Submissions are not allowed to use software engineering approaches to speed up low-level, primitive operations in \jax, \pytorch, their dependencies, or the operating system. For example, it is prohibited to introduce new compiler functionality, using faster GPU kernels, or make similar modifications that could generally benefit any submission.
Submissions must also use the versions of \pytorch or \jax (and their dependencies) specified by the benchmark.

Submissions are free to add software packages that support novel algorithmic and mathematical ideas, as long as they do not circumvent the intention of the benchmark.
For example, submitters are allowed to use packages such as \backpack \citep{Dangel2020}, which extracts additional information from the backward pass. We also recognize that the way a method is implemented will impact its performance in the benchmark, and it is generally acceptable to make clever and efficient use of the \jax and \pytorch APIs from within the submission functions. For example, it would be acceptable to use CUDA streams to schedule the transfer of data from CPU to GPU while performing other computations. However, under the rules there are periodic untimed model evaluations that do not contribute to the submissions score and it would not be acceptable for a submission to schedule asynchronous computations that are being performed during these untimed evaluations.

\subsubsection{Incentivizing Generally Useful Submissions}
\label{sec:rls_general_submissions}

We want to disallow submissions if they clearly violate the spirit of the benchmark, even if these submissions perform well in our benchmark.
Most importantly, this includes overly benchmark-specific methods that cannot be applied to generic \dl \wls.

It is impossible to define rules that clearly distinguish between allowed and prohibited submissions in all possible scenarios.
This section provides some guidelines to clarify whether or not a submission violates the spirit of the rules, and thus should be disqualified by the spirit jury.
As a rule of thumb, a submission should be allowed if it will run and do something reasonable on unseen \wls without requiring additional human engineering effort.
Two essential questions can guide this distinction:
\begin{enumerate}[itemsep=0mm]
    \item What \textbf{information} is used by the submission?
    \item What \textbf{action} is the submission code taking based on this information?
\end{enumerate}
Generally, both parts are needed to decide if a particular piece of code is within the spirit of the rules.
Below are some specific examples intended to illustrate the policy.
Additional cases of allowed and disallowed submissions, along with further clarifications, can be found in the complete rules.\footnote{See the ``Valid Submission'' Section in \url{https://github.com/mlcommons/algorithmic-efficiency/blob/main/RULES.md\#valid-submissions}.}

\paragraph{Using shape and layer information of the model.}

Submissions may use the provided model parameter shape information if the resulting action can be applied to generic \wls.
Some examples of allowed uses include
(a) using the shape information of each layer to switch between a high-memory and a low-memory routine,
(b) using different update rules (\eg \adam and \sgd) for different layer types (\eg convolutional layer, \batchnorm layer, fully-connected layer), or
(c) leveraging the order of the layers to train layers in an organized fashion.

However, submissions may not use this same information to identify the specific \wl and use \wl-specific settings.
A clear case of using this information in a way that violates the spirit of the benchmark would be using the shapes of the model parameters as a workload ``fingerprint" and then loading or looking up the predetermined optimal \hps.
In general, any hard-coded behavior based on identifying a specific \wl is prohibited.
At the same time, it is entirely acceptable to take action based on basic modules such as the layer type. In other words, a submission may run different code whenever it encounters a model with convolutional layers, but it shouldn't need to specifically know that it is training a ResNet-50 on ImageNet.

\paragraph{Using expensive offline computations.}
Submissions may not circumvent the tuning rules by looking up the workload-specific results of offline computations that have been performed ahead of time.
This includes looking up optimal \hp settings for each specific \wl or even looking up (pre-trained) model parameters.

In contrast, it is perfectly fine to hard-code a single \hp setting, \eg, a default \hp setting, even when found using an expensive offline search because the \hp will need to perform well on all \wls simultaneously and thus could be expected to have some hope of generalizing to new workloads.
We also allow submissions based on learned \tas, which may include using a learned set of \hps.
In this case, we ask submitters to disclose information about the training set used to develop these learned \tas.

\subsection{Workloads}
\label{sec:rls_workloads}

A \emph{\wl} consists of a \dataset (including any preprocessing), model, and loss function, along with a target that is defined using some evaluation metric.
This evaluation metric can be identical to the \wl's loss function, or it could be a \wl-specific metric such as the \textit{word error rate} (WER) or the \textit{mean average precision} (mAP).
For example, training \resnetfifty on \imagenet using the \textit{cross-entropy} loss (CE) until a target error of \resnettesttarget on the test set has been reached, would constitute one \wl.

A diverse set of \wls that reflect multiple important application areas is necessary to assess the suitability of an algorithm as a general-purpose \ta (\Cref{sec:chal_workload_implications}).
We selected our list of workloads to cover several different tasks and data modalities, including image classification, machine translation, speech recognition, and other typical \ml tasks, focusing on today's practically relevant \wls.
The set of \wls will need to be extended in future iterations of the benchmark in order to remain relevant and to match advancements in the field. We intentionally restricted our current set of workloads to supervised learning tasks (although we could easily accommodate self-supervised tasks) and excluded reinforcement learning problems that might require fundamentally different methods and evaluation protocols.

\subsubsection{Fixed and Randomized Workloads}
\label{sec:public_held_out_workloads}

In service of identifying generally useful \tas, our benchmark includes two types of \wls: fixed \wls and randomized \wls.
\emph{Fixed \wls} are fully specified in the benchmark and completely known to the submitters.
\Cref{tab:workloads} provides an overview of the \nworkloads \emph{fixed} workloads used in the first iteration of this benchmark (\Cref{app:workload_details} contains the individual \wl details).
Additionally, we also provide \emph{randomized \wls} which are only defined as distributions over \wls.
Once all the benchmark submissions are frozen, we will sample specific instances from these randomized \wls that we call \emph{\heldout \wls}.
A submission's score is a function of its performance on the fixed \wls as well as these \heldout \wls.

\begin{table}[!htp]
	\centering
	\scriptsize
	\begin{tabularx}{0.98\textwidth}{@{}XllllS[table-format=2.6]S[table-format=2.7]S[table-format=6.0]@{}}
	\toprule
	&&&&& {Validation} & {Test} & {Maximum} \\
	\textbf{Task} & \textbf{Dataset} & \textbf{Model} & \textbf{Loss} & \textbf{Metric} & \textbf{Target} & \textbf{Target} & \textbf{\Runtime}\\
	\midrule
	Clickthrough rate prediction & \criteo & \dlrmsmall & CE & CE & 0.123649 & 0.126060 & 7703\\ \addlinespace
	MRI reconstruction & \fastmri & \unet & L1 & SSIM & 0.7344 & 0.741652 & 8859\\ \addlinespace
	Image & \imagenet & \resnetfifty & CE & ER & 0.22569 & 0.3440 & 63008 \\
	classification &  & \vit & CE & ER & 0.22691 & 0.3481 & 77520\\ \addlinespace
	Speech& \librispeech & \conformer & CTC & WER & 0.078477 & 0.046973  & 101780\\
	recognition &  & \deepspeech & CTC & WER & 0.1162 & 0.068093 & 92509\\  \addlinespace	
	Molecular property prediction & \ogbg & \gnn & CE & mAP & 0.28098 & 0.268729 & 18477\\ \addlinespace
	Translation & \wmt & \transformer & CE & BLEU & 30.8491 & 30.7219 & 48151 \\	
	\bottomrule
\end{tabularx}
 	\caption{\textbf{Summary of the \emph{fixed} \wls used in our benchmark.} The possible losses are the cross-entropy loss (CE), the mean absolute error (L1), and the Connectionist Temporal Classification loss (CTC). The evaluation metrics additionally include the structural similarity index measure (SSIM), the error rate (ER), the word error rate (WER), the mean average precision (mAP), and the bilingual evaluation understudy score (BLEU).}
	\label{tab:workloads}
\end{table}

Each randomized \wl introduces minor modifications to an associated fixed \emph{base} \wl.
These modifications include, for example, altering the data augmentation strategies or modifying aspects of the model architecture, such as the activation function or the number of layers.
Each randomized \wl defines a \emph{distribution} over workloads. For the first iteration of the benchmark, for convenience, we used particularly simple discrete distributions that sample one concrete workload variant out of a set of three possible, hand-designed variants of the base workload.
Only once all submissions are frozen do we select the specific instance of the \heldout \wl that will be used during scoring. Thus, although submitters know the three possible variants that might be sampled for each randomized workload, they will not know which of the three will be sampled during scoring.
Defining distributions instead of directly specifying instances ensures that while the entire process is public and transparent, neither the submitters nor the benchmark organizers know the specific \heldout \wls beforehand.

The random \heldout \wls function similarly to a \heldout test set, discouraging \tas that overfit to the fixed \wls.
The randomized \wls are intended to simulate novel-but-related \wls to ensure that the proposed \tas generalize to unknown problems. Although submitters can enumerate all possible \heldout workloads, by creating a larger set of possibilities than the fixed workloads alone, the randomized workloads should encourage more robust \tas.
Ideally, the randomized \wls should strike a balance between modifying a base \wl enough to generate a different \wl, but not so much that they produce something impossible to train.
If the modifications are too conservative, then the \heldout \wl will simply be a copy of the base \wl and not provide a generalization challenge that simulates novel \wls.
However, if the modifications are too drastic, then the \wl could lose its practical relevance.

Consequently, our randomized \wl distribution should only introduce ``natural changes'' that a practitioner might want to experiment with.
These include modifications such as changing the activation function, the type of normalization layer, or modifying the number and width of the layers. Any modification that actually improves results is \emph{prima facie} natural, as are modifications that occur in the literature or are obvious extensions of common practice. We want to avoid extremely contrived changes purely designed to make the training problem harder. That said, we are not above changes that might arise based on an insufficiently careful initial weight distribution, badly scaled parameters, or other changes of that nature.
At the end of the day, we want changes that may elucidate robustness properties of \tas that actually provide value for practitioners.

For the benchmark's first iteration, we manually designed three different \wl variants for each fixed \wl (see \Cref{tab:workload_variants}) from which to draw \heldout \wls.\footnote{In preliminary experiments, we briefly attempted to construct randomized \wl distributions with support over much larger sets of concrete \wls before deciding to take a simpler approach for the first iteration of the benchmark. See \Cref{app:ogbg_randomized} for information on these preliminary experiments.}
For each fixed \wl, one of these \wl variants will be randomly selected after submissions are frozen.
The randomized \wls use the same procedure as the fixed \wls to define validation and test targets (\Cref{sec:rls_timetoresults}). However, to save computational resources, we only tuned two \tas for the randomized \wls, \nadamw and the other best-performing \ta on the corresponding base \wl.

In total, a \ta submitted to the benchmark is evaluated on \nworkloads fixed and \nworkloads \heldout \wls. However, scoring uses the \heldout \wls only to penalize submissions that can't handle them, and reserves the fixed \wls for timing measurements. \Cref{sec:rls_heldout_scoring} describes the precise way \heldout workloads affect submissions scores, but, at a high level, we wanted to prioritize the fixed workloads for timing measurements since they are the most relevant variant and merely use the \heldout \wls as a deterrent for brittle submissions. Finally, \Cref{sec:exp_randomized_workloads_process} describes the experimental protocol for selecting possible \wl variants to serve as components of randomized \wls.

\subsection{Tuning}
\label{sec:rls_tuning}

Given our goal of evaluating generally applicable training methods, any \wl-specific adaptation (or tuning) should be an automated part of the algorithm.
In our benchmark, we provided two tuning \rulesets that govern how \hps may be tuned.
In the more permissive, external tuning \ruleset (\Cref{sec:rls_external_tuning}), each submission may define a list of \hps along with a search space to tune them over. To evaluate a submission using external tuning, we tune its \hps using \quasirandom search (with a modest, fixed budget of tuning trials) over the submission's search space. The runtime for a submission under this ruleset on a given workload is then the best-performing (fastest to reach the target) \hp setting for that workload across the tuning trials.
In contrast, the more restrictive, self-tuning \ruleset (\Cref{sec:rls_self_tuning}) does not allow any tuning outside of the timed operation of the submission itself on the benchmarking system. Instead,  the \ta must be \hp-free or, equivalently, it must automatically set any \hps it happens to define. 
The main difference between these two \rulesets is that in the self-tuning \ruleset every part of the algorithm---which may include any arbitrary internal tuning procedure---is performed ``on the clock.''
In contrast, the external tuning \ruleset allows for some parallelization of \hp tuning, where only the fastest \hp setting for each workload is used for scoring.

Both \rulesets are essentially independent competitions, where submissions only compete with methods adhering to the same \ruleset.
The two \rulesets cover two different practical scenarios: tuning on a single machine sequentially, and tuning in parallel in one shot across a modest number of machines. Any valid submission for the self-tuning \ruleset is valid under the external tuning \ruleset, but not vice-versa. Both tuning \rulesets share all other non-tuning benchmark rules, unless otherwise specified.
Below we describe the each version of the tuning rules in more detail.

\subsubsection{External Tuning}
\label{sec:rls_external_tuning}

For each \wl and each submission requiring external tuning, the \hps are tuned using $20$ tuning \emph{trials} drawn with \quasirandom search \citep{Bousquet2017} from the workload-agnostic search space specified in the submission.
In lieu of independent samples from their search space, submissions can instead supply a fixed list of $20$ \hp points that will be sampled without replacement.\footnote{Following \citet{Metz2020}, we also refer to this approach as a ``learned optimizer list'' or, abbreviated, as an \optlisttext.}
Using more than $20$ tuning trials would increase the computational burden of the benchmark. At the same time, we want the external tuning \ruleset to be sufficiently different from the self-tuning \ruleset, which effectively uses a single tuning trial.
To produce lower variance scores, the rules require repeating the tuning for five independent \emph{studies}, resulting in a total of $100$ trials.
For each of the five studies and for each workload, the \hp setting that reaches the validation target the fastest will be selected among the $20$ tuning trials.
For each workload, the score for a submission is the median of these five per-study training times and becomes an input into the overall benchmark score (\Cref{sec:rls_benchmark_score}).
While we use the time to reach the \emph{validation} set target for selecting the \hp point, we use the time to reach the \emph{test} set target for that \hp setting for scoring.
The five independent studies effectively simulate five hypothetical independent practitioners training the algorithm on a \wl using the same search space.
Using the median score allows us to assess what training speed the average practitioner can expect from this method.

\subsubsection{Self-Tuning}
\label{sec:rls_self_tuning}

Submissions to the self-tuning \ruleset are not allowed to have user-defined \hps and therefore receive no extra tuning.
Instead, the \tas in this \ruleset need to perform all necessary adaptations to the \wl autonomously.
This adaptation could, for example, come in the form of inner-loop tuning, \eg line search approaches, sequential outer-loop tuning, freeze-thaw methods, or algorithms that use the same \hps for all workloads, \eg \adam with its default parameters.
Any tuning effort will be part of the per-workload score and thus any tuning should save more time than it costs.

Compared to the external tuning \ruleset, there are no tuning trials but only a single run per study.
Once again, the median training time of the $5$ studies represents the per-workload score and is incorporated into the benchmark score.
Since we do not use any (external) tuning in this \ruleset, the time to reach the validation target is irrelevant and only the time to the test target is considered.
To account for the lack of external tuning, submissions have a longer per-workload time budget.
Compared to the external tuning \ruleset, all maximum allowed \runtimes are tripled, \ie $3\times\!$ the maximum \runtime reported in \Cref{tab:workloads}.

At this time, we do not anticipate self-tuning submissions will be competitive with externally tuned methods in terms of their per-\wl score.
However, closing the gap between fully automatic methods in the self-tuning \ruleset and more traditional externally tuned methods could drastically reduce the compute requirements of \dl.

\subsection{Scoring and Reporting Results}
\label{sec:rls_scoring}

So far, we have described how to evaluate a submission on a workload and produce a raw training time score. Suppose that for each submission $s$ and each \wl $w$, we determine a training time $t_{s,w} \in [0,\infty)$ using one of the rulesets described in \Cref{sec:rls_tuning}.
This training time is the wall-clock \runtime it took the submission to first reach the test target on this particular \wl. In the case of the external tuning rule set, we measure the time it takes to reach the test target \emph{only} for the \hp setting that achieved the validation target the fastest.
If a submission is unable to reach the target within the given \runtime budget, it will receive a score of $t_{s,w}\!=\!\infty$ for this particular \wl. These raw training time scores are already useful if we care about a single workload, \eg, by ranking submissions on the benchmark workload that is most similar to the real-world problem we are trying to solve. However, the raw training time scores are too detailed in most situations. 

Depending on our goal, we will need different ways of summarizing the raw training time data and converting it into scores, visualizations, and summary statistics. In general, in addition to providing guidance to practitioners that only care about a single workload, we need the results we report to help us:
\begin{enumerate}[itemsep=0mm]
    \item Qualitatively compare submissions across all workloads at once,
    \item Construct a leaderboard for submissions that aggregates across workloads, and
    \item Summarize year-over-year benchmark progress.
\end{enumerate}

\subsubsection{Aggregation Using Performance Profiles}
\label{sec:rls_performance_profiles}

A natural way to compare a given submission $s$ within a larger pool of competing submissions is to look at the fraction of workloads where $s$ trains the fastest (\ie, $t_{s,w}$ is the smallest). However, if there are workloads where a few strong submissions get nearly the same raw training time score, it might make sense to look at the fraction of workloads where $s$ is either the fastest or \emph{nearly} the fastest by being within, say, 1\% of the runtime of the best submission on that workload. Since our choice of 1\% is arbitrary, we also might ask the same question with different notions of whether a submission is sufficiently close to the fastest. Performance profiles \citep{Dolan2002} conveniently generalize this idea. Specifically, a performance profile is a plot of the fraction of workloads where a given method is within some ratio of the best per-workload training time.

Performance profiles are straightforward to compute given the raw training times, $t_{s,w}$, for a set of $k$ submissions $\mathcal{S}\!=\!\{s_1, s_2, \dots, s_k\}$ measured on a set of $n$ \wls $\mathcal{W}\!=\!\{w_1, w_2, \dots, w_n\}$. 
Specifically, for a submission $\bar{s}$ on a particular \wl $\bar{w}$ we define its performance ratio as:
\begin{align}\label{eq:performance_ratio}
    r_{\bar{s}, \bar{w}} = \frac{t_{\bar{s}, \bar{w}}}{\min_{s \in \mathcal{S}} t_{s, \bar{w}}} \,.
\end{align}
In other words, the performance ratio $r_{\bar{s}, \bar{w}}$ expresses how much time the submission $\bar{s}$ took on \wl $\bar{w}$ compared to the best submission on $\bar{w}$.
For example, if a submission takes twice as long on a particular \wl compared to the best submission, it will receive a performance ratio of $2$. By definition, $r_{\bar{s},\bar{w}} \ge 1$ for all submissions $\bar{s}$ and workloads $\bar{w}$.

The performance profile $\rho_{\bar{s}}(\tau)$ for a submission $\bar{s}$ is the probability that, on a random workload $w$ drawn uniformly from $\mathcal{W}$, $\bar{s}$ will have a performance ratio $r_{\bar{s}, w}$ of at most $\tau$. 
Specifically, for $\tau \in [1, \infty)$ the performance profile is defined as
$\bar{s}$:
\begin{align}\label{eq:performance_profile}
    \rho_{\bar{s}}(\tau) = \left( \frac{1}{n} \right) \cdot \left| \{ \bar{w} : r_{\bar{s}, \bar{w}} \leq \tau \} \right| \,.
\end{align}
Since $\rho_{\bar{s}}(\tau)$ expresses the fraction of \wls where a submission is less than $\tau$ away from the optimal submission, it is piecewise constant, monotonically non-decreasing with $\tau$, and  bounded between $0$ and $1$.
A perfect submission that always achieves the fastest training time on every \wl would have a performance profile that immediately jumps to $1$ at $\tau = 1$.
\Cref{fig:perf_profiles_core_baseline} in \Cref{sec:exp_baselines} shows performance profiles for our baseline submissions.

Performance profiles are a convenient way to summarize the overall performance of a submission relative to other submissions on the benchmark workloads. Since performance ratios are relative to the best submission on each workload, performance profiles are also inherently relative and will change as submissions are added or removed from the comparison set. Although performance profiles are great to qualitatively compare submissions, to construct a leaderboard we need a single scalar benchmark score.

\subsubsection{Integrating Performance Profiles for the Benchmark Score}
\label{sec:rls_benchmark_score}

To calculate the scalar benchmark score $B_{\bar{s}}$ of a submission $\bar{s}$, we integrate its performance profile up to a maximum ratio $r_{\text{max}}$:
\begin{align}\label{eq:benchmark_score}
    B_{\bar{s}} = \frac{1}{r_{\text{max}}-1} \int_{1}^{r_{\text{max}}} \rho_{\bar{s}}(\tau) \, d\tau \,.
\end{align}
Since we normalize by $r_{\text{max}}-1$, $B_{\bar{s}} \in [0,1]$, with higher benchmark scores being better.
A benchmark score of $B_{\bar{s}} = 1$ would indicate that the submission $\bar{s}$ was the fastest on every workload.
\Cref{tab:perf_scores_core_baselines} presents the benchmark scores of the baselines shown in \Cref{sec:exp_baselines}.

We set the upper integration limit to $r_{\text{max}} = 4$, which also serves as the right-hand limit of any performance profile plot.
This choice means that any submission that requires more than four times the \runtime of the fastest submission will not get any credit on this \wl and will be treated the same as a \ta that is unable to successfully reach the target within the maximum allowed \runtime budget. Although the exact integration bound is somewhat arbitrary, we want to encourage algorithms that are robust to different workloads, and in practice we are likely to rank the best-performing submissions similarly for most reasonable values of $r_{\text{max}}$. If there exists a specialized algorithm that is four times faster than any generic training algorithm on a particular problem, it seems likely that practitioners will prefer it even if it is only useful for a small number of problems.

\subsubsection{Using Held-Out \Wls in Scoring}
\label{sec:rls_heldout_scoring}

The benchmark score computation is based on a performance profile over only the fixed \wls. However, we penalize submissions that perform poorly on the \heldout \wls. If a submission does not perform well enough on a given \heldout \wl, then we score the submission on the corresponding fixed \wl as if that submission did not reach the fixed-\wl target. Specifically, for a submission to get credit for a finite training time on a particular fixed \wl, it must:
\begin{enumerate}[itemsep=0mm]
    \item Reach the validation and test target on the fixed \wl within the \runtime budget.
    \item Reach the validation and test target on the fixed \wl within $4\times\!$ of the fastest submission.
    \item Reach the validation and test target on the \heldout \wl corresponding to the fixed \wl within the maximum \runtime.
    \item Reach the validation and test target on the \heldout \wl corresponding to the fixed \wl within $4\times$ of the fastest submission.\footnote{To determine the fastest submission on a \heldout \wl, we only consider submissions that reached the target on the corresponding fixed \wl. This protects us against extremely fast submissions that only work on a specific \heldout \wl and are useless as general algorithms.}
\end{enumerate}
Only if all four requirements are met does the submission get credit for a finite score on that particular \wl. Otherwise, a submission will have an infinite training time on the fixed \wl.
This rule means that being unable to successfully train a \heldout \wl will disqualify a submission from getting credit for a good training time on the corresponding fixed \wl. We thus require submissions to be robust enough to not completely fail when faced with minor \wl variations. This ensures that we prioritize the fixed \wls for scoring since they are the most relevant version of that \wl in practice. However, we also protect the benchmark from egregious \wl-specific tuning and penalize brittle methods that break with slight modifications of the \wl.

\subsubsection{Measuring Year-Over-Year Benchmark Progress}
\label{sec:rls_yoy_score}

Scores derived from performance profiles make sense for relative comparisons within a set of submissions, but the benchmark scores $B_{s}$ of the winning submission between different iterations of the benchmark tell us nothing about how much faster training algorithms have become in an absolute sense. In order to measure year-over-year progress in reducing training time, we can use the geometric mean across workloads of the raw training times for the best submission. 

In general, we recommend that anyone who reports results on the benchmark should report raw training times in addition to any performance profiles and benchmark scores. The raw scores allow other researchers to include the submissions in their own performance profile comparisons or compute geometric means of any speedups if they want to measure progress through time.
Ideally, the raw times would be measured on the official competition hardware. However, if practical considerations require using a different system, additional baseline results from previous work should also be reported to facilitate comparisons. In the event that future iterations of the benchmark change the competition hardware, we can run the most crucial previous submissions on the new system.

\section{Target-Setting Experiments}
\label{sec:exp_target_setting}

As a time-to-result benchmark (see \Cref{sec:rls_timetoresults}), we need to set validation and test targets for all \wls.
To set the targets, we considered 4 \tas, \adamw \citep{Kingma2015,Loshchilov2019}, \nadamw \citep{Dozat2016,Loshchilov2019}, \nesterov \citep{Nesterov1983}, \heavyball \citep{Polyak1964}. 
We tuned all four target-setting \tas over relatively broad search spaces
(see \Cref{tab:target-setting-search-spaces} for exact search spaces).
For each algorithm, we sampled $200$ trials quasirandomly from its search space and selected the trial with the best validation metric.
\Cref{tab:target-setting-algorithm-performance} shows these resulting validation evaluation metric values for each algorithm, with the boldface values denoting the best for each workload.

As mentioned in \Cref{sec:rls_timetoresults}, each algorithm ran for $0.75 \times\!$ the maximum \runtime shown in \Cref{tab:workloads}. More precisely, since these four algorithms happen to have nearly identical step times, in these experiments we used a maximum number of steps as a proxy for runtime.\footnote{On all workloads, the step time differences between these four optimizers are negligible compared to the time required to complete one gradient calculation.} We determined the maximum number of steps that would fit within the given runtime budget, and used this same number of steps for every \ta.

\begin{table}[!htp]
	\centering
	\scriptsize
	\begin{tabular}{@{}lllll@{}}
\toprule
\textbf{\Hp}             & \textbf{\adamw}                       & \textbf{\nadamw}                      & \textbf{\heavyball}                   & \textbf{\nesterov}                    \\
\midrule
Base LR          & Log $[\SI{1e-5}, \SI{1e-1}]$ & Log $[\SI{1e-5}, \SI{1e-1}]$ & Log $[\SI{1e-3}, 10]$        & Log $[\SI{1e-3}, 10]$        \\
Weight decay    & Log $[\SI{1e-5}, 1]$         & Log $[\SI{1e-5}, 1]$         & Log $[\SI{1e-7}, \SI{1e-2}]$ & Log $[\SI{1e-7}, \SI{1e-2}]$ \\
1 - \betaone    & Log $[\SI{1e-3}, 1]$         & Log $[\SI{1e-3}, 1]$         & Log $[\SI{1e-3}, 1]$         & Log $[\SI{1e-3}, 1]$         \\
1 - \betatwo    & Log $[\SI{1e-3}, 1]$         & Log $[\SI{1e-3}, 1]$         & NA                           & NA                           \\
\addlinespace
Schedule        & \warmup                      & \warmup                      & \warmup                      & \warmup                      \\
                & + \cosinedecay               & + \cosinedecay               & + \lineardecay               & + \lineardecay               \\
                &                              &                              & + \constantschedule          & + \constantschedule          \\
\addlinespace
\Warmup         & $\{2\%, 5\%, 10\%\}$   & $\{2\%, 5\%, 10\%\}$   & 5\%                        & 5\%                        \\
Decay factor    & -                           & -                           & $\{\SI{1e-2}, \SI{1e-3}\}$   & $\{\SI{1e-2}, \SI{1e-3}\}$   \\
Decay steps     & -                           & -                           & Linear $[0.8, 1.0]$          & Linear $[0.8, 1.0]$          \\
\addlinespace
Dropout & $\{0.0, 0.1\}$          &  $\{0.0, 0.1\}$          &  $\{0.0, 0.1\}$          &  $\{0.0, 0.1\}$          \\
Aux.~dropout &  $\{0.0, 0.1\}$          &  $\{0.0, 0.1\}$          &  $\{0.0, 0.1\}$          &  $\{0.0, 0.1\}$          \\
Label smoothing &  $\{0.0, 0.1, 0.2\}$          &  $\{0.0, 0.1, 0.2\}$          &  $\{0.0, 0.1, 0.2\}$          &  $\{0.0, 0.1, 0.2\}$          \\ \bottomrule
\end{tabular}
 	\caption{\textbf{\Hp search spaces for the target-setting \tas.} Descriptions of the learning rate schedules can be found in \Cref{app:exp_details_chal_schedules}. The regularization \hps are tuned only for those workloads where they are applicable.}
	\label{tab:target-setting-search-spaces}
\end{table}
\begin{table}[!htp]
	\centering
	\begin{tabular}{@{} l *{4}{S[table-format=2.6]}@{}}\toprule
	\textbf{Workload} &\textbf{\criteo} &\textbf{\fastmri} &\textbf{\imagenet} &\textbf{\imagenet} \\
	&\textbf{\dlrmsmall} &\textbf{\unet} &\textbf{\resnetfifty} &\textbf{\vit} \\ \midrule
	{Metric} &{CE} \lib &{SSIM} \gib  &{Error Rate} \lib &{Error Rate} \lib\\\midrule
	\adamw &0.123675 & 0.734330	& 0.23034	& 0.22614 \\
	\nadamw & \B 0.123609 & 0.734523 & 0.22702 & \B 0.22534 \\
	\heavyball &0.125913 & 0.733828 & \B 0.22534 & 0.24486 \\
	\nesterov &0.126139 & \B 0.734645 & 0.22660 & 0.24318 \\
	\bottomrule
	\\
	\toprule
	\textbf{Workload} &\textbf{\librispeech} &\textbf{\librispeech} &\textbf{\ogbg} &\textbf{\wmt} \\
	&\textbf{\conformer} &\textbf{\deepspeech} &\textbf{\gnn} &\textbf{\transformer} \\ \midrule
	{Metric} &{WER} \lib &{WER} \lib &{mAP} \gib  &{BLEU} \gib \\\midrule
	\adamw &0.078327 &	0.114152 & 	0.277534 &	30.6939 \\
	\nadamw &\B 0.077790 & \B 0.113950 & 0.280012 & \B 30.8534 \\
	\heavyball & 0.132797 & 0.161977 & 0.276148 & 30.6431 \\
	\nesterov & 0.130823 & 0.171137 & \B 0.283124 & 30.1074 \\
	\bottomrule
\end{tabular}
 	\caption{\textbf{Performance of the target-setting \tas on each of the \wls.} Each entry contains the validation performance for the best tuning trial for a given algorithm and workload. We often see some ``reversion to the mean'' when rerunning this trial 20 times (cf. \Cref{tab:target-setting-top-20trials-valid}). The boldface values indicate the best result in each column.}
	\label{tab:target-setting-algorithm-performance}
\end{table}

After running the 200 \hp trials, for every \wl we took the best configuration (\ta and \hp settings) on the validation set (detailed in \Cref{tab:target-setting-opt-hparams}) and retrained it $20$ times with different random seeds. The final validation targets were the median values achieved over these $20$ repetitions, while the test targets were the worst-case test set performance achieved across those $10$ repetitions that hit the validation target.
The results of these $20$ repetitions, for each \wl, along with the associated \ta, \hp settings, and additional statistics are included in \Cref{tab:target-setting-top-20trials-valid,tab:target-setting-top-20trials-test} in \Cref{app:exp_details_exp_target_setting}. \Cref{tab:workloads} shows the final targets for all fixed \wls.
\begin{table}[!htp]
	\centering
	\scriptsize
	\setlength\tabcolsep{1.5pt} \begin{tabular}{@{}llllllllll@{}}\toprule
    &\textbf{\criteo} &\textbf{\fastmri} &\multicolumn{2}{c}{\textbf{\imagenet}} & \multicolumn{2}{c}{\textbf{\librispeech}} &\textbf{\ogbg} &\textbf{\wmt} \\
    \cmidrule(lr){4-5} \cmidrule(lr){6-7}
    &\textbf{\dlrmsmall} &\textbf{\unet} &\textbf{\resnetfifty} &\textbf{\vit} &\textbf{\conformer} &\textbf{\deepspeech} &\textbf{\gnn} &\textbf{\transformer} \\\midrule
    Algorithm &\nadamw &\nesterov &\heavyball &\nadamw &\nadamw &\nadamw &\nesterov &\nadamw & \\\addlinespace
    Base LR &0.003331 &0.028609 &4.131896 &0.000844 &0.001308 &0.004958 &2.491773 &0.001749 & \\
    Weight decay &0.003578 &0.000577 &5.67e-6 &0.081354 &0.163753 &0.114739 &1.29e-7 &0.081216 & \\
    \betaone &0.948 &0.981543 &0.927476 &0.889576 &0.973133 &0.863744 &0.944937 &0.932661 & \\
    \betatwo &0.998793 & & &0.99785 &0.998123 &0.629185 & &0.995516 & \\ \addlinespace
    Warmup &2\% &5\% &5\% &5\% &10\% &2\% &5\% &2\% & \\
    Decay factor & - &0.01 &0.001 &- &- &- &0.001 &- & \\
    Decay steps &- &0.984398 &0.900777 &- &- &- &0.861509 &- & \\ \addlinespace
    Dropout &0.1 &0 &0 &0 &0 &0 &0.1 &0.1 & \\
    Aux.~dropout &- &- &- &- &0 &0.1 &- &0.1 & \\
    Label&- &- &0.2 &0.2 &- &- &0 &0 & \\
    smoothing & & & & & & & & & \\
    \bottomrule
\end{tabular}
 	\caption{\textbf{Optimal \ta and \hp settings used for target setting on each \wl.} The reported configuration (\ta and \hp setting) achieved the best performance on the validation set.}
	\label{tab:target-setting-opt-hparams}
\end{table}

For the randomized \wls, we used nearly an identical procedure to set targets for each \wl variant. To save computational resources, we only tuned two training algorithms instead of four. For each workload variant, we used \nadamw and the other best-performing \ta on the corresponding base \wl.

\paragraph{Comparison to results in the literature}
The ideal target-setting procedure would produce targets that compare favorably to results reported in the literature for similar setups, although in many cases it will not be possible to find published results that are an exact match for our \wls (i.e.~training budget, evaluation metrics, data preparation, or model architecture might be different).
Therefore, this comparison is neither a comprehensive literature review, nor---by itself---a direct assessment of our target-setting procedure.
Instead, our goal is to provide context for a holistic evaluation of the relevance and competitiveness of our targets.
\begin{itemize}
    \item \textbf{\criteo} On this \wl, our target-setting procedure resulted in a validation target (binary cross entropy) loss of $0.123649$ and a test target loss of $0.126060$. In comparison, \citet{Sterbenz2017} reported a combined loss on our validation and test sets of $0.1250$ (our combined loss would be $0.1248545$), leading to a reported AUC score of $0.8002$. It is worth noting that \citet{Sterbenz2017} used a larger model and trained it for more than three times as long as we did. \citet{Nvidia2023} trained a model with the same dimensions as the one used in our workload for one epoch and reported AUC scores on our validation set between $0.802509$ and $0.802784$. However, it is important to note that their test \datasets used frequency thresholding, which we did not apply.
    
    \item \textbf{\fastmri} The target-setting procedure achieved an SSIM of $0.7344$ on our validation set and $0.741652$ on our test set. The paper introducing the \fastmri benchmark provided a \unet baseline with an SSIM score of $0.72$ on our combined validation and test set \citep[][Table 8, note that model selection was done using NMSE and not SSIM]{Zbontar2018}, compared to our slightly better combined score of $0.738$. In the 2019 \fastmri challenge \citep{Knoll2020}, the winning submission for the single-coil knee \dataset achieved an SSIM of $0.751$ on our combined validation and test set, using a custom i-RIM model \citep{Putzky2019}.
    
    \item \textbf{\imagenet} For the \resnetfifty \wl, our target error rate on the validation set of $22.57\%$ improves over the performance reported in the original \resnet paper by \citet{He2016} (their most similar setup reaches $24.7\%$). More recent studies have achieved lower error rates by using variants of the original \resnet architecture or by employing improved training recipes. \citet[][\textsc{ResNet-RS-50} in Table 7]{Bello2021} report a validation error of $21.2\%$, and \citet[][A2 in Table 1]{Wightman2021} report an error of $20.2\%$ when training for roughly $2.67 \times\!$ the training budget of our \wl. With a slightly shorter training time than our \wl, \citet[][A3 in Table 1]{Wightman2021} achieved an error rate of $21.9\%$ using a lower training resolution, RandAugment, CutMix, and Mixup compared to our workload.\\
    For the \vit \wl, our target-setting procedure yielded a validation error rate of $22.69\%$. The paper introducing Vision Transformers reported an error rate of $22.09\%$ for a slightly larger model trained exclusively on \imagenet \citep[][\textsc{ViT-B/16} in Table 5]{Dosovitskiy2021}. We can compare our results with a training budget of roughly $112$ epochs, to \citet{Beyer2022}, which reported error rates of $23.5\%$ and $21.5\%$ for the same \textsc{ViT-S/16} model used in our \wl for a training budget of $90$ and $150$ epoch respectively (note that they used our validation set as a test set and trained only on $99\%$ of the training data to use the remaining $1\%$ as their validation set).
    
    \item \textbf{\librispeech} Our target-setting procedure for the \librispeech \dataset resulted in a validation word error rate (WER) of $0.1162$ for the \deepspeech \wl and $0.078477$ for the \conformer \wl. For easier comparison with the literature, we can consider the WERs on our test set (the \texttt{test\_clean} split), which are $0.067976$ and $0.046696$ for \deepspeech and \conformer, respectively. The original paper introducing the \librispeech \dataset reported a baseline with a WER of $0.0551$ \citep[][Table 3]{Panayotov2015}, although the model uses a completely different approach than our \wls. Subsequently, \citet[][Table 4]{Amodei2015} reported an improvement to $0.0515$ with the \deepspeech 2 model. The \deepspeech 2 model uses a much larger training set beyond just \librispeech and uses beam search decoding, unlike our \deepspeech \wl. \citet[][Table 2, Conformer(S) without LM]{Gulati2020} introduced the \conformer architecture and reported $0.027$ test WER. However, \citet{Gulati2020} used a much larger \conformer model, used beam search decoding instead of greedy decoding, trained for more steps, and included a few other more minor model differences.
    
    \item \textbf{\ogbg} The target-setting procedure for the \ogbg \wl resulted in a mean average precision (mAP) on the validation set of $0.28098$. When introducing the \textsc{Open Graph Benchmark} (\textsc{OGB}), \citet{Hu2020} also provided six baselines for the \textsc{ogbg-molpcba} \dataset used in our \wl. The best baseline reached a validation mAP of $0.2798$. Subsequent approaches were able to achieve higher results with different models and training techniques, such as a $0.3012$ mAP by \citet{Wang2022}, or $0.3252$ mAP using additional training data \citep{Wang2021}.
    
    \item \textbf{\wmt} For our \wmt workload, we follow the setup adopted by the \flax \citep{flax2020github} WMT example. \wmt \datasets for different years aim to establish benchmarks for different problems in the machine translation domain, e.g. low-resource translation, domain \& style of translation, or long-sequence translation. Consequently, over the past decade, the neural machine translation literature has used different combinations of \wmt \datasets depending on the desired language pair, recency of the data, amount of data, and underlying translation problem. Since our goal is to provide a training algorithms benchmark, we decided to stay close to a high quality open source example (in this case the \flax \wmt example). Specifically, the models are trained on ``train'' split of \textsc{WMT2017} translation \dataset \citep[]{Bojar2017} for German to English (De $\!\rightarrow\!$ En), use the ``dev'' set from the \textsc{WMT2014} dataset ~\citep[]{Bojar2014} and use \texttt{newstest2014} as the test set. Our models achieve a BLEU score of 30.72 on \texttt{newstest2014}. \citet[Table 10]{gao-etal-2022-bi} also evaluate their models on \texttt{newstest2014} De $\rightarrow$ En, and achieve higher BLEU scores (in the range 33.60 -- 35.15). However, they pre-train the models in a bidirectional manner (De $\!\rightarrow\!$ En and En $\!\rightarrow\!$ De) before fine-tuning on the De $\!\rightarrow\!$ En translation direction. Additionally, \citet[Table 6]{ma2023mega} evaluate their models on \texttt{newstest2014} De $\!\rightarrow\!$ En direction and report BLEU score in the range 31.33 -- 32.35, although they made architectural changes to the attention mechanism.  
\end{itemize}

\section{Randomized Workloads Experiments}
\label{sec:exp_randomized_workloads}

The primary goal of our randomized \wls is to help deter brittle submissions that only work on the original---and relatively standard---fixed \wls, and instead
encourage more robust \tas that also perform well on novel \dl \wls.
An additional concern is that by using only a small set of \nworkloads (fixed) \wls, submissions in the external tuning \ruleset (which are permitted to sample $20$ different \hp points per \wl) could effectively perform \wl-specific tuning.
This risk is amplified since we permit submissions with a fixed \hp list (\ie, \optlisttext approaches) instead of search spaces.
For instance, a submission could use a \hp list of the best \hp configurations for each fixed \wl.
Such approaches could potentially lead to generally useful \tas and thus should be allowed.
However, unless it generalizes to novel \wls the expensive offline computation it would require cannot be justified.
Without \heldout \wls, the benchmark would not test whether submissions that tune over a list of configurations that perform well on the fixed \wls can generalize to \wls outside of the \nworkloads fixed \wls.
To encourage robust submissions and avoid vitiating the limits on \wl-specific tuning, we manually created three variations of each fixed, base workload.
The three variants together form a randomized \wl that will be used to sample one specific concrete variant to use as a \heldout \wl during scoring. Since \heldout \wl sampling occurs after all submission code has been frozen, submissions may need to perform well on any of the possible variants.

\subsection{Desiderata for Workload Variants}
\label{sec:exp_randomized_workloads_desiderata}

Although any \wl change could potentially pose challenges for some hypothetical submissions, an ideal \wl variant intended for use as part of a randomized \wl would have the following properties.

\paragraph{Representative of real workloads}
The best \wl variants would be as representative as possible of changes that occur in practice. Extremely contrived changes to the base \wl do not help the community develop robust and general \tas. The most natural changes are ones that already occur in the wild (\eg, architectural changes described in the literature). However, when judging how natural a particular modification is for a \wl, it is important to distinguish between changes that affect the optimization dynamics and more surface level architectural changes. Even if we would never expect a particular change to be applied in a real application, if it results in optimization dynamics that \emph{do} occur in the wild, it might still be worth considering. The ``attention temperature'' change described below is arguably a bit contrived in terms of model architecture, but it reproduces a type of training instability that occurs in practice, especially as Transformer models become larger, and therefore \tas that handle it well could be useful \citep{gilmer2023intriguing, kim2021lipschitz, Dehghani2023}.

\paragraph{Trainable in the original \runtime budget}
We want \wl variants that are trainable. Furthermore, they should reach reasonable validation and test evaluation metric values \emph{within the \runtime budget of the base \wl}. Specifically, there should exist a \ta with some \hp setting that achieves good results. It is very easy to make changes that completely ruin a \wl and produce a model incapable of performing well on its task no matter how it is trained, but these models are not useful. Exhibiting a configuration of the \hps that achieves a good validation error on the original task proves that it is possible for the \wl variant to yield a useful trained model. 

\paragraph{Distinct from the base \wl (and other variants)}
A \wl variant only adds information to the benchmark if it is distinct enough from the original fixed \wl it is based on. Ideally, the entire pool of variants of a given \wl would be mutually distinct from each other---and the base \wl---while each presenting an interesting new challenge for submissions. Although there are many ways to define distinctiveness, we tried to create variants that meaningfully changed the optimal \hps of the \tas used during target setting. In this way, we hoped to encourage submissions that are easier to tune than current popular techniques that use straightforward search spaces, as represented by our target-setting algorithms.

Achieving all the desiderata listed above simultaneously, on command, is unfortunately difficult. Nonetheless, \wl variants that meet these requirements definitely exist, especially if we make a few practical concessions, so giving up does not seem appropriate either. 
Originally, we hoped to specify randomized \wls via distributions with support over a very large number of variants by randomizing different pieces of the \wl definition. However, without a better scientific understanding of the effects of various \wl modifications, constructing such distributions was simply too onerous (see \Cref{app:ogbg_randomized} for details on some of our attempts), which is why we elected to manually design a small number of specific, concrete \wl variants instead. Even this goal was more challenging than we expected, but \emph{not} because it is impossible to construct modifications that require re-tuning the \hps. Instead, the challenge in designing interesting \wl variants comes from the conjunction of requirements that need to be achieved simultaneously.

\subsection{Creating and Testing Workload Variants}
\label{sec:exp_randomized_workloads_process}

While creating \wl variants, we explored various modifications to the base \wls. The following families of modifications ended up being used in the benchmark variants:

\begin{itemize}
    \item \textbf{Activation function:} Most base \wls employed \relu as the activation function and we explored alternative activation functions such as \gelu, \silu, or \Tanh.
    
    \item \textbf{\preln vs \postln:} For Transformer-based models, the base \wl was usually the \prelnfull (\preln) \citep{Xiong2020} version. We changed these to \postlnfull (\postln, see \Cref{fig:pre_post_ln}). 

    \item \textbf{Attention temperature:} For the \wmt \transformer, we modified the attention layers to compute $\text{Softmax}\left(\frac{c X W^Q (X W^K)^\top}{\sqrt{D/H}}\right)$ where $c$ is a constant scalar denoting the attention temperature. The default self-attention implementation sets $c=1$. In order to artificially induce instabilities similar to those faced by larger versions of these models, we set $c=1.6$.
    
    \item \textbf{Initialization scales:} For the \dlrm model, changing the scale of the initial weights of the embedding layer resulted in a variant. For the \resnet model, changing the initial \batchnorm layer scale weights resulted in a \wl variant.
    
    \item \textbf{Normalization layer:} We changed the type of normalization layer employed in the model. Common changes included interchanging \batchnorm with \layernorm, as well as \instancenorm with \layernorm. 

    \item \textbf{Width, depth, and channels:} We explored changing model width, depth, and number of channels, as applicable. 

    \item \textbf{Input pipeline:} On \librispeech, we found changing \specaug strength to be an effective strategy. 

    \item \textbf{Residual connection structure and scaling:} For the \dlrm model, we created a variant with additional residual connections. For \deepspeech, we removed residual connections from the model. 

    \item \textbf{Pooling layer type:} Changing the pooling layer type from \textit{global average pooling} to \textit{max average pooling} resulted in a variant for the \vit \wl.
\end{itemize}

We used a relatively permissive protocol to decide whether a candidate change to a workload produced an \textit{acceptable} variant, since strictly achieving the strongest versions of all of our desiderata is quite difficult. Specifically, we tested whether the optimal \hp setting for one of the more robust, standard algorithms (\nadamw) were sufficiently different between the candidate variant and its base \wl.
We chose \nadamw as the training algorithm to measure variant distinctiveness because, in our experiments with our search spaces, its optimal \hps tended to transfer much better than, for instance, \heavyball or \nesterov. \nadamw also happened to be the algorithm that set targets on the most \wls ($5/8$).

In addition to testing distinctiveness, we also rejected variants that degraded validation performance too much. As a rule of thumb, we accepted workload variants for which we could achieve a performance within 10\% of the original validation performance (in the original target setting budget). 
More precisely, given a base workload $w$ and a candidate workload variant $w'$, we re-ran the exact same collection of 200 \hp settings that were used for target-setting (Table~\ref{tab:target-setting-search-spaces}) for \nadamw on the candidate variant. Let $H = \{h_1, \ldots h_{200}\}$ denote this set of points in \hp space. Furthermore, let $H(w) = \{h_i(w)\}_{i=1}^{200}$, where $h_i(w)$ denotes the validation evaluation metric of \hp setting $h_i$ on workload $w$. 
Let $h^*(w)$ and $h^*(w')$ denote the best \hp setting for $w$ and $w'$, respectively. Additionally, let $\mathrm{rank}(h, H(w))$ denote the rank of the \hp setting $h$ in the set $H(w)$ when the elements are ordered according to validation performance on workload $w$ (lower ranks indicate better performance). In general, we sought variants $w'$ for which the following quantity was as large as possible: 
$$\min\left\{\mathrm{rank}\left(h^*(w), H(w')\right), \mathrm{rank}\left(h^*(w'), H(w)\right)\right\}.$$ 
In other words, when both ranks are greater than $c$, then the optimal \hp setting on $w$ performs worse on $w'$ than at least $c$ other \hp points, and the optimal \hp setting on $w'$ performs worse on $w$ than at least $c$ other \hps as well. Note, if $w$ and $w'$ share the same optimal \hp, then these ranks will be $0$. In the rest of this section, to simplify notation we write $\mathrm{rank}(w\rightarrow w')$ and $\mathrm{rank}(w'\!\rightarrow\!w)$ instead of
$\mathrm{rank}(h^*(w), H(w'))$ and $\mathrm{rank}(h^*(w'), H(w))$, respectively.

Ultimately, we used human judgement instead of strict thresholds to decide if $\mathrm{rank}(w\!\rightarrow\!w')$ and  $\mathrm{rank}(w'\!\rightarrow\!w)$ were large enough for a given candidate variant and if it achieved a tolerable validation performance. In some cases (for the \resnet and \conformer models) we did not find variants that substantially changed the rank of the optimal \hp point. In these cases, we fell back to creating variants that had different optimal base learning rates in simple one-dimensional learning rate sweeps. See \Cref{app:workload_details} for additional details and results from our variant testing protocol.

Our procedure for testing and rejecting candidate variants is not guaranteed to produce a set of variants that achieve all of our desiderata. Although, by design, all our variants are at least somewhat representative of real \wls and are trainable in the original budget (perhaps with some degradation in performance), we potentially sacrificed distinctiveness. We made no attempt to ensure that different variants of the same base \wl were \emph{mutually} distinct from each other in terms of optimal \hps, although we did not repeat the same modifications within a set. Furthermore, our operational definition of distinctiveness depends on the particular set of \hp points we used and our choice of \ta. Finally, we did not always produce variants that made very drastic changes to the ranks of the best \hp points.

\subsection{Workload Variants of the Benchmark}
\label{sec:exp_randomized_workloads_selected_variants}

\Cref{tab:workload_variants} contains a brief description of the changes that produced the three variants of every fixed \wl. See \Cref{app:workload_details} for additional details about the changes and variants, as well as workload-specific results of the variant testing protocol (described above). Although it was hard to predict how candidate variants would perform in our tests, in hindsight some patterns emerged. For example, after switching the activation function from \relu to \gelu in the \conformer model, higher learning rates performed better (and in many cases the resulting model performance itself was improved).

We found producing variants which successfully changed the optimal \nadamw \hps to be surprisingly difficult. This speaks to the robustness of \nadamw to \wl variations, a desirable property that perhaps helps explain the success of preconditioned \tas. In contrast, the optimal learning rate for momentum methods is highly sensitive to seemingly trivial variations in the workload (\eg the \wideresnet stride change experiment in \Cref{sec:chal_model_changes}). Note, this does not mean that the performance of \nadamw was completely robust to changes in the \wls. In fact, it was quite easy to design a \wl variant which was more difficult for \nadamw to optimize. However, for many such variants the optimal \nadamw \hps did not change. 

There is at least one known case for which we expect the optimal \nadamw \hps to change: when the batch size changes \citep{Nado2021}. However, this was not a valid option for us because the submission hardware is fixed and the submitter is allowed to choose the batch size. We were initially hopeful that other methods for increasing stochasticity in optimization problem would decrease the optimal \nadamw learning rate. Indeed, although it isn't part of the \wl definition, changing the value of dropout can have this effect. However, in many cases when we tried to achieve similar effects by increasing the severity of data augmentation or adding label noise, we did not see a significant change in the optimal learning rate, although we did manage to hurt overall performance. 

Another surprising failure was playing with residual scales. For example we tried modifying the traditional residual connection of $x\!+\!F(x)$ to $\alpha x\!+\!(1\!-\!\alpha) F(x)$ for $\alpha\!\in\![0,1]$. By sweeping $\alpha$ between 0 and 1, we can interpolate between a model with no residual connections ($\alpha\!=\!0$) to the default setting $\alpha\!=\!0.5$ to a model which ignores all intermediate blocks $\alpha\!=\!1.0$---varying $\alpha$ between these 3 extremes allows us to explore interesting variations of residual connections. Indeed we found $\alpha$ to be a very important parameter for overall performance (somewhat surprisingly $\alpha\!=\!0.5$ was not always the optimal value), however the optimal learning rate for any given $\alpha$ did not seem to change. We also tried $\alpha x\!+\!F(x)$ which did not work either, though did improve performance for $\alpha\!=\!2$ and $\alpha\!=\!4$.

Finally we would like to highlight some of our variants that actually outperform the base workload in validation error. The changes which we saw leading to improvement in models were changing from \relu to \gelu (\resnet), changing from \relu to \silu (\resnet and \gnn), introducing GLU (\vit) and \emph{removing} residual connections (\deepspeech).

\begin{table}[!htp]\centering
	\scriptsize
	\begin{tabularx}{0.99\textwidth}{@{}llXS[table-format=2.6]S[table-format=2.7]@{}}
\toprule
Base &  & Variant & \multicolumn{1}{l}{Validation} & \multicolumn{1}{l}{Test} \\
\textbf{Workload} & \textbf{Variant} & \textbf{Description} & \textbf{Target} & \textbf{Target} \\ \midrule
\textbf{\criteo} \\ \quad \dlrmsmall & \textsc{Embed Init Scale} & Changes initialization scale of the embedding layer from $1/\sqrt{\text{vocab size}}$ to $1$ & 0.124286 & 0.126725 \\
& \textsc{LayerNorm} & Adds \layernorm to the network & 0.123744 & 0.126161 \\
& \textsc{Residual} & Adds residual connections to the network & 0.124027 & 0.126470 \\
\addlinespace\textbf{\fastmri} \\ \quad \unet & \textsc{Channels \& Pooling} & Increases number of channels and decreases number of pool layers & 0.734376 & 0.741547 \\
& \textsc{TanH} & Replaces all \leakyrelu activations with \Tanh & 0.729633 & 0.736727 \\
& \textsc{LayerNorm} & Replaces \instancenorm with \layernorm with learnable parameters & \B 0.734861 & \B 0.741982 \\
\addlinespace\textbf{\imagenet} \\ \quad \resnetfifty & \textsc{SiLU} & Replaces all \relu activations with \silu & \B 0.220090 & \B 0.342600 \\
& \textsc{GELU} & Replaces all \relu activations with \gelu & \B 0.220770 & \B 0.340200 \\
& \textsc{BN Init Scale} & Increases the scale of the initialization of \batchnorm scale variables & 0.234740 & 0.357700 \\ 
\addlinespace\quad \vit & \postln & Uses \postln instead of \preln & 0.246880 & 0.371400 \\
& \textsc{MAP} & Changes pooling type from global to max average & 0.228860 & 0.347700 \\
& \textsc{GLU} & Include GLU in the \textsc{MLPBlock} & \B 0.223300 & \B 0.345500 \\
\addlinespace\textbf{\librispeech} \\ \quad \conformer & \textsc{GELU} & Replaces all \relu activations with \gelu & 0.077958 & 0.047643 \\
& \textsc{LayerNorm Change} & The LayerNorm before the final readout layer was removed & 0.085371 & 0.053096 \\
& \textsc{Attention Temp} & Increases attention temp from $1$ to $1.6$ & 0.082665 & 0.050168 \\
\addlinespace\quad \deepspeech & \textsc{TanH} & Replaces all \relu activations with \Tanh & 0.133449 & 0.079810 \\
& \textsc{No Residual} & Removes residual connections.  & \B 0.105042 & \B 0.060388 \\
& \textsc{Norm \& SpecAugment} &  Removes decoder \layernorm layer \& replaces \batchnorm with \layernorm. Changes \specaug specifications. & 0.131553 & 0.082442 \\
\addlinespace\textbf{\ogbg} \\ \quad \gnn & \textsc{GELU} & Replaces all \relu activations with \gelu & 0.277710  & 0.262926  \\
& \textsc{SiLU} & Replaces all \relu activations with \silu & \B 0.282178 & \B 0.272144 \\
& \textsc{Altered Layers} & Adds a hidden layer,  decreases latent dimension. Reduces the number of message passing steps and changes \layernorm to \batchnorm. & 0.269446 & 0.253051 \\
\addlinespace\textbf{\wmt} \\ \quad \transformer & \postln & Uses \postln instead of \preln & 30.2003 & 29.8982 \\
& \textsc{Attention Temp} & Increases attention temperature from $1$ to $4.0$ & 30.0756 & 29.8094 \\
& \textsc{GLU \& \Tanh} & Uses GLUs in the MLP blocks and replaces all \relu activations to \Tanh & 30.0002 & 29.8139 \\ \bottomrule
\end{tabularx}
 	\caption{\textbf{Overview of \wl variants used for randomized \wls.} See \Cref{app:workload_details} for additional details of each \wl variant. Targets that are better than the corresponding base workload target are in bold. Unsurprisingly, the variants where the validation targets improved were the same as the ones where the test targets improved.} 
	\label{tab:workload_variants}
\end{table}

\section{Baseline Submissions}
\label{sec:exp_baselines}

We constructed baseline submissions using eight different \ta families: \adamw \citep{Kingma2015,Loshchilov2019}, \nadamw \cite{Dozat2016,Loshchilov2019}, \nesterov \citep{Nesterov1983}, \heavyball \citep{Polyak1964}, \lamb \citep{You2019}, \adafactor \citep{Shazeer2018}, \samadam \citep{Foret2021} and \distshampoo \citep{gupta2018,anil2020}. For each \ta family, we designed one or more search spaces to create valid submissions for the external tuning \ruleset. For \adamw, \nadamw, \nesterov and \heavyball, we compared submissions derived from search spaces that fixed the relevant first moment parameter ($\beta_1$) to a default value with submissions that tuned it, denoted by names ending with \textsc{fixed} $\beta_1$ or \textsc{tuned} $\beta_1$, respectively. Additionally, for these four \ta families, we also constructed baselines that use a list of $20$ specific \hp configurations to sample from, without replacement. We denote baselines making use of these kinds of search spaces by names ending with \optlisttext (\eg \adamw \optlisttext).

\subsection{Baseline Creation Procedure}
\label{sec:exp_baselines_creation}

We used the results of the target-setting experiments to guide the creation of search spaces for baseline submissions. Since \lamb, \adafactor, \samadam and \distshampoo were not used in the target-setting procedure, we collected similar tuning data for them by sampling $200$ trials from a broad search space for each algorithm. We used search spaces with the same ranges as in \Cref{tab:target-setting-search-spaces}, except defined over the analogous \hps in \lamb, \adafactor, \samadam and \distshampoo (\ie the \adamw $1 - \beta_1$ range was used as the $1 - \beta_1$ range for the analogous first moment parameters in the other algorithms). \samadam has an additional $\rho$ \hp that we search over using the recommended discrete search space of $\{0.01, 0.02, 0.05, 0.1, 0.2, 0.5\}$.
Since the target setting procedure violates the submission rules by using too many tuning trials, we needed to tighten the search spaces used in the $200$-trial searches to produce strong baselines for the $20$ trial budget allowed in the external tuning \ruleset. We used the following recipe to produce narrower search spaces for all baseline submissions (see \Cref{tab:baseline-search-spaces} for the resulting exact search spaces).
\begin{enumerate}[itemsep=0mm]
    \item \textbf{Learning rate and weight decay:} For a given algorithm, for every fixed workload, we first found the \hp setting with the best validation error among the 200 trials, yielding a set of at most \nworkloads \hp points (up to one per fixed \wl).
    We set the boundaries for the search spaces for the (base) learning rate and weight decay to include the maximum and minimum values observed in this set, plus a little extra on each side.
    While this procedure led to a small search space for preconditioned algorithms such as \adamw and \nadamw, the search spaces for \heavyball and \nesterov were prima facie too wide to allow for efficient sampling, so we narrowed them by excluding outliers when building the set of per-workload \hp points. In particular, we removed \hp points from consideration when we could find an alternative \hp point in the target-setting experiments with very similar performance, but within our narrowed-down search space. 

    \item \textbf{$\beta_1$ and $\beta_2$:} As mentioned above, some baseline submissions searched over $\beta_1$ (\eg \adamw \textsc{tuned} $\beta_1$, \heavyball \textsc{tuned} $\beta_1$, etc.) and others fixed it to a default value (\eg \adamw \textsc{fixed} $\beta_1$).\footnote{For the purposes of this paper, we call the momentum parameter $\beta_1$ in \heavyball and \nesterov although it obviously is a different \hp in each of those algorithms (as well as in \adamw).}
    When tuning $\beta_1$, we followed the same procedure as described for the learning rate to find the extreme values among the best per-\wl \hp settings, and set the search space to be within these (rounded) bounds. As is common, we reparameterized the search space to explore $1 - \beta_1$ in logspace. For $\beta_2$ (where applicable) we used its default value of $0.999$. 
    
    \item \textbf{Learning rate schedule parameters:} We used a fixed $5\%$ linear \warmup from $0.0$ for all runs.\footnote{Although using a learning rate of zero on the first step is somewhat perverse, it simplifies the logic and is common in many codebases.} Baselines using the Linear Decay + Constant schedule family always searched the decay factor in the set $\{0.01, 0.001\}$ (as in target-setting). In target-setting, both of these values appeared in top-performing \hp settings in roughly equal proportion. For the decay steps factor parameter, we used a fixed value of $0.9$; we saw the best trials in the $200$-trial searches concentrated near that value and preliminary experiments seemed to suggest tuning it did not matter very much. 
    
    \item \textbf{Regularization Parameters:} For dropout, we noticed that, in the best trials, dropout and aux.~dropout parameters tended to be the same (either both $0.0$ or both $0.1$).
    Therefore, we tied these parameters together in our baselines and searched them in the discrete set $\{0.0, 0.1\}$. For label smoothing, we noticed that whenever the workload allows it, the best-performing trials used either $0.1$ or $0.2$, therefore we searched within that set whenever the workload supported label smoothing. 
    
    \item \textbf{\optlisttext baselines:} To build \optlisttext baselines, we ranked the $200$ \hp configurations from the broad searches independently on each of the \nworkloads fixed \wls. Then we greedily grew a set of $20$ \hp configurations by cycling through \wls in an arbitrary, round-robin order and adding the top  configurations on each \wl that wasn't already in the set. 
 \end{enumerate}

\begin{table}[!htp]
	\centering
	\scriptsize
	
\begin{tabular}{@{}lllll@{}}
\toprule
\textbf{\Hp}                 & \textbf{\adamw}              & \textbf{\nadamw}             & \textbf{\heavyball}          & \textbf{\nesterov}           \\ \midrule
Base LR                      & Log $[\SI{1e-4}, \SI{1e-2}]$ & Log $[\SI{1e-4}, \SI{1e-2}]$ & Log $[\SI{1e-1}, 10]$        & Log $[\SI{1e-1}, 10]$        \\
Weight decay                 & Log $[\SI{5e-3}, 1]$         & Log $[\SI{5e-3}, 1]$         & Log $[\SI{1e-7}, \SI{1e-5}]$ & Log $[\SI{1e-7}, \SI{1e-5}]$ \\
1 - \betaone [Default]       & $0.1$                        & $0.1$                        & $0.1$                        & $0.1$                        \\
1 - \betaone [Tuned]         & Log $[\SI{2e-2}, 0.5]$       & Log $[\SI{4e-3}, 0.1]$       & Log $[\SI{5e-3}, 0.3]$       & Log $[\SI{5e-3}, 0.3]$       \\
1 - \betatwo                 & $0.999$                      & $0.999$                      & NA                           & NA                           \\
\addlinespace
Schedule        & \warmup                      & \warmup                      & \warmup                      & \warmup                      \\
                             & + \cosinedecay               & + \cosinedecay               & + \lineardecay               & + \lineardecay               \\
\addlinespace
\Warmup         & $5\%$                      & $5\%$                      & $5\%$                      & $5\%$                      \\
Decay factor                 & NA                           & NA                           & $\{\SI{1e-2}, \SI{1e-3}\}$   & $\{\SI{1e-2}, \SI{1e-3}\}$   \\
Decay steps                  & NA                           & NA                           & $0.9$                        & $0.9$                        \\
\addlinespace
Label smoothing & $\{0.1, 0.2\}$               & $\{0.1, 0.2\}$               & $\{0.1, 0.2\}$               & $\{0.1, 0.2\}$               \\
Dropout (Tied)               & $\{0.0, 0.1\}$               & $\{0.0, 0.1\}$               & $\{0.0, 0.1\}$               & $\{0.0, 0.1\}$               \\
\bottomrule
\\
\toprule
\textbf{\Hp}                 & \textbf{\lamb}               & \textbf{\adafactor}          & \textbf{\samadam}            & \textbf{\distshampoo}       \\ \midrule
Base LR                      & Log $[\SI{1e-4}, \SI{1e-2}]$ & Log $[\SI{1e-4}, \SI{1e-2}]$ & Log $[\SI{1e-4}, \SI{1e-2}]$ & Log $[\SI{1e-4}, \SI{1e-2}]$ \\
Weight decay                 & Log $[\SI{1e-3}, 1]$         & Log $[\SI{1e-3}, 1]$         & Log $[\SI{1e-2}, 0.2]$       & Log $[\SI{5e-3}, 1]$         \\
1 - \betaone [Default]       & $0.1$                        & $0.1$                        & $0.1$                        & $0.1$                        \\
1 - \betaone [Tuned]         & Log $[\SI{2e-2}, 0.5]$       & Log $[\SI{1e-2}, 0.45]$      & Log $[\SI{5e-2}, 0.43]$      & Log $[\SI{1e-2}, 0.15]$      \\
1 - \betatwo                 & $0.999$                      & $0.999$                      & $0.999$                      & $0.999$                      \\
$\rho$                       & NA                           & NA                           & $\{0.01, 0.02, 0.05\}$       & NA                           \\
\addlinespace
Schedule        & \warmup                      & \warmup                      & \warmup                      & \warmup                      \\
                             & + \cosinedecay               & + \cosinedecay               & + \cosinedecay               & + \cosinedecay               \\
\addlinespace
\Warmup         & $5\%$                      & $5\%$                      & $5\%$                      & $5\%$                      \\
Decay factor                 & NA                           & NA                           & NA                           & NA                           \\
Decay steps                  & NA                           & NA                           & NA                           & NA                           \\
\addlinespace
Label smoothing & $\{0.1, 0.2\}$               & $\{0.1, 0.2\}$               & $\{0.1, 0.2\}$               & $\{0.1, 0.2\}$               \\
Dropout (Tied)               & $\{0.0, 0.1\}$               & $\{0.0, 0.1\}$               & $\{0.0, 0.1\}$               & $\{0.0, 0.1\}$               \\ \bottomrule
\end{tabular}
 	\caption{\textbf{\Hp search space for the baseline submissions.} Descriptions of the learning rate schedules can be found in \Cref{app:exp_details_chal_schedules}. The regularization \hps are tuned only for those workloads where they are applicable. Dropout and aux.~dropout are set to the same value.}
	\label{tab:baseline-search-spaces}
\end{table}

\subsection{Baseline Timing}
\label{sec:exp_baselines_timing}

In order to run more extensive experiments, we used Google TPUs \citep{jouppi2020tpu} whenever we could, thus deviating from the official, \benchmarkinghardware benchmark system. To estimate training time for our baselines on the official benchmark system, we timed each \ta on the true competition system, on each \wl, for a reduced number of training steps. Our timing measurements have some noise, possibly exacerbated by non-local disks. However, variations smaller than the time between off-the-clock evaluations (typically about 1\% of total allowed \runtime) are unlikely to have a large effect on the benchmark results. 
To reduce the time needed on the official benchmark system, we ran each algorithm (except \distshampoo) for 20\% of the allowed number of steps, extrapolated to the full number of steps, and averaged across two runs on two different machines (see  \Cref{tab:baselines-runtimes} in the Appendix for the timing results). As of the time of writing, we were not able to run \distshampoo on all \wls on the official benchmark system due to various memory and configuration issues that we hope to correct in future versions of this work. Therefore, we omit \distshampoo when presenting results with respect to runtime. Since we \emph{were} able to run \distshampoo on TPUs, we have included \distshampoo in any results that don't require \runtime measurements on the official benchmark system.

All the baseline algorithms should take roughly the same time per step, except for \samadam and \distshampoo. In practice, \samadam and \distshampoo typically require $1.5\times$--$2\times$ the time per step (for \distshampoo this measurement is on TPU), depending on the workload. Note however that \distshampoo can amortize these costs by increasing the batch size which we leave for future work as it requires careful work for fair comparisons. Although all of our main, target-setting \tas had timing results in line with our expectations, specifically on the competition hardware using GPUs, our implementation of \adafactor took around 10-20\% more time per step than \adamw on multiple \wls. In contrast, on TPU, \adafactor step times were much closer to \adamw, although they still seemed slightly slower. Due to these unexpectedly slower step times and the max \runtime limit, our \adafactor baseline misses a couple of targets that it would be able to hit if it ran for the same number of steps as \adamw. Although \adafactor was not one of the target-setting algorithms and we did not find any obvious issues in the implementation, we would like to investigate this slowdown more in the future.

On the competition hardware, using our current implementations, three out of the eight \wls, namely  \criteo \dlrmsmall, \fastmri \unet and \ogbg \gnn, are data-pipeline-bound when training with any of the target-setting algorithms. In other words, reading, preprocessing, and preparing batches of training data takes much longer than computing gradients and weight updates. The exact bottleneck varies across \wls, and isn't necessarily transferring data from disk. For example, on \ogbg \gnn creating and padding batches of graphs is a bottleneck. On data-pipeline-bound \wls, techniques such as data echoing \citep{choi2019_data_echoing} can accelerate training. Additionally, on such \wls, optimizers that perform more work per batch and might otherwise increase the time-per-step (e.g. \samadam) can reclaim idle accelerator time. Although data-pipeline-bound \wls exist in the wild and are a legitimate part of a representative benchmark suite,\footnote{Given the end of Moore's law, we should expect accelerator improvements to continue to outpace improvements in general purpose processors. For that reason, the problem of bottlenecks upstream of the part of the training pipeline that runs on the accelerator(s) is not likely to go away.} as researchers iterate on a specific \wl, they tend to make changes that reduce, or eliminate, idle accelerator time. For example, they might optimize the input pipeline code, run multiple independent copies of the data pipeline in parallel, or switch to a larger, more computationally intensive model. For these reasons, input pipeline bottlenecks are best viewed not as an immutable property of the problem, but instead as a consequence of a particular model size, amount of engineering resources, and amount of computational resources. Although a small number are fine, we generally view data-pipeline-bound \wls as undesirable for a \tas benchmark because such \wls tend to favor expensive training algorithms that wouldn't be as useful when we have the resources to remove the bottleneck. In the future, we would like to optimize the data pipelines for \criteo \dlrmsmall, \fastmri \unet and \ogbg \gnn in hopes of removing bottlenecks and increasing GPU utilization.

\subsection{Baseline Results}
\label{sec:exp_baseline_results_core_baselines}

\begin{figure}
	\centering
	\begin{subfigure}[b]{\textwidth}
		\centering
		\includegraphics[width=\textwidth]{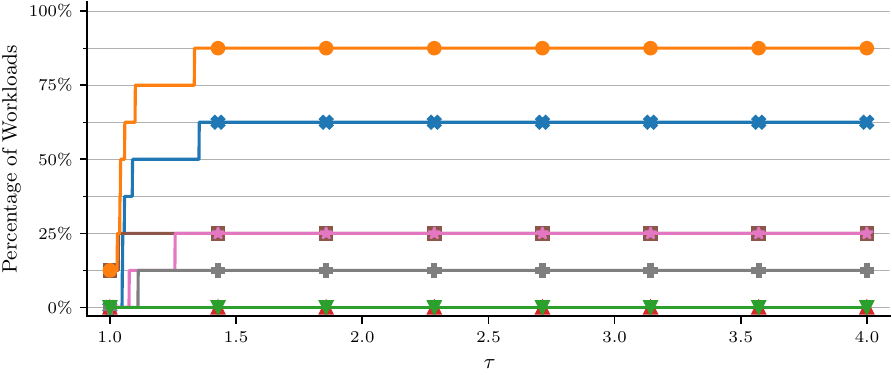}
		\caption{Performance profiles when measuring \textbf{runtime} to target}
    	\label{fig:perf_profiles_core_baseline_time}
	\end{subfigure}
	\par\bigskip \begin{subfigure}[b]{\textwidth}
		\centering
		\includegraphics[width=\textwidth]{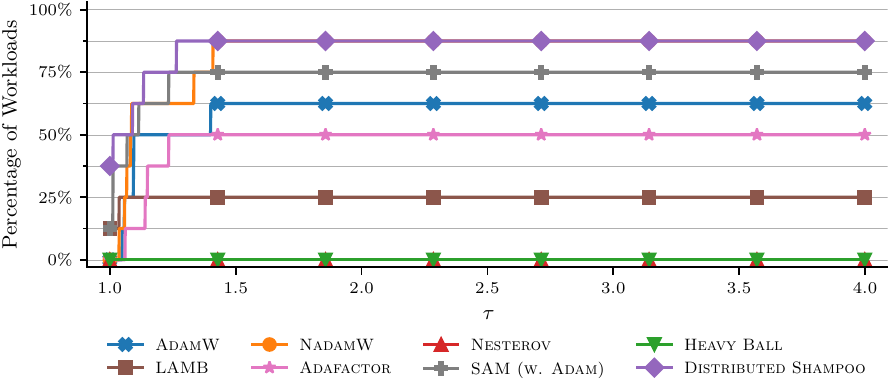}
		\caption{Performance profiles when measuring \textbf{steps} to target}
    	\label{fig:perf_profiles_core_baseline_steps}
	\end{subfigure}
	\caption{\textbf{Performance profiles of our baseline submissions.} Each line in these plots is the performance profile of a single baseline submission.
	A step in any line occurring at a value $\tau$ indicates that for one additional \wl the corresponding submission achieves the target within a $\tau$ factor of the \runtime of the best submission. For example in (b), \nadamw (\colorline{SNSorange}) has a bump just before $\tau = 1.5$. This indicates that on one additional \wl, \nadamw requires a bit less than $1.5\times$ the number of steps as the fastest submission to reach the target on this \wl.
	If a submission does not reach the target on one or more \wls, its performance profile will not reach $100\%$ at the very right of the plot. A flat line at $0\%$ indicates that for all \wls the submission either did not hit the target at all or did so in time/steps that is at least $4$ times worse than the time/steps to target of the best submission. As mentioned earlier, we computed the performance profile for the entire set of baselines presented in the paper together and have removed the profiles for other baselines from this figure for readability.}
	\label{fig:perf_profiles_core_baseline}
\end{figure}

\Cref{fig:perf_profiles_core_baseline} shows performance profiles for \adamw, \nadamw, \nesterov, \heavyball, \lamb, \adafactor, \samadam and \distshampoo using the search spaces described in \Cref{tab:baseline-search-spaces} with the $\beta_1$ parameter tuned. 
\Cref{fig:perf_profiles_core_baseline_time} shows the performance profile with respect to \runtime,  \Cref{fig:perf_profiles_core_baseline_steps} shows the same performance profile with respect to steps, and \Cref{tab:perf_scores_core_baselines} shows the benchmark scores measuring the normalized area under the performance profile curves. Unlike the official scoring procedure we will use for real submissions, baselines in \Cref{fig:perf_profiles_core_baseline} ignore the held-out \wl criterion described in \Cref{sec:rls_heldout_scoring}, which will depend on the specific held-out \wls sampled during official scoring, after submission code has been frozen. 
Since the performance profile (and thus the benchmark score) for any individual baseline depends upon the entire set of baselines included in the performance profile, we computed performance profiles and benchmark scores for the complete set of baselines presented across all sections, together (e.g. including those presented in \Cref{sec:exp_baseline_results_search spaces}) to ensure consistency. Throughout the paper, when comparing a subset of baselines, we might elide some baselines from a figure to improve readability, but the curves are always computed based on the full set.
Please see \Cref{tab:baselines-benchmark-scores} in the appendix for the complete set of benchmark scores for all baselines considered in the paper.

Although our benchmark is designed around measuring time-to-result, looking at performance profiles based on steps as well can sometimes be illuminating. Given that all target-setting algorithms take roughly the same time per step, the relative ranking of the baselines derived from those algorithms does not change when measuring steps instead of \runtime. Since \samadam and \distshampoo typically require $1.5\times$--$2\times$ the time per step as the target-setting algorithms, in order to create performance profiles based on steps as well as \runtime, we ran \samadam and \distshampoo for the same number of steps as the other baselines and used learning rate schedules based on this step budget. These step-budget-optimized schedules likely decay the learning rate too slowly to be ideal when measured with respect to \runtime. With faster implementations and \runtime-tuned learning rate schedules, it might be possible to make these \tas competitive in \runtime as well as in number of steps. That said, \samadam has a very large deficit to make up, and might still struggle to be competitive. Regardless, we hope proponents of these algorithms create submissions that achieve the best possible \runtime results on our benchmark.

We can observe several interesting results from \Cref{fig:perf_profiles_core_baseline}, \Cref{tab:perf_scores_core_baselines} and the workload-specific breakdown in \Cref{tab:baseline-performance-time} and  \Cref{tab:baseline-performance-steps} presented in the appendix.
\begin{itemize}
    \item There is no single baseline \ta that hits the targets on every workload, but on every \wl there exists at least one (and usually more than one) baseline that hits the target. \distshampoo and \nadamw both reach the target on 7 out of 8 workloads. However, as described above, \distshampoo has an unfair budget advantage, so \nadamw hits the targets on the largest number of \wls among baselines adhering to the strict runtime budget. \distshampoo misses the target on \criteo \dlrmsmall, whereas \nadam misses the target on \imagenet \resnetfifty.
    
    \item With our search spaces, \nadamw performs significantly better than \adamw both in terms of \runtime and steps. Despite this impressive result, it seems to be much less popular than \adamw. 

    \item We found that \adafactor performs somewhat worse than \adamw in terms of steps to target. \adafactor is able to hit targets on \sfrac{4}{8} workload. \adamw hits the same targets and additionally the target on \criteo \dlrmsmall.
    
    \item \lamb hits the target on only 2 of our workloads, \librispeech \deepspeech and \wmt \transformer and thus does not do well in terms of the benchmark score. However, for \wmt \transformer in particular, \lamb is the fastest algorithm to the target (by a 7\% margin).
    
    \item The non-preconditioned baselines, \heavyball and \nesterov, are not able to hit the target on any workload. Two factors explain this surprisingly poor result: (1) how we constructed search spaces for these \tas, and (2) the way our benchmark rules require consistent performance and force \tas to account for the \wl-specific tuning they require to achieve good results. Both \heavyball and \nesterov have some trials which hit the target on \imagenetresnet and \fastmri, but they are not frequent enough to show up in the median over the five studies. Indeed, search spaces for these algorithms that are better tailored to the tuning budget can hit more targets, which we show later in \Cref{sec:exp_baseline_results_search spaces}. However, as the target-setting experiments suggest, outperforming the pre-conditioned baselines would be a tall order.
\end{itemize}

\begin{table}[!htp]
	\centering
	\begin{tabular}{lS[table-format=2.6]S[table-format=2.6]}
	\toprule 
	\textbf{Submission} & \multicolumn{2}{c}{\textbf{Benchmark Score}} \\
	\cmidrule(lr){2-3}
	 & {Runtime} & {Steps} \\
	\midrule
	\adamw      & 0.600141 &  0.596116 \\
	  	\nadamw      & \B 0.849960 & 0.830414 \\
	   	\nesterov      & 0.0 & 0.0 \\
	   	\heavyball      & 0.0 & 0.0 \\
        \lamb &0.248619 & 0.248494 \\
        \adafactor & 0.236111 &0.475760 \\
        \samadam & 0.120368 &0.731717 \\
        \distshampoo & {-} & \B 0.854210 \\
	\bottomrule
\end{tabular}
 	\caption{\textbf{The benchmark scores for our baseline submissions.} These are the integrated performance profiles shown in \Cref{fig:perf_profiles_core_baseline_time} (\runtime) and \Cref{fig:perf_profiles_core_baseline_steps} (steps).}
	\label{tab:perf_scores_core_baselines}
\end{table}

\subsubsection{Baseline Results Comparing Search Spaces}
\label{sec:exp_baseline_results_search spaces}

\begin{figure}
	\centering
	\begin{subfigure}[b]{\textwidth}
		\centering
		\includegraphics[width=\textwidth]{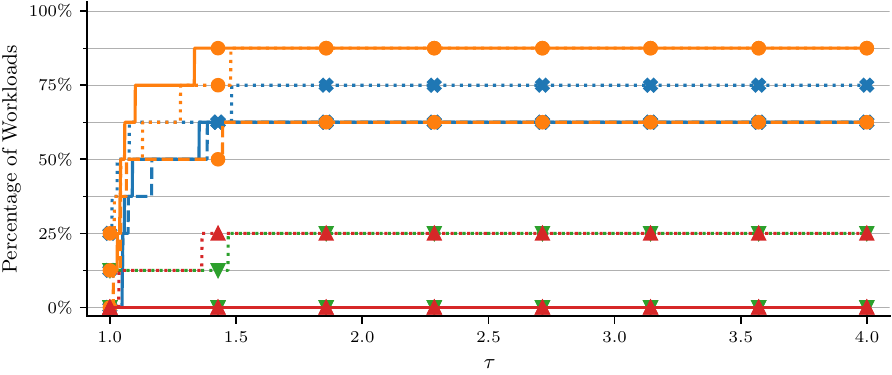}
		\caption{Performance profiles when measuring \textbf{runtime} to target}
	    \label{fig:perf_profiles_search_space_baseline_time}
	\end{subfigure}
	\par\bigskip \begin{subfigure}[b]{\textwidth}
		\centering
		\includegraphics[width=\textwidth]{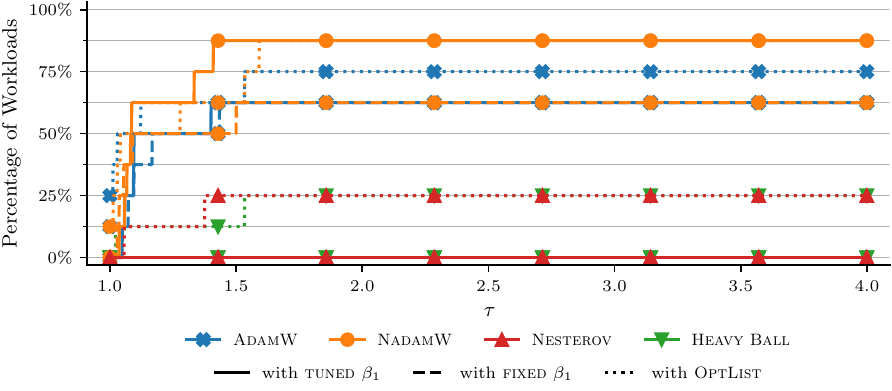}
		\caption{Performance profiles when measuring \textbf{steps} to target}
	    \label{fig:perf_profiles_search_space_baseline_steps}
	\end{subfigure}
	\caption{\textbf{Performance profiles of our baseline submissions with different search spaces.} Each line in these plots is the performance profile of a single baseline submission (analogously to \Cref{fig:perf_profiles_core_baseline}). As mentioned earlier we computed the performance profile for the entire set of baselines presented in the paper together and have suppressed the profiles for other baselines from this figure for readability.}
	\label{fig:perf_profiles_search_space_baselines}
\end{figure}

In order to reveal more about the role of the \hp search space, we compared baselines using the target-setting algorithms (\adamw, \nadamw, \nesterov, \heavyball) with different search spaces.
For each algorithm, we compared two search spaces (specified in \Cref{tab:baseline-search-spaces}): one tuning the $\beta_1$ parameter, and the other keeping the $\beta_1$ parameter fixed to $0.9$.
Additionally, for each algorithm, we also prepared an \textsc{OptList} baseline, as described above, that samples without replacement from a list of \hp configurations that performed well on at least one \wl during target-setting.
Tables \ref{tab:opt-list-adam}, \ref{tab:opt-list-nadam}, \ref{tab:opt-list-nesterov} and \ref{tab:opt-list-heavyball} in the Appendix contain the complete list of configurations used in the \textsc{OptList} baselines. 
\Cref{fig:perf_profiles_search_space_baselines} shows the resulting performance profiles for this comparison and \Cref{tab:perf_scores_search_space_baselines} shows the corresponding benchmark scores given by the (normalized) area under the curve. 
The raw data used to prepare these plots can be found in Tables \ref{tab:baseline-performance-time} and \ref{tab:baseline-performance-steps} in the Appendix.

As expected, the search space plays a large role in the results.
\begin{itemize}
    \item Although overall \nadamw with a search space that tunes $\beta_1$ performed the best among these \tas in terms of runtime (and steps), \nadamw performed  roughly the same as \adamw when the $\beta_1$ parameter isn't tuned (fixed to the default value of 0.9). On the other hand, for \adamw, tuning $\beta_1$ shows little advantage (at least at the limited number of 20 tuning trials).
    
    \item The \optlisttext baselines for \nadamw and \adamw seem to be roughly the same in terms of performance as compared to the best box polytope search spaces for those \tas. On the other hand, for \nesterov and \heavyball, the \optlisttext baselines perform substantially better (and are the only ones to hit any target).
    This result is likely a consequence of our search space construction procedure and the diversity of \hp values that worked well for \nesterov and \heavyball. In other words, our procedure for constructing box polytope search spaces needed to cover much wider \hp ranges for \nesterov and \heavyball that also contain a large volume of bad points, making it harder to get good results when sampling from these types of search spaces, for these algorithms.
\end{itemize}

\begin{table}[!htp]
	\centering
	\begin{tabular}{llS[table-format=2.6]S[table-format=2.6]}
	\toprule 
	\textbf{Submission} & \textbf{Version} & \multicolumn{2}{c}{\textbf{Benchmark Score}}\\
	\cmidrule(lr){3-4}
	 & & {Runtime} & {Steps} \\
	\midrule
	\adamw    & \textsc{tuned} $\beta_1$ & 0.600141 &  0.596116 \\
	& \textsc{fixed} $\beta_1$ & 0.596985 &  0.593047 \\
	& \textsc{opt-list}   & 0.725260 &  0.721035 \\
	  	\nadamw   & \textsc{tuned} $\beta_1$ &  \B 0.849960 & \B 0.830414 \\
	  & \textsc{fixed} $\beta_1$ & 0.599691 & 0.595478 \\
	& \textsc{opt-list}  & 0.835602 &  0.813194 \\
	   	\nesterov   & \textsc{tuned} $\beta_1$  & 0.0 & 0.0 \\
	   		  & \textsc{fixed} $\beta_1$ & 0.0 &  0.0 \\
	& \textsc{opt-list}  & 0.233373 &  0.232048\\
	   	\heavyball  &  \textsc{tuned} $\beta_1$  & 0.0 & 0.0 \\
  &  \textsc{fixed} $\beta_1$  & 0.0 & 0.0 \\
  & \textsc{opt-list}  & 0.230504 &  0.226860 \\
	\bottomrule
\end{tabular}
 	\caption{\textbf{The benchmark scores for our baseline submissions.} These are the integrated performance profiles shown in \Cref{fig:perf_profiles_search_space_baseline_time} (\runtime) and \Cref{fig:perf_profiles_search_space_baseline_steps} (steps).}
	\label{tab:perf_scores_search_space_baselines}
\end{table}

\section{Discussion}
\label{sec:discussion}

\subsection{Target Setting}
\label{sec:disc_target_setting}

Fundamentally, our target setting procedure is a \emph{generic} recipe for training neural networks that achieves (arguably) a near state-of-the-art  result on all of our fixed \wls and could, in principle, be applied to any workload. The procedure assumes the model architecture, dataset, loss function, and other \wl details are given and also assumes we have a known limit for the maximum \runtime allowed for the training program. Furthermore, it assumes we can afford to sample hundreds of trials from the search spaces we designed for the various \tas. Although it depends on a lot of assumptions, it \emph{does} give us an automatic process for getting good results on our fixed \wls. 

Naturally, we might ask ourselves how good this procedure is compared to other alternatives. 
Unfortunately, our procedure is the only one of its kind we are aware of in the deep learning literature. Although \citet{tuningplaybookgithub} make an attempt to systematize the applied deep learning workflow, the process they describe is still far from a repeatable, mechanical procedure we can use for setting targets. Instead of generic procedures for training and tuning neural networks, the literature is filled with particular configurations that worked well on specific problems, only occasionally accompanied by the tuning search spaces that helped discover them. These \wl-specific training recipes typically do not generalize to new \wls. This gap in the literature makes it difficult to know how competitive the validation and test targets we set with our procedure actually are, or how realistic they are to reach for completely general, \wl-agnostic \tas. 

Two steps in the deep learning workflow seem to be especially neglected by the literature. First, we are not aware of algorithms for determining how long to train, or papers that tackle the concomitant philosophical issues surrounding how best to frame this choice. Second, existing tools for automatic \hp tuning still require search spaces as input, i.e., we need to somehow determine which \hps to tune and what set of values they should be allowed to take on. Although the blackbox optimization literature has explored methods that automatically grow or shrink search spaces, the deep learning literature contains shockingly little research on how to design the best search spaces for use with existing tools and arbitrary budgets. Indeed, this profound gap in the literature was part of our motivation for forcing submissions in our own benchmark to grapple with tuning challenges directly.

Our own target-setting procedure is far from the final word on these issues. In an effort to keep the protocol \wl-agnostic and generic, we ended up with broad search spaces that required a large number of tuning trials, even though preliminary experiments showed that some \hps only needed to be tuned on a subset of the \wls. For example, supporting and tuning dropout was not always necessary or helpful. Furthermore, our search spaces (\Cref{tab:target-setting-search-spaces}), despite being relatively broad, had to be constructed through manual trial and error. In several cases, we had to refer to previous published results on similar \wls to determine that our initial search spaces needed to be adjusted. We hope that the community scrutinizes our target setting procedure and comes up with better alternatives. Although we tried our best to create an objective protocol that minimized human judgement, we were unable to remove it entirely. Even if full automation is out of reach, we hope the community takes the problem of \hp search space construction, specifically in deep learning, much more seriously.

Stepping back, our target setting procedure essentially simulates a competition between \tas to get the best validation performance within a fixed \runtime budget. This protocol is a mirror image of our time-to-result benchmark. One possibility might be to formalize this relationship and alternate competitions to improve targets given fixed \runtime budgets, and \runtimes given fixed targets. Even if we don't take things quite so far, we could revise targets in future benchmark iterations by incorporating winning submissions into the target-setting protocol. The targets we selected using our procedure represent what is possible with currently popular \tas applied in a \wl-agnostic manner. These targets are generally \emph{not} going to be the state-of-the-art error rates on the tasks our \wls cover.

\subsection{Randomized Workloads}
\label{sec:disc_randomized_workloads}

As described in \Cref{sec:exp_randomized_workloads}, although we found realistic \wl variants that simultaneously changed the best-performing \nadamw \hps in our search spaces and reached reasonable error rates in the original \runtime budget, discovering such variants was surprisingly difficult and far too labor intensive.
Our difficulties highlight the desperate need for new research to predict when \hp settings that perform well on one \wl will transfer to a related \wl and, more generally, predict the effect of various \wl modifications on the optimal \hps. While there has been some recent work in this vein, such as \citet{Yang2021} which focuses on model size, we are still far from being able to reliably predict when good \hps will transfer. A theory of \hp transfer for all of the most common \hps (and most popular \tas) would not just make it easier to produce interesting \wl variants, but would likely go a long way towards achieving the underlying purpose of our randomized \wls: encouraging \tas that are robust to the exact details of the workload and are easy to tune. Even if such a theory did not immediately lead to more robust and convenient algorithms, it would let us save tuning effort by extrapolating to new experiments from related tuning results.

Our randomized \wls built from \wl variants highlight an important set of trade-offs in the design of our benchmark. The more \wls in the benchmark, the more expensive running a submission through the benchmark becomes, especially in the more expensive external tuning \ruleset.
Both fixed \wls and randomized \wls contribute to this cost, but because we sample a single variant from each randomized \wl to construct the held-out \wls, the fixed and randomized \wls currently contribute roughly equally to the total cost.
Similarly, the more tuning trials we allow in the external tuning ruleset, the more expensive the benchmark becomes. Since larger tuning budgets let submissions adapt more to specific \wls, they also increase the importance of using a large enough and diverse enough set of \wls, potentially through randomized \wls. However, without a better understanding of when existing \tas need to be re-tuned, it is very hard to know when a new \wl or \wl variant is worth the increase in cost. Even worse, a variant that provides a useful challenge for one \ta might be completely redundant for another.

\subsection{Baselines}
\label{sec:disc_baselines}

Our baseline results show that our per-workload targets are achievable, reveal clear gaps between different \tas, and suggest performance on the benchmark is far from saturated. On each \wl, at least one baseline was able to reach the target, but no baseline reached the targets on all \wls simultaneously. \nadamw with our tuned $\beta_1$ search space reached the target on seven out of the eight fixed \wls, and constitutes a provisional state of the art on our benchmark. The performance profiles that form the basis of the benchmark scores showed stark differences between the tested methods.

Our baseline results also showed that the current practice of trying to compare abstract update rules with free parameters, divorced from the tuning protocol that would instantiate them, is doomed to produce perpetually conflicting results. \Hp tuning search spaces (and tuning protocols more generally) play a crucial role in the effectiveness of a \ta. By picking specific search spaces, even when they are reasonable \emph{a priori}, we could claim our results showed that \adamw is better than \nadamw, or vice versa, when in reality we are only showing that \nadamw \emph{with a particular tuning protocol} is better than \adamw with some other specific tuning protocol. We need to view the tuning protocol as an inseparable part of the \ta if we are ever going to determine a meaningful notion of the state-of-the-art \ta. Similarly, although \heavyball or \nesterov was the target-setting algorithm on a combined three out of the eight \wls, our \heavyball or \nesterov with box polytope search spaces failed to reach the target reliably on any \wls. This is not a general judgment about \heavyball or \nesterov as update rules. Instead, this result once again highlights the importance of the tuning protocol and, if anything, might indicate that these algorithms require more \wl-specific tuning effort. New training algorithms with \hps that must be tuned should ideally provide (budget-dependent) tuning procedures, or at a minimum, guidance on how to tune at a variety of budgets.

\subsection{Benchmark Limitations}
\label{sec:disc_limitations}

The benchmark we presented in this work, like any benchmark, has a variety of limitations.
These limitations fall into several, broad categories. First, the benchmark has limited coverage of possible submissions. In other words, the benchmark rules and software end up prohibiting, or effectively prohibiting, some potentially interesting submissions that we would have preferred to allow, in principle. Second, there are limitations that could affect whether benchmark scores truly measure what we intend them to measure (i.e. benchmark ``validity'' in the parlance of psychometrics). Third, there are limitations of the scope of the benchmark that could affect its relevance to the actual practice of deep learning. Finally, there are limitations that affect the accessibility of the benchmark, primarily in terms of how easy and affordable it is for researchers to score new submissions.

\paragraph{Coverage of the space of potentially interesting submissions}
Benchmark submissions must adhere to a specific \ta API and interoperate with \wl implementations in either \jax or \pytorch. Although this restriction is an essential design choice intended to, among other things, help isolate the effects of the \ta, it does make it so some potentially interesting submissions cannot be supported under the current rules. For example, although submissions can employ arbitrary \hp tuning procedures while being timed, submissions adhering to the external tuning \ruleset and making use of parallel tuning resources must use random search. Instead of being allowed to employ more sophisticated black-box optimization algorithms, they must only rely on their control over the tuning search space.  
Although we could potentially relax this limitation in future versions, allowing submissions complete control over how they utilize the tuning resources would require   much more complicated orchestration code and APIs.

Similarly, although not against the rules, the API does not provide a way to implement model parallelism since it isn't situated in the submission code, as we define it, even if we could imagine a hypothetical exotic \ta that depended on it somehow.
As another example, submissions are not allowed to access arbitrary information about the current \wl. Instead, they can only obtain basic layer metadata and dataset information. This restriction precludes submissions based on optimizers, such as \kfac \citep{martens2015}, that require detailed architectural information. Although \kfac itself is non-trivial to apply to new model architectures even when detailed architectural information is available, hypothetical generic \tas that required such information, but worked for any arbitrary neural network, would likely also face difficulties. 

Submissions are also constrained from a software implementation standpoint. They must interoperate with either \jax or \pytorch \wl implementations, prohibiting other frameworks and software stacks. Although in theory there is nothing stopping us from porting our \wls to additional frameworks, in practice it is far too costly in terms of engineering resources. 

Ultimately, our initial rules and API err on the side of caution to make sure we isolate the effects of the \ta. Moving forward, we intend to keep a close eye on the types of algorithmic modifications that are of interest to the community, but are not possible within our rules or API, and solicit suggestions on ways they could be accommodated. 

\paragraph{Experimental protocol}
Although we designed our benchmark to prioritize producing convincing measurements, there are still ways the experimental protocol could be strengthened. Specifically in the external tuning \ruleset, the ratio between the number of \wls (eight fixed along with eight more \heldout \wls) and the number of tuning trials allowed per study (twenty) could allow for too much \wl-specific tuning and thus risk rewarding submissions that overfit to the particular suite of \wls. Although the randomized \wls we draw \heldout \wls from are designed to mitigate this issue, the \wl variants did not fully achieve all of our desiderata. Although all three variants of a given base \wl will require different \hp settings than the base \wl, they might not require mutually distinct \hp settings from each other, risking a set of variants that doesn't challenge submissions as much as initial appearances might suggest. Additionally, the variants of different \emph{base} \wls often repeat similar changes. For example, we generated variants of multiple base \wls by making activation function changes.
Moving beyond \wl overfitting concerns, neural network training is a noisy process, especially when \hp tuning using random search is involved. Our strategy of repeating measurements with different random seeds during scoring can help, but the best-performing optimization \hps for the most popular optimizers are often near the ``edge of stability'' \citep{cohen2022adaptive}, so some submissions can get unlucky and have training diverge more than is typical when we only have a small number of repetitions.

\paragraph{Scope and relationship to current practice}

Our choices of what \wls and conditions to study in our benchmark necessarily constrain the set of situations the results will be most relevant for.
Even restricting our attention to supervised and self-supervised learning, we cannot hope to cover every practically relevant data modality, let alone every practically relevant dataset and model. For example, our benchmark currently does not contain any \wls for object detection from point cloud data, weather prediction, language modeling, video understanding, or image generation. Our choice to prioritize currently popular, easily-accessible, and well-studied datasets and models could end up reinforcing existing selection effects. Perhaps because currently popular \ta co-evolved with the most popular application domains, they work unusually well together. In this case, a benchmark emphasizing existing popular \wls will make it hard to break out of what could be a methodological local optimum. The advent of new \tas could potentially unlock efficient training for entirely different models, ones that are not covered by our benchmark. In the near-term, it is impossible to resolve these types of counterfactuals, but we could potentially create a more diverse---along every dimension we can measure---set of benchmark \wls, if we saved resources in other ways.
Additionally, as discussed in \Cref{sec:exp_baselines_timing}, we would like to have slightly fewer data-pipeline-bound \wls.
Moving away from the limitations of our particular set of current \wls, the benchmark today includes only a single hardware weight class. Although we believe results at this scale are relevant at a variety of interesting scales, the gold standard would be to directly measure much larger and smaller scales. That said, even if the benchmark results are informative for many different scales, the relative performance of different \tas might systematically vary with batch size. By only using a single specific system with a particular amount of accelerator memory, the benchmark effectively covers only the narrow range of batch sizes for each \wl that are close to the largest batch size that can fit in memory for the most competitive submissions.

\paragraph{Accessibility}
In order for our benchmark to be useful, scoring the most intriguing new \tas developed by the research community needs to be feasible. A large part of whether it is feasible for a particular group to evaluate a submission on the benchmark is the compute costs of the scoring protocol, running on the official benchmark system. \Cref{tab:compute_costs} lists the provisional\footnote{We plan to reduce the running time of some of the more expensive \wls before issuing a call for submissions, so these costs are likely a bit of an overestimate, although they should be the correct order of magnitude.} number of machine-hours required, on the official benchmark system using \benchmarkinghardware, to evaluate submissions under both tuning rulesets. Although that tuning can run on a different machine and need not use the official benchmark system, it is still worth computing the number of machine-hours on the benchmark system as a reference point. There are a variety of ways these costs could come down. For example, we could switch to a system with newer GPUs that speeds up training enough to be worth any concomitant increase in the price per machine-hour. Or, as submissions become more competitive, we could shrink the maximum \runtime allowed.

\begin{table}[!htp]
	\centering
	\scriptsize
	\begin{tabular}{lS[table-format=5.2]}
\toprule
\textbf{Setting}                        & \textbf{Time (h)}                             \\
\midrule
\textbf{External Tuning Ruleset}                                               &                              \\
\quad One \hp                                                  & 232.23                       \\
\quad Scoring a submission                                      & 1161.13                      \\
\quad Tuning a submission                                      & 23222.61                     \\
\addlinespace\textit{Qualification Set}                                 &                              \\
\quad One \hp                                                    & 20.65                        \\
\quad Scoring a submission                                       & 103.24                       \\
\quad Tuning a submission                                       & 2064.75                      \\
                                                                        &                              \\
\textbf{Self-tuning Ruleset}                                            &                              \\
\quad One \hp                                                   & 696.68                       \\
\quad Scoring a submission                                     & 3483.39                      \\
\addlinespace\textit{Qualification Set}                                 &                              \\
\quad One \hp                                                    & 61.94                        \\
\quad Scoring a submission                                      & 309.71                       \\
\bottomrule
\end{tabular}

 	\caption{\textbf{Estimated required \runtime of the benchmark.} \textit{One \hp} refers to running all eight fixed and eight \heldout \wls once, \ie with a single \hp. \textit{Scoring a submission} involves repeating this process for each study, \ie five times. To fully \textit{tune a submission}, each study uses twenty tuning trial to identify the best \hp setting (note that this need not use the benchmark system). Running a single \hp is more expensive in the self-tuning \ruleset since it has a three times larger \runtime budget for every \wl to compensate for the lack of external tuning. The \textit{qualification set} only consists of three (out of the eight) fixed \wls, without \heldout \wls, and thus offers a reduce cost.}
	\label{tab:compute_costs}
\end{table}

Nevertheless, as it stands, for groups that don't have several \benchmarkinghardware machines on premises, cloud costs to score submissions could be a significant hurdle. On the other hand, groups that train very large language models routinely run experiments eclipsing these scoring costs by orders of magnitude. Ultimately, whether or not evaluating a submission is affordable is a relative question that will depend on the specific research group. For now, our solution is to obtain compute sponsorship to help groups with more limited resources evaluate and score promising submissions.

In order to allocate limited sponsorship funding among submissions from groups unable to self-fund scoring costs, we plan to use performance on a qualification set of \wls that excludes some of the most expensive \wls. Specifically, the qualification set consists of the \criteodlrm, \ogbggnn, and \wmttransformer \wls, without any \heldout \wls. Submitters that do not have the resources to self-report results on the full set of \wls may instead report results on this smaller qualification set. As shown in \Cref{tab:compute_costs}, evaluating submissions on the qualification set is about an order of magnitude cheaper than the full set of \wls.

Stepping back, part of the issue of affordability is that our benchmark is all-or-nothing. To compute a benchmark score, we need to run submissions on all \wls. Thus, for any specific level of compute resources necessary for scoring, some groups will struggle to participate without compute support. It is hard to imagine this approach scaling to thousands of realistic \wls, especially if we consider tuning costs, without excluding far too many groups. Ideally, we would have a benchmark with a flexible mechanism for incrementally investing computational resources to gather more and more information about the performance of a submission, eventually culminating in a complete set of experiments on a large, highly-diverse set of fixed and held-out \wls. Even with our current design, we can compute an upper bound on a submission's benchmark score based on training on a subset of \wls and/or with a reduced \runtime limit, but our scoring procedure would probably need to be revised if we wanted it to scale to a much larger number of still-relatively-costly \wls.

\subsection{Future Work}
\label{sec:disc_future_work}

In addition to work on improving the benchmark itself and building new, and even stronger, baselines (especially for the self-tuning \ruleset), there are several related areas that would benefit from more research.
One appealing area of future work motivated by our challenges with building randomized \wls, as discussed previously in \Cref{sec:disc_randomized_workloads}, would be a theory to predict the effect of various \wl modifications on the optimal \hps, paving the way for a better understanding of how to perform \wl-specific tuning, how to build randomized \wls with support over a combinatorial space of variants, and when \wl-specific tuning is even necessary. Another direction with immense practical potential is developing completely \wl-agnostic training recipes that, when given an arbitrary neural network training \wl (model, dataset, and loss function) along with a budget, produce the best possible result. Even though we could use such recipes for target-setting, to be maximally interesting, they would need to go far beyond our target-setting procedure and be useful enough to be adopted by practitioners.

Although we restricted our attention to \tas in this work, there are other parts of the deep learning training pipeline that affect training speed. For the purposes of our \tas benchmark, the preprocessing, model architecture, and loss function are fixed components of the \wl, but they could all benefit from algorithmic improvements.
The initial proposal for the benchmark rules also included a separate, time-to-result model benchmark that measured training speedups due to model changes. In order to isolate the effects of model changes, this model benchmark required submissions to train using a small set of standard \tas. Unlike in the \ta benchmark, models would only be required to perform well on a single task, albeit across different \datasets (e.g. different language pairs in machine translation). Consequently, task-specific components of the \ml pipeline, such as data augmentation, could be incorporated as part of the submission in a hypothetical future \algoperf \textsc{Models} benchmark. With results from both \tas and separate model benchmarks, we could determine the relative responsibility for speedups on different tasks of the \tas vs the model architecture, and develop a complete picture of the most promising directions for accelerating neural network training.

\section{Conclusion}
\label{sec:conclusion}

Neural networks must be trained to be useful, making \tas essential for creating useful deep learning models. 
Unfortunately, due to the lack of a standard, convincing protocol for empirically comparing \tas, progress on better \tas has stalled.
To address this pressing issue, in this work, we introduced the \benchmarkname benchmark, a competitive, time-to-result benchmark covering multiple realistic \wls, running on fixed hardware.
This benchmark represents the collective efforts, over multiple years, of the members of the MLCommons Algorithms working group to remove the measurement barriers frustrating progress on neural network \tas.

Although we believe the benchmark we have created represents an important advance, it is far from perfect and has a variety of limitations.
We extend an open invitation to the entire community---and in particular, those who disagree with our design choices---to join the working group and collaborate on improving the benchmark further, either before we issue the initial call for submissions, or after. We would also be delighted by any novel solutions for the challenges with \ta comparisons we described in \Cref{sec:challenges}, whether they are possible to incorporate into \benchmarkname or necessitate an additional benchmark with a fundamentally different approach.

We urge researchers developing new training algorithms to submit them to the \benchmarkname benchmark competition once the call for submissions has been issued. Crafting a valid submission might require thinking more about \hp tuning than researchers inventing new optimizers are used to, but this work will help us break free of the cycle of hype and abandonment currently facing new algorithms. Outside the competition schedule, researchers can still use the benchmark to measure the performance of new algorithms as long as they adhere to the rules and report raw \wl times, in addition to recomputing unofficial benchmark scores. Researchers should feel free to reach out to the working group for guidance, as needed, or to find potential collaborators with the resources to run larger, more comprehensive experiments.
Finally, we believe that the presented benchmark constitutes a new state of the art for empirical comparisons of \tas, and should be viewed as a first step towards a more reproducible and empirically rigorous scientific literature on neural network \tas.

\newpage
\section*{Author Contributions and Acknowledgments}

\begin{itemize}
	\item \textbf{George E.~Dahl:} Founded and chaired the working group. Co-authored the initial rules proposal and shaped the rules of the benchmark. Recruited contributors to complete the necessary engineering work. Co-led paper experiments and co-authored the \href{\initcodebase}{codebase used for paper experiments} \citep{init2winit}. Co-led the paper's writing process and supervised the writing contributions from other authors. Directly contributed to every aspect of the writing process including outlining, drafting sections, creating figures, and editing. Served as overall project coordinator.
	\item \textbf{Frank Schneider:} Chaired the working group. Significantly influenced the rules of the benchmark.  Co-led the paper's writing process and supervised the writing contributions from other authors. Directly contributed to every aspect of the writing process including outlining, drafting sections, creating figures, and editing.
	\item \textbf{Zachary Nado:} Lead engineer, tech lead, and supervisor for building the \href{\mlccodebase}{benchmark codebase} and cloud infrastructure, including the implementation of the workloads in both \jax and \pytorch. Coordinated verifying the correctness across infrastructure implementations. Co-authored the \href{\initcodebase}{codebase for paper experiments}. Co-authored the initial rules proposal. Made major writing contributions to the paper.
	\item \textbf{Naman Agarwal:} Co-led paper experiments. Designed and ran a large number of critical experiments including target-setting experiments, baselines, \wl variants, and many others. Worked on initial implementations of the \fastmri and \vit \wls. Made major writing contributions to the paper. Significantly influenced the rules of the benchmark.
	\item \textbf{Chandramouli Shama Sastry:} Implemented both \librispeech \wls in \pytorch. Developed test suites to compare the implementations of \wls across frameworks that caught several implementation deficiencies. Contributed to the benchmark infrastructure.	
	\item \textbf{Philipp Hennig:} Significantly influenced the rules of the benchmark and made major writing contributions to the paper.	
	\item \textbf{Sourabh Medapati:} Assisted with the writing of the paper. Implemented both \librispeech \wls in \jax and helped debug them in \pytorch. Conducted several experiments. Helped maintain the \href{\initcodebase}{paper experiment codebase}. Contributed to the benchmark infrastructure.	
	\item \textbf{Runa Eschenhagen:} Significantly contributed to the benchmark infrastructure, leading the \pytorch development for a large portion of the development cycle. Implemented the \wmt \wl in \pytorch and the \ogbg \wl in both \pytorch and \jax. Debugged numerous critical issues in the \href{\mlccodebase}{benchmark codebase} and helped refine APIs.	
	\item \textbf{Priya Kasimbeg:} Led the timing experiments between our \pytorch and \jax implementations and debugged several critical issues in the \href{\mlccodebase}{benchmark codebase}. Contributed to the benchmark infrastructure. Helped maintain the \href{\initcodebase}{paper experiment codebase}.
	\item \textbf{Daniel Suo:} Implemented the initial version of the \fastmri and \vit \wls in \jax. Created the performance profile and scoring infrastructure. Made a large number of essential contributions to the \href{\initcodebase}{main codebase for paper experiments}. Contributed to the benchmark infrastructure. Assisted with the writing of the paper.
	\item \textbf{Juhan Bae:} Implemented the \vit, \fastmri, and \criteo \wls in \pytorch. Debugged several critical issues in the \href{\mlccodebase}{benchmark codebase}. Contributed to the benchmark infrastructure and documentation.	
	\item \textbf{Justin Gilmer:} Made major writing contributions to the paper. Conducted experiments highlighting the need for \wl standardization and designed several \heldout \wl variants. Co-authored the initial rules proposal. Co-authored the \href{\initcodebase}{codebase for paper experiments}.	
	\item \textbf{Abel L.~Peirson:} Explored ways to define randomized \wl distributions for the \ogbg base \wl and implemented the \cifarten development \wl for \jax. Assisted with the writing.	
	\item \textbf{Bilal Khan:} Conducted the baseline experiments for \distshampoo, \sam, \adafactor, and \lamb. Assisted with the writing of the paper.	
	\item \textbf{Rohan Anil:} Co-authored the initial rules proposal. Advised on the setup of the \jax baselines, and optimizer configuration details, generally helped with performance tuning and debugging \jax/\pytorch differences, and assisted with writing the paper.	
	\item \textbf{Mike Rabbat:} Significantly influenced the rules of the benchmark, assisted with the \pytorch implementation, provided support for the \fastmri \wl, and made major writing contributions to the paper.	
	\item \textbf{Shankar Krishnan:} Assisted with the writing of the paper. Conducted several experiments.
	\item \textbf{Daniel Snider:} Implemented the \resnetfifty \wl in \jax. Conducted preliminary experiments for randomized \wls on \ogbg, wrote code for logging, and explored options for serving the benchmark. Drafted rules around software dependencies.
	\item \textbf{Ehsan Amid:} Assisted with the writing of the paper. Supported the implementation of \criteo for \jax.
	\item \textbf{Kongtao Chen:} Implemented the initial drafts of the \deepspeech \wl in \pytorch and the \wmt \wl in \jax.
	\item \textbf{Chris J.~Maddison:} Co-authored the initial rules proposal.
	\item \textbf{Rakshith Vasudev:} Supported the implementation of the \criteo \wl for \pytorch.
	\item \textbf{Michal Badura:} Assisted with the writing of the paper. Wrote the initial \jax implementation of the \ogbg \wl in the \href{\initcodebase}{main paper experiment codebase}.
	\item \textbf{Ankush Garg:} Supported the implementation of the \wmt \wl. Assisted with the writing of the paper.
	\item \textbf{Peter Mattson:} Co-authored the initial rules proposal.
\end{itemize}

The Brain team at Google Research supported the initial work on the benchmark and the \jax baselines.

The authors would like to express their gratitude to David Kanter and the entire MLCommons organization for their support throughout the project. Many thanks to Toby Boyd for his help in preparing a request for compute sponsorship and additional logistical support. We are thankful to Hanlin Tang for his help in designing the initial \pytorch API requirements and implementing the \resnetfifty\wl in \pytorch. We thank Leda Sari for providing a reference implementation and her expertise for the \librispeech workloads. Thanks to Guodong Zhang for helpful suggestions regarding the rules and the submission API. We thank Dami Choi and Roger Grosse for helpful discussions. Furthermore, we would like to thank Varun Godbole for helpful feedback on this manuscript, Lucas Nestler for his help in implementing \librispeech \wls in \jax, and Kamal Raj for helping formulate Docker instructions. Finally, we'd like to especially thank all the members of the MLCommons Algorithms working group.

Frank Schneider is supported by funds from the Cyber Valley Research Fund.
Philipp Hennig and Frank Schneider gratefully acknowledge financial support by the European Research Council through ERC StG Action 757275/PANAMA; the DFG Cluster of Excellence ``Machine Learning - New Perspectives for Science'', EXC 2064/1, project number 390727645; the German Federal Ministry of Education and Research (BMBF) through the Tübingen AI Center (FKZ:01IS18039A); and funds from the Ministry of Science, Research and Arts of the State of Baden-Württemberg.
Daniel Snider is supported by funds from the Canada Foundation for Innovation JELF grant, NSERC Discovery grant, AWS Machine Learning Research Award (MLRA), Facebook Faculty Research Award, Google Scholar Research Award, and VMware Early Career Faculty Grant.
 
\newpage

\begin{appendices}

\section{Experimental Details for \texorpdfstring{\Cref{sec:challenges}}{Section 2}}
\label{app:exp_details_chal}

\subsection{Learning Rate Schedules}
\label{app:exp_details_chal_schedules}

\begin{figure}[!ht]
	\includegraphics[width=\textwidth]{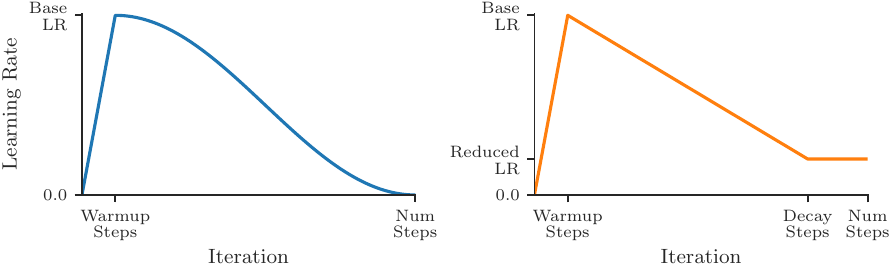}
	\caption{\textbf{The two learning rate schedules used in our experiments.} We use a cosine decay schedule with a learning rate \warmup phase (\textbf{\wcd}, \textit{left}, \colorline{SNSblue}) and a linear decay schedule with a learning rate \warmup phase and a constant phase at the end (\textbf{\wldc}, \textit{right}, \colorline{SNSorange}). The schedules are scaled to fill the entire $\mathrm{Num Steps}$ and are further parameterized by the parameters shown in the respective figures.}
	\label{fig:lr_schedules}
\end{figure}

Throughout the experiments in this paper (not only those reported in \Cref{sec:challenges}), we use two types of learning rate schedules.
Both schedules are illustrated in \Cref{fig:lr_schedules} and their details are provided in the following.

\subsubsection{\Warmup Cosine Decay}

The cosine decay with \warmup (denoted \textit{\wcd}) is characterized by a linear \warmup phase followed by a cosine decay \citep{Loshchilov2017} of the learning rate.
It is parameterized by the following parameters:
\begin{itemize}
	\item \textbf{Base LR:} The base learning rate. It is used as the peak learning rate of the \warmup part and the cosine decay.
	\item \textbf{Num Steps:} The total number of steps of the schedule.
	\item \textbf{\Warmup Steps:} The number of \warmup steps over which the learning rate is linearly increased from zero to the base learning rate. Here, the \warmup steps, characterize the absolute number of steps over which the \warmup is performed. For simplicity, in our code, we instead provide the \warmup phase in terms of the percentage of the total number of steps in the schedule (see for example \Cref{tab:target-setting-opt-hparams}).
\end{itemize}

Mathematically, the learning rate $\mathrm{LR}$ at any step $t \geq 0$ of the \textit{\wcd} schedule is defined by\begin{align*}
	\mathrm{LR}(t) = \begin{cases}
		\mathrm{BaseLR}\,\frac{t}{\mathrm{WarmupSteps}} &\text{$t \leq \mathrm{WarmupSteps}$} \\[0.25em]
		\frac{\mathrm{BaseLR}}{2}\left( 1 + \cos\left(\pi \left( \frac{t - \mathrm{WarmupSteps}}{\mathrm{NumSteps} - \mathrm{WarmupSteps}} \right)\right) \right) &\text{otherwise} \, .
	\end{cases}
\end{align*}

\subsubsection{\Warmup Linear Decay Constant}

The other learning rate schedule (denoted \textit{\wldc}) also employs a linear \warmup phase, this time followed by a \emph{linear} decay and a final phase of constant learning rate.
It is parameterized by the following parameters:
\begin{itemize}
	\item \textbf{Base LR:} The base learning rate. It is used as the upper learning rate of the \warmup part and the linear decay.
	\item \textbf{Num Steps:} The total number of steps of the schedule.
	\item \textbf{\Warmup Steps:} The number of \warmup steps over which the learning rate is linearly increased from zero to the base learning rate. Here, the \warmup steps, characterize the absolute number of steps over which the \warmup is performed. For simplicity, in our code and the rest of the paper, we instead provide the \warmup phase in terms of the percentage of the total number of steps in the schedule (see the entry \textit{warmup} for example in \Cref{tab:target-setting-opt-hparams}).
	\item \textbf{Reduced LR:} The reduced learning rate is the lower bound of the linear decay and is used for the final constant phase of the schedule. Here, the reduced learning rate denotes the absolute learning rate. For simplicity, in our code and the remaining paper, we characterize the reduced learning rate by a \textit{decay factor} (see the entry \textit{decay factor} for example in \Cref{tab:target-setting-opt-hparams}) relative to the base learning rate.
	\item \textbf{Decay Steps:} The absolute number of steps of the linear decay (including the \warmup steps). Once again, for simplicity in our code and throughout the paper, we use a slightly different version. There, we define the \textit{decay steps} relative to the total number of steps excluding the \warmup phase (see the entry \textit{decay steps} for example in \Cref{tab:target-setting-opt-hparams}).
\end{itemize}

Mathematically, the learning rate $\mathrm{LR}$ at any step $t \geq 0$ of the \textit{\wldc} schedule is defined by
\begin{align*}
	\mathrm{LR}(t) = \begin{cases}
		\mathrm{BaseLR}\,\frac{t}{\mathrm{WarmupSteps}} & t \leq \mathrm{WarmupSteps} \\[0.25em]
		\mathrm{BaseLR}\,\frac{\mathrm{DecaySteps} - t}{\mathrm{DecaySteps} - \mathrm{WarmupSteps}} \\+ \mathrm{ReducedLR}\,\frac{t - \mathrm{WarmupSteps}}{\mathrm{DecaySteps} - \mathrm{WarmupSteps}} & \mathrm{WarmupSteps} < t \leq \mathrm{DecaySteps} \\[0.25em]
		\mathrm{BaseLR}\,\mathrm{DecayFactor} & \text{otherwise} \, .
	\end{cases}
\end{align*}

\subsection{Details for Training Curves that Cross}
\label{app:exp_details_chal_crossing_curves}

The training curves presented in \Cref{fig:crossing_curves} in \Cref{sec:chal_training_speed} were drawn from preliminary target-setting experiments for \resnetfifty trained on \imagenet using \adamw.
They use the \wl configuration described in \Cref{app:wl_details_imagenet_resnet}.
The trials were selected from two arbitrary preliminary tuning studies that each had a budget of $100$ trials.
To select the two trials, we plotted all the training curves on a single plot, selecting two arbitrary trials that crossed and also achieved relatively good final validation error rates (in the top $20$ trials for both studies).
It was not necessary to find trials from different studies given how plentiful training curves that cross are in typical tuning studies.
The search spaces were the same as the \adamw search space listed in \Cref{tab:target-setting-search-spaces}, except they were from experiments that predate the final choice of allowed options for learning rate \warmup lengths, disabled dropout, and tuned label smoothing on a continuous range of $[0.0, 0.2]$.
These preliminary experiments also happened to be from studies using a batch size of $8192$ and $32768$, respectively.
The particular trials selected used a \warmup length of $15\%$ and $10\%$; the trial with the better final validation error used a batch size of $8192$ and a \warmup length of $15\%$.
Many pairs of trials in these studies cross, and the specific trials we selected were typical well-performing ones.
It is easy to find training curves that cross multiple times among the top dozen trials in almost any of our larger tuning studies.

\subsection{Details for Sensitivity of Optimizer Ranking to the Model Architecture}
\label{app:exp_details_chal_model_changes}

\subsubsection{\wideresnet with Stride Changes}
\label{app:exp_details_chal_wideresnet}

\begin{figure}[!ht]
	\includegraphics[width=\textwidth]{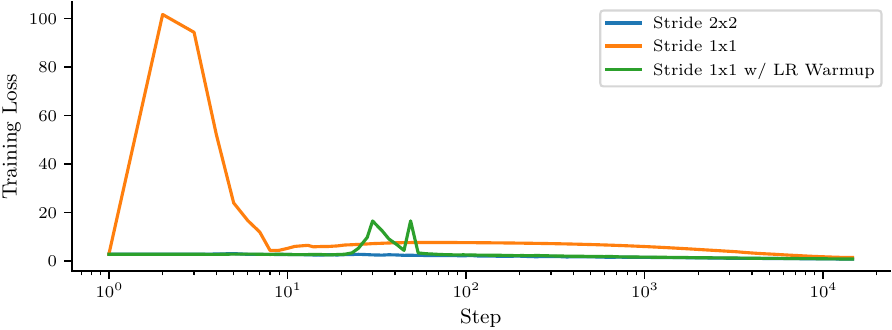}
	\caption{{\bf Visualizing the training instability caused by changes to the \wideresnet convolutional stride.} For the standard Stride 2x2 architecture (\colorline{SNSblue}), the training loss decreases almost monotonically throughout training. However, the Stride 1x1 architecture (\colorline{SNSorange}) shows a brief period of training instability, followed by a recover all within the first $10$ steps of training. Despite the recovery, the long term optimization trajectory is affected and the stride change under-performs the standard architecture. Learning rate \warmup helps mitigate this instability and allows the Stride 1x1 architecture to match the performance of the original model (\colorline{SNSgreen}). All runs are shown for a learning rate of $0.215$.}
	\label{fig:wrn_losscurve}
\end{figure}

The standard \wideresnet 28-10 architecture consists of 3 groups of 4 residual blocks, each block containing 2 convolutional layers. The default strides used in each of the three groups are 1$\times$1, 2$\times$2, and 2$\times$2 in groups 1, 2, and 3 respectively. Our modified architectures changes the strides in group 3 from 2$\times$2 to 1$\times$1. The strides are what reduce the height  and width  dimensions in the embedded image tensor, so changing the strides from 2$\times$2 to 1$\times$1 means that no reduction occurs in the final group. The final operation before the dense layer is an 8$\times$8 average pooling operation followed by a flatten operation. So all remaining height ($H$), width ($W$) and channel ($C$) dimensions are flattened into a single dim of size $H \times W \times C$. In the stride 1x1 variant both $H$ and $W$ increase by a factor of $9$, resulting in a factor of $81$ increase in the fan-in of the final dense layer (in particular the dense layer shape changes from 640$\times$10 to 51840$\times$10). This dense layer is initialized with the LeCun normal initialization, which makes the variance of the forward pass invariant to the fan-in. However, the statistics of the backward pass are not invariant to the fan-in (and more importantly the loss curvature itself is not invariant) and so the resulting model has an instability particularly with respect to the final output layer (see \Cref{fig:wrn_losscurve}).

\subsubsection{Architectural Modifications of Transformer Models}
\label{app:exp_details_pre_post_ln}

This section provides additional data for the experiment presented in \Cref{sec:chal_model_changes}. Specifically, \Cref{tab:pre_post_ln_ce} is a variant of \Cref{tab:pre_post_ln_bleu} but showing the cross-entropy loss instead of the BLEU score.
\begin{table}[!ht]
	\centering
	\begin{tabular}{lcc@{\hskip 0.05\textwidth}cc}
    \toprule
    \textbf{Training} & \multicolumn{2}{c}{\textbf{\preln}} & \multicolumn{2}{c}{\textbf{\postln}} \\
    \cmidrule(lr){2-3} \cmidrule(lr){4-5}
    \textbf{Algorithm} & \textit{Best} & \textit{Confidence interval} & \textit{Best} & \textit{Confidence interval} \\
    \midrule
    \adamw & $1.3093$ & $1.3449 \pm 0.0337$ & $1.3632$ & $1.4934 \pm 0.1327$ \\
    \nadamw & $1.3070$ & $1.3308 \pm 0.0240$ & $1.3539$ & $1.5072 \pm 0.1822$ \\
    \nesterov & $1.5668$ & $1.6362 \pm 0.1047$ & $1.9577$ & $2.5451 \pm 0.6894$ \\
    \shampoo & $1.4096$ & $1.4400 \pm 0.0318$ & $1.4516$ & $1.5330 \pm 0.1452$ \\
    \bottomrule
\end{tabular}
 	\caption{\textbf{Test set cross-entropy score for \prelnfull (\preln) and \postlnfull (\postln) architectures for different training algorithms.} The architectural modification of \preln vs. \postln affects \nesterov momentum, \adamw, \nadamw, and \shampoo differently. This is the same as \Cref{tab:pre_post_ln_bleu} but showing the cross-entropy loss instead of the \bleu score.}
	\label{tab:pre_post_ln_ce}
\end{table}

\subsection{Details for Comparing Instances of Training Algorithms}
\label{app:exp_details_chal_comparing_algorithms}

\subsubsection{Training Algorithms with Different Hyperparameters}
\label{app:exp_details_chal_comparing_hyperparameters}

This section provides additional details for the experiments described in \Cref{sec:chal_comparing_hyperparams} and summarized in \Cref{tab:hparams_workloads_joint}.
We report the performance of the \textit{per-workload optimal} \hps, of the \textit{overall optimal} \hps, as well as the resulting relative performance degradation $\phi(H)$ from using a single shared \hp setting, which is defined as
\begin{align*}
	\phi_w(H) = \left| \frac{\val(w, h^\ast) - \val_H(w)}{\val_H(w)} \right| \,
\end{align*}
where $h^\ast$ is the overall optimal \hp setting. It holds that $\max_w \phi_w(H) = \Phi(H)$ as reported in \Cref{sec:chal_comparing_hyperparams}. \Cref{tab:hparams_workloads_adam,tab:hparams_workloads_nadam,tab:hparams_workloads_nesterov,tab:hparams_workloads_heavyball} show the quantities mentioned above for the four target-setting \tas.
\begin{table}[!htp]
	\centering
	\small

\begin{tabular}{llS[table-format=2.6]S[table-format=2.6]S[table-format=1.6]}
	\toprule
	\textbf{Workload} & & \multicolumn{2}{c}{\textbf{Performance of the}} & $\bm{\phi_w(H)}$\\
	\cmidrule(lr){3-4}
	& & \textit{Per-Workload Optimal} & \textit{Overall Optimal} & \\
	& & \textit{Hyperparameters} & \textit{Hyperparameters} &\\
	\midrule
    \criteo &\dlrmsmall &0.123675 &   0.124022 &  0.002806 \\ \addlinespace
    \fastmri &\unet &0.734330 &   0.731403 &  0.003986 \\ \addlinespace
    \imagenet &\resnet50 &0.230340 &   0.267460 &  0.161153 \\
     &\vit &0.226140 &   0.263300 &  0.164323 \\ \addlinespace
    \librispeech &\conformer &0.078327 &   0.093633 &  \B 0.195425 \\
     &\deepspeech &0.114152 &   0.133201 &  0.166873 \\ \addlinespace
    \ogbg &\gnn &0.277534 &   0.252594 &  0.089862 \\ \addlinespace
    \wmt &\transformer &30.693876 &  29.067333 &  0.052992 \\
    \bottomrule
\end{tabular}

 	\caption{\textbf{Performance of the \textit{per-workload optimal} and \textit{overall optimal} \hps for \adamw.} A column with $\phi_w(H)=0$ indicates that the overall optimal \hps achieved the same performance as the per-\wl optimal \hps on this \wl. The $\phi_w$-value in bold indicates the largest value observed across all \wls, which is reported as $\Phi(H)$ in \Cref{tab:hparams_workloads_joint}.}
	\label{tab:hparams_workloads_adam}
\end{table}
\begin{table}[!htp]
	\centering
	\small
	
\begin{tabular}{llS[table-format=2.6]S[table-format=2.6]S[table-format=1.6]}
	\toprule
	\textbf{Workload} & & \multicolumn{2}{c}{\textbf{Performance of the}} & $\bm{\phi_w(H)}$\\
	\cmidrule(lr){3-4}
	& & \textit{Per-Workload Optimal} & \textit{Overall Optimal} & \\
	& & \textit{Hyperparameters} & \textit{Hyperparameters} &\\
	\midrule
    \criteo &\dlrmsmall &0.123609 &   0.135988 &  0.100147 \\ \addlinespace
    \fastmri &\unet &0.734523 &   0.710679 &  0.032462 \\ \addlinespace
    \imagenet &\resnet50 &0.227020 &   0.261320 &  0.151088 \\
     &\vit &0.225300 &   0.263420 & \B 0.169197 \\ \addlinespace
    \librispeech &\conformer &0.077790 &   0.089341 &  0.148486 \\
     &\deepspeech &0.113950 &   0.121887 &  0.069654 \\ \addlinespace
    \ogbg &\gnn &0.280012 &   0.274704 &  0.018954 \\ \addlinespace
    \wmt &\transformer &30.853422 &  29.747388 &  0.035848 \\
    \bottomrule
\end{tabular}
 	\caption{\textbf{Performance of the \textit{per-workload optimal} and \textit{overall optimal} \hps for \nadamw.} A column with $\phi_w(H)=0$ indicates that the overall optimal \hps achieved the same performance as the per-\wl optimal \hps on this \wl. The $\phi_w$-value in bold indicates the largest value observed across all \wls, which is reported as $\Phi(H)$ in \Cref{tab:hparams_workloads_joint}.}
	\label{tab:hparams_workloads_nadam}
\end{table}
\begin{table}[!htp]
	\centering
	\small
	
\begin{tabular}{llS[table-format=2.6]S[table-format=2.6]S[table-format=1.6]}
	\toprule
	\textbf{Workload} & & \multicolumn{2}{c}{\textbf{Performance of the}} & $\bm{\phi_w(H)}$\\
	\cmidrule(lr){3-4}
	& & \textit{Per-Workload Optimal} & \textit{Overall Optimal} & \\
	& & \textit{Hyperparameters} & \textit{Hyperparameters} &\\
	\midrule
    \criteo &\dlrmsmall &0.126139 &   0.144934 &  0.149004 \\\addlinespace
    \fastmri &\unet &0.734645 &   0.733440 &  0.001640 \\\addlinespace
    \imagenet &\resnet50 &0.226600 &   0.278720 &  \B 0.230009 \\
     &\vit &0.243180 &   0.278700 &  0.146065 \\\addlinespace
    \librispeech &\conformer &0.130823 &   0.130823 &  0.0 \\
     &\deepspeech &0.171137 &   0.192623 &  0.125546 \\\addlinespace
    \ogbg &\gnn &0.283124 &   0.226850 &  0.198761 \\\addlinespace
    \wmt &\transformer &30.107387 &  26.977169 &  0.103968 \\
    \bottomrule
\end{tabular}

 	\caption{\textbf{Performance of the \textit{per-workload optimal} and \textit{overall optimal} \hps for \nesterov.} A column with $\phi_w(H)=0$ indicates that the overall optimal \hps achieved the same performance as the per-\wl optimal \hps on this \wl. The $\phi_w$-value in bold indicates the largest value observed across all \wls, which is reported as $\Phi(H)$ in \Cref{tab:hparams_workloads_joint}.}
	\label{tab:hparams_workloads_nesterov}
\end{table}
\begin{table}[!htp]
	\centering
	\small

\begin{tabular}{llS[table-format=2.6]S[table-format=2.6]S[table-format=1.6]}
	\toprule
	\textbf{Workload} & & \multicolumn{2}{c}{\textbf{Performance of the}} & $\bm{\phi_w(H)}$\\
	\cmidrule(lr){3-4}
	& & \textit{Per-Workload Optimal} & \textit{Overall Optimal} & \\
	& & \textit{Hyperparameters} & \textit{Hyperparameters} &\\
	\midrule
    \criteo &\dlrmsmall &0.125913 &   0.145933 &  0.158998 \\ \addlinespace
    \fastmri &\unet &0.733828 &   0.731964 &  0.002539 \\ \addlinespace
    \imagenet &\resnet50 &0.225340 &   0.279280 &  \B 0.239372 \\
     &\vit &0.244860 &   0.286660 &  0.170710 \\ \addlinespace
    \librispeech &\conformer &0.132797 &   0.134879 &  0.015684 \\
     &\deepspeech &0.161977 &   0.186385 &  0.150687 \\ \addlinespace
    \ogbg &\gnn &0.276148 &   0.223901 &  0.189199 \\ \addlinespace
    \wmt &\transformer &30.643066 &  26.705241 &  0.128506 \\
    \bottomrule
\end{tabular}
 	\caption{\textbf{Performance of the \textit{per-workload optimal} and \textit{overall optimal} \hps for \heavyball.} A column with $\phi_w(H)=0$ indicates that the overall optimal \hps achieved the same performance as the per-\wl optimal \hps on this \wl. The $\phi_w$-value in bold indicates the largest value observed across all \wls, which is reported as $\Phi(H)$ in \Cref{tab:hparams_workloads_joint}.}
	\label{tab:hparams_workloads_heavyball}
\end{table}

\subsubsection{Training Algorithms with Different \Hp Search Spaces}
\label{app:exp_details_chal_comparing_search_spaces}

In \Cref{tab:search_space_importance_25p_t5}, we report additional data for the \adamw search space comparison presented in \Cref{sec:chal_comparing_search_spaces}. Specifically, we show the results presented in \Cref{tab:search_space_importance_results} but showing a different budget of tuning trials, \ie $5$ trials instead of the $20$ trials reported in the main text.

\begin{table}[!htp]
	\centering
	\scriptsize
	\begin{tabular}{
		l l S[table-format=2.6] S[table-format=2.6] S[table-format=2.6] S[table-format=2.6] S[table-format=2.6] S[table-format=2.6] }
\toprule
\textbf{Workload}         &              & \multicolumn{3}{c}{\textbf{\textsc{AdamW Narrow}}} & \multicolumn{3}{c}{\textbf{\textsc{AdamW Broad}}} \\ \cmidrule(lr){3-5} \cmidrule(lr){6-8}
&              & {Median}      & {$Q_1$} & {$Q_3$} & {Median}   & {$Q_1$} & {$Q_3$} \\
\midrule
\criteo            & \dlrmsmall   & \B 0.124039  & 0.124016       & 0.124074       & 0.124319 & 0.124118        & 0.124634       \\
\addlinespace
\fastmri     & \unet        & \B 0.734512    & 0.734335   & 0.734682   & 0.733791    & 0.733094    & 0.734065    \\
\addlinespace
\imagenet    & \resnetfifty & \B 0.23382     & 0.23298    & 0.2357    & 0.26842     & 0.24708     & 0.291495     \\
& \vit         & \B 0.22324  & 0.22108        & 0.22526        & 0.24690  & 0.23924        & 0.27028        \\
\addlinespace
\librispeech & \conformer   & \B 0.077553    & 0.076817   & 0.079047   & 0.090477    & 0.085484    & 0.106039    \\
& \deepspeech  & \B 0.115082 & 0.112909       & 0.117964       & 0.131462 & 0.124718       & 0.155395       \\
\addlinespace
\ogbg & \gnn         & \B 0.279426  & 0.277748       & 0.281602       & 0.269102 & 0.265382       & 0.275955       \\
\addlinespace
\wmt  & \transformer & \B 31.1849  & 30.9471        & 31.2992        & 30.4967  & 29.2948        & 30.9648 \\
\bottomrule
\end{tabular}
 	\caption{\textbf{Performance across multiple \wls for \adamw with two different hyperparameter search spaces.} Shown are the median, as well as the lower and upper quartiles ($Q_1$ and $Q_3$) of the best observed validation metric. The results are for a budget of $T\!=\!5$ trials across $1000$ simulations. This table is similar to \Cref{tab:search_space_importance_results} but showing a different budget of tuning trials ($5$ instead of $20$).}
	\label{tab:search_space_importance_25p_t5}
\end{table}

\subsubsection{Training Algorithms with Different Tuning Goals}
\label{app:exp_details_chal_comparing_tuning_goals}

In this section, we provide more details from the experiment presented in \Cref{sec:chal_comparing_search_spaces} and illustrated in \Cref{fig:different_tuning_goal}. For this, we ran \hp tuning studies for \resnetfifty on \imagenet trained with \adamw with two different training step budgets, $186,666$ and $233,333$. Both studies used the same search space and used $100$ tuning trials of \quasirandom search. They both used a \cosinedecay schedule and a linear learning rate \warmup. Although the \cosinedecay schedule is a function of the maximum number of training steps, the only learning rate schedule parameter that was tuned beyond the peak learning rate was the length of the \warmup, which was tuned over three discrete options: $2\%$, $5\%$, or $10\%$ of the step budget (for the complete search space see \Cref{tab:target-setting-search-spaces}, \adamw). We used the same seed for \quasirandom search in both studies to generate the exact same set of $100$ \hp points. With the search space parameterized in this way to be relative to the training step budget and with the exact same set of 100 \hp points, the \hp setting achieving the best validation error happened to be the same across the two studies (\Cref{tab:different-step-budget-opt-hparams}).
\begin{table}[!htp]
	\centering
	\begin{tabular}{lS[table-format=1.20]}
    \toprule
    \textbf{\Hp} & \textbf{Value} \\
    \midrule
    Base LR & 0.00040908031497988146 \\
    Weight decay & 0.5107969085979827 \\
    \betaone & 0.9978657786056152 \\
    \betatwo & 0.9961526971493336 \\
    Warmup & \text{$5\%$} \\
    Label smoothing & 0.1986402587653765\\
    \bottomrule
\end{tabular}
 	\caption{\textbf{\Hp values for both runs shown in \Cref{fig:different_tuning_goal}.} Although both trials use a different step budget ($186,666$ vs. $233,333$) the \hp values found after tuning are the same.}
	\label{tab:different-step-budget-opt-hparams}
\end{table}
We selected the best trial based on minimum validation error achieved at \emph{any} point during training, not just at the end of training, but with a \cosinelrdecay schedule we should expect the best result to be at---or near---the end of training.

\section{Details for Target-Setting Experiments (\texorpdfstring{\Cref{sec:exp_target_setting}}{Section 5})}
\label{app:exp_details_exp_target_setting}

\Cref{tab:target-setting-top-20trials-valid,tab:target-setting-top-20trials-test} provide the results of all $20$ reruns of the target-setting \ta for each \wl.
\begin{table}[!htp]
	\centering
	\scriptsize
	\setlength\tabcolsep{2pt} \begin{tabular}{@{}l@{\hskip-6pt}S[table-format=1.6]S[table-format=1.6]S[table-format=1.6]S[table-format=1.6]S[table-format=2.6]S[table-format=2.6]S[table-format=1.6]S[table-format=2.4]@{}}
	\toprule
	\textbf{Workload} &\textbf{\criteo} &\textbf{\fastmri} & \multicolumn{2}{c}{\textbf{\imagenet}} & \multicolumn{2}{c}{\textbf{\librispeech}} &\textbf{\ogbg} &\textbf{\wmt} \\
	\cmidrule(lr){4-5} \cmidrule(lr){6-7}
	&\textbf{\dlrmsmall} &\textbf{\unet} &\textbf{\resnetfifty} &\textbf{\vit} &\textbf{\conformer} &\textbf{\deepspeech} &\textbf{\gnn} &\textbf{\transformer} \\
	\midrule
	Metric & {CE} \lib & {SSIM} \gib & \multicolumn{2}{c}{{Error Rate} \lib} & \multicolumn{2}{c}{{WER} \lib} & {mAP} \gib & {\bleu} \gib \\
	\midrule
	0 &0.123585 &0.733429 &0.22484 &0.22436 &0.076226 &0.113374 &0.275715 &30.7062 \\
	1 &0.123589 &0.733868 &0.22488 &0.2255 &0.07658 &0.115416 &0.276308 &30.7142 \\
	2 &0.123609 &0.733994 &0.2249 &0.22606 &0.076726 &0.115487 &0.277504 &30.726 \\
	3 &0.123619 &0.734157 &0.2251 &0.22618 &0.076862 &0.115659 &0.277586 &30.7814 \\
	4 &0.123624 &0.734161 &0.22512 &0.22626 &0.077372 &0.11573 &0.277711 &30.7992 \\
	5 &0.123631 &0.734162 &0.22512 &0.22628 &0.077453 &0.11574 &0.278202 &30.7993 \\
	6 &0.123634 &0.734205 &0.22518 &0.22632 &0.077517 &0.115922 &0.278523 &30.8069 \\
	7 &0.123635 &0.734223 &0.22534 &0.22674 &0.077599 &0.115982 &0.278594 &30.8229 \\
	8 &0.123637 &0.734276 &0.22546 &0.2268 &0.077653 &0.116134 &0.279844 &30.829 \\
	9 &0.123649 &0.73432 &0.22562 &0.22688 &0.078145 &0.116144 &0.280716 &30.8458 \\
	10 &0.123649 &0.73448 &0.22576 &0.22694 &0.078809 &0.116255 &0.281243 &30.8524 \\
	11 &0.123662 &0.734483 &0.2258 &0.22726 &0.078881 &0.116478 &0.281428 &30.8571 \\
	12 &0.123664 &0.734532 &0.22586 &0.22766 &0.079063 &0.116791 &0.281518 &30.8631 \\
	13 &0.123668 &0.734587 &0.22598 &0.22776 &0.079718 &0.117135 &0.281907 &30.8787 \\
	14 &0.123673 &0.734605 &0.22598 &0.22796 &0.079718 &0.117378 &0.282107 &30.8835 \\
	15 &0.123687 &0.734658 &0.22608 &0.2284 &0.084566 &0.117398 &0.282539 &30.9064 \\
	16 &0.1237 &0.734674 &0.22628 &0.2284 &0.091796 &0.117438 &0.282713 &30.9196 \\
	17 &0.123715 &0.734712 &0.22664 &0.22854 &0.093352 &0.117812 &0.282918 &30.9267 \\
	18 &0.123728 &0.734804 &0.22706 &0.22886 &0.098336 &0.131573 &0.283148 &30.9446 \\
	19 &0.12374 &0.735031 &0.22746 &0.2293 &0.104256 &0.171481 &0.288737 &31.0684 \\
	\addlinespace
	min &0.123585 &0.733429 &0.22484 &0.22436 &0.076226 &0.113374 &0.275715 &30.7062 \\
	25th &0.12363 &0.734162 &0.22512 &0.226275 &0.077433 &0.115737 &0.278079 &30.7993 \\
	median &0.123649 &0.7344 &0.22569 &0.22691 &0.078477 &0.1162 &0.28098 &30.8491 \\
	75th &0.123676 &0.734619 &0.226005 &0.22807 &0.08093 &0.117383 &0.282215 &30.8892 \\
	max &0.12374 &0.735031 &0.22746 &0.2293 &0.104256 &0.171481 &0.288737 &31.0684 \\
	\addlinespace
	$\text{max} - \text{min}$ &0.000155 &0.001602 &0.00262 &0.00494 &0.028031 &0.058107 &0.013022 &0.3621 \\
	$\frac{\text{max} - \text{min}}{\text{min}}$ $(\%)$ &0.13 &0.22 &1.17 &2.20 &36.77 &51.25 &4.72 &1.18 \\
	\bottomrule
\end{tabular}
 	\caption{\textbf{Validation evaluation metric results of the $20$ reruns of the top \ta and \hp combination for each \wl.}}
	\label{tab:target-setting-top-20trials-valid}
\end{table}
\begin{table}[!htp]
	\centering
	\scriptsize
	\setlength\tabcolsep{2pt} \begin{tabular}{@{}l@{\hskip-6pt}S[table-format=1.6]S[table-format=1.6]S[table-format=1.6]S[table-format=1.6]S[table-format=2.6]S[table-format=2.6]S[table-format=1.6]S[table-format=2.4]@{}}
	\toprule
	\textbf{Workload} &\textbf{\criteo} &\textbf{\fastmri} & \multicolumn{2}{c}{\textbf{\imagenet}} & \multicolumn{2}{c}{\textbf{\librispeech}} &\textbf{\ogbg} &\textbf{\wmt} \\
	\cmidrule(lr){4-5} \cmidrule(lr){6-7}
	&\textbf{\dlrmsmall} &\textbf{\unet} &\textbf{\resnetfifty} &\textbf{\vit} &\textbf{\conformer} &\textbf{\deepspeech} &\textbf{\gnn} &\textbf{\transformer} \\
	\midrule
	Metric & {CE} \lib & {SSIM} \gib & \multicolumn{2}{c}{{Error Rate} \lib} & \multicolumn{2}{c}{{WER} \lib} & {mAP} \gib & {\bleu} \gib \\
	\midrule
	0 &0.125989 &0.7404 &0.3371 &0.3396 &0.044773 &0.065546 &0.263254 &30.6455 \\
	1 &0.126002 &0.74106 &0.3405 &0.3403 &0.044849 &0.066671 &0.265461 &30.7219 \\
	2 &0.126029 &0.741231 &0.3412 &0.3408 &0.045175 &0.066862 &0.266552 &30.7898 \\
	3 &0.126035 &0.741236 &0.3415 &0.3414 &0.04571 &0.066883 &0.267611 &30.8858 \\
	4 &0.126039 &0.741361 &0.3417 &0.3416 &0.046055 &0.066926 &0.268155 &30.8996 \\
	5 &0.126042 &0.741451 &0.3425 &0.3422 &0.046189 &0.067499 &0.268729 &30.9216 \\
	6 &0.126043 &0.74157 &0.343 &0.3422 &0.046265 &0.067584 &0.268964 &30.9228 \\
	7 &0.126044 &0.741643 &0.3432 &0.3426 &0.046476 &0.067732 &0.26905 &30.947 \\
	8 &0.126051 &0.741652 &0.3432 &0.3428 &0.046552 &0.067944 &0.269792 &30.9692 \\
	9 &0.126052 &0.741671 &0.3434 &0.3432 &0.046629 &0.067944 &0.269988 &30.9959 \\
	10 &0.126054 &0.741699 &0.3436 &0.3432 &0.046763 &0.068008 &0.270238 &30.9987 \\
	11 &0.12606 &0.741721 &0.3437 &0.3437 &0.046954 &0.068029 &0.270672 &31.005 \\
	12 &0.126071 &0.741738 &0.3438 &0.3438 &0.046973 &0.068093 &0.270756 &31.0129 \\
	13 &0.126076 &0.741778 &0.3438 &0.3439 &0.047184 &0.068369 &0.271233 &31.0221 \\
	14 &0.126076 &0.741827 &0.3439 &0.3444 &0.048638 &0.068454 &0.271476 &31.0408 \\
	15 &0.126087 &0.7419 &0.344 &0.3451 &0.050551 &0.069112 &0.272067 &31.0412 \\
	16 &0.126087 &0.741909 &0.3449 &0.3451 &0.055086 &0.069282 &0.272105 &31.1161 \\
	17 &0.126126 &0.741915 &0.3453 &0.3457 &0.057286 &0.069494 &0.27285 &31.1407 \\
	18 &0.126128 &0.742025 &0.3462 &0.3466 &0.060749 &0.081402 &0.276531 &31.1446 \\
	19 &0.126136 &0.742195 &0.3481 &0.3481 &0.066317 &0.110948 &0.276976 &31.1679 \\
	\addlinespace
	min &0.125989 &0.7404 &0.3371 &0.3396 &0.044773 &0.065546 &0.263254 &30.6455 \\
	25th &0.126041 &0.741428 &0.3423 &0.34205 &0.046155 &0.067355 &0.268586 &30.9161 \\
	median &0.126053 &0.741685 &0.3435 &0.3432 &0.046696 &0.067976 &0.270113 &30.9973 \\
	75th &0.126079 &0.741845 &0.343925 &0.344575 &0.049116 &0.068618 &0.271624 &31.0409 \\
	max &0.126136 &0.742195 &0.3481 &0.3481 &0.066317 &0.110948 &0.276976 &31.1679 \\
	\addlinespace
	$\text{max} - \text{min}$ &0.000148 &0.001796 &0.011 &0.0085 &0.021544 &0.045402 &0.013722 &0.5225 \\
	$\frac{\text{max} - \text{min}}{\text{min}}$ $(\%)$ &0.12 &0.24 &3.26 &2.50 &48.12 &69.27 &5.21 &1.70 \\
	\bottomrule
\end{tabular}
 	\caption{\textbf{Test evaluation metric results of the $20$ reruns of the top \ta and \hp combination for each \wl.}}
	\label{tab:target-setting-top-20trials-test}
\end{table}

\section{Details for Baseline Experiments (\texorpdfstring{\Cref{sec:exp_baselines}}{Section 7})}
\label{app:exp_details_exp_baselines}

In this section, we provide additional results from the baseline experiments presented in \Cref{sec:exp_baselines}.
\Cref{tab:baseline-performance-time,tab:baseline-performance-steps} report the total \runtime and number of steps needed to achieve the target performance for all baselines.
The benchmark scores for all baselines presented in the paper are listed in \Cref{tab:baselines-benchmark-scores}.
\Cref{tab:baselines-runtimes} presents the estimated \runtimes for a given number of steps of the different \tas used in the baselines.
We report the $20$ \hp points used in our \optlisttext baselines in \Cref{tab:opt-list-adam,tab:opt-list-nadam,tab:opt-list-nesterov,tab:opt-list-heavyball}.

\begin{table}[!htp]
	\centering
	\scriptsize
	\setlength\tabcolsep{3pt} \begin{tabular}{@{}l@{\hskip-17pt}S[table-format=4]S[table-format=4]S[table-format=5]S[table-format=5]S[table-format=5]S[table-format=5]S[table-format=5]S[table-format=5]@{}}
	\toprule
	\textbf{Algorithm}  &\textbf{\criteo} &\textbf{\fastmri} &\multicolumn{2}{c}{\textbf{\imagenet}} &\multicolumn{2}{c}{\textbf{\librispeech}} &\textbf{\ogbg} &\textbf{\wmt} \\
	\cmidrule(lr){4-5} \cmidrule(lr){6-7}
	 &\textbf{\dlrmsmall} &\textbf{\unet} &\textbf{\resnetfifty} &\textbf{\vit} &\textbf{\conformer} &\textbf{\deepspeech} &\textbf{\gnn} &\textbf{\transformer} \\
	\midrule
	\textbf{\adamw} \\
		\quad \textsc{tuned} $\beta_1$ &5622 &{inf} &{inf} &62667 &95222 &80106 &{inf} &40534 \\
		\quad \textsc{fixed} $\beta_1$ &\B 5320 &7473 &{inf} &62667 &{inf} &81946 &{inf} &41499 \\
		\quad \textsc{opt-list} &5471 &\B 6415 &{inf} &64213 &88135 &\B 76427 &{inf} &44391 \\
	\addlinespace \textbf{\heavyball} \\
		\quad \textsc{tuned} $\beta_1$ &{inf} &{inf} &{inf} &{inf} &{inf} &{inf} &{inf} &{inf} \\
		\quad \textsc{fixed} $\beta_1$ &{inf} &{inf} &{inf} &{inf} &{inf} &{inf} &{inf} &{inf} \\
		\quad \textsc{opt-list} &{inf} &{inf} &\B  57321 &{inf} &{inf} &{inf} &{inf} &43984 \\
	\addlinespace \textbf{\lamb} \\
		\quad \textsc{tuned} $\beta_1$ &{inf} &{inf} &{inf} &{inf} &{inf} &78966 &{inf} &\B  29962 \\
	\addlinespace \textbf{\nadamw} \\
		\quad \textsc{tuned} $\beta_1$ &5850 &8559 &{inf} &62005 &92558 &79569 &\B  11441 &30822 \\
		\quad \textsc{fixed} $\beta_1$ &5544 &{inf} &61049 &60457 &{inf} &79569 &{inf} &43329 \\
		\quad \textsc{opt-list} &5544 &8205 &{inf} &\B 59682 &\B 87475 &77721 &12914 &44291 \\
	\addlinespace \textbf{\nesterov} \\
		\quad \textsc{tuned} $\beta_1$ &{inf} &{inf} &{inf} &{inf} &{inf} &{inf} &{inf} &{inf} \\
		\quad \textsc{fixed} $\beta_1$ &{inf} &{inf} &{inf} &{inf} &{inf} &{inf} &{inf} &{inf} \\
		\quad \textsc{opt-list} &{inf} &8750 &59330 &{inf} &{inf} &{inf} &{inf} &{inf} \\
	\addlinespace \textbf{\adafactor} \\
		\quad \textsc{tuned} $\beta_1$ &{inf} &{inf} &{inf} &{inf} &{inf} &82214 &{inf} &37679 \\
	\addlinespace \textbf{\samadam} \\
		\quad \textsc{tuned} $\beta_1$ &5910 &{inf} & {inf} &{inf} &{inf} &{inf} &{inf} &{inf} \\
	\bottomrule
\end{tabular}
 	\caption{\textbf{Total \runtime to achieve the target performance for different baselines.} These numbers are used to plot the performance profiles in \Cref{fig:perf_profiles_core_baseline_time} and \Cref{fig:perf_profiles_search_space_baseline_time}. A value of inf indicates that that baseline was unable to achieve the target within the maximum allowed \runtime.}
	\label{tab:baseline-performance-time}
\end{table}

\begin{table}[!htp]
	\centering
	\scriptsize
	\setlength\tabcolsep{3pt} \begin{tabular}{@{}l@{\hskip-43pt}S[table-format=4]S[table-format=5]S[table-format=6]S[table-format=6]S[table-format=6]S[table-format=5]S[table-format=5]S[table-format=5]@{}}
	\toprule
	\textbf{Algorithm}  &\textbf{\criteo} &\textbf{\fastmri} &\multicolumn{2}{c}{\textbf{\imagenet}} &\multicolumn{2}{c}{\textbf{\librispeech}} &\textbf{\ogbg} &\textbf{\wmt} \\
	\cmidrule(lr){4-5} \cmidrule(lr){6-7}
	&\textbf{\dlrmsmall} &\textbf{\unet} &\textbf{\resnetfifty} &\textbf{\vit} &\textbf{\conformer} &\textbf{\deepspeech} &\textbf{\gnn} &\textbf{\transformer} \\
	\midrule
	\textbf{\adamw} \\
		\quad \textsc{tuned} $\beta_1$ &7881 &{inf} &{inf} &151146 &250604 &69600 &{inf} &111972 \\
		\quad \textsc{fixed} $\beta_1$ &\B 7455 &30324 &{inf} &151146 &{inf} &71200 &{inf} &114638 \\
		\quad \textsc{opt-list} &7668 &\B 25992 &{inf} &154878 &231942 &\B 66400 &{inf} &122636 \\
	\addlinespace \textbf{\distshampoo} \\
		\quad \textsc{tuned} $\beta_1$ &{inf} &32851 &181002 &\B 138084 &\B 229276 &67200 &\B 35200 &90644 \\
	\addlinespace \textbf{\heavyball} \\
		\quad \textsc{tuned} $\beta_1$ &{inf} &{inf} &{inf} &{inf} &{inf} &{inf} &{inf} &{inf} \\
		\quad \textsc{fixed} $\beta_1$ &{inf} &{inf} &{inf} &{inf} &{inf} &{inf} &{inf} &{inf} \\
		\quad \textsc{opt-list} &{inf} &{inf} &169806 &{inf} &{inf} &{inf} &{inf} &122636 \\
	\addlinespace \textbf{\lamb} \\
		\quad \textsc{tuned} $\beta_1$ &{inf} &{inf} &{inf} &{inf} &{inf} &68800 &{inf} &\B 79980 \\
	\addlinespace \textbf{\nadamw} \\
		\quad \textsc{tuned} $\beta_1$ &8094 &34656 &{inf} &149280 &242606 &68800 &49600 &85312 \\
		\quad \textsc{fixed} $\beta_1$ &7668 &{inf} &181002 &145548 &{inf} &68800 &{inf} &119970 \\
		\quad \textsc{opt-list} &7668 &33212 &{inf} &143682 &\B 229276 &67200 &56000 &122636 \\
	\addlinespace \textbf{\nesterov} \\
		\quad \textsc{tuned} $\beta_1$ &{inf} &{inf} &{inf} &{inf} &{inf} &{inf} &{inf} &{inf} \\
		\quad \textsc{fixed} $\beta_1$ &{inf} &{inf} &{inf} &{inf} &{inf} &{inf} &{inf} &{inf} \\
		\quad \textsc{opt-list} &{inf} &35739 &175404 &{inf} &{inf} &{inf} &{inf} &{inf} \\
	\addlinespace \textbf{\adafactor} \\
		\quad \textsc{tuned} $\beta_1$ &{inf} &{inf} &{inf} &158610 &261268 &70400 &{inf} &98642 \\
	\addlinespace \textbf{\samadam} \\
		\quad \textsc{tuned} $\beta_1$ &8307 &{inf} &\B 166118 &147414 &231942 &67200 &{inf} &98642 \\
	\bottomrule
\end{tabular}
 	\caption{\textbf{Number of steps to achieve the target performance for different baselines.} These numbers are used to plot the performance profiles in \Cref{fig:perf_profiles_core_baseline_steps} and \Cref{fig:perf_profiles_search_space_baseline_steps}. A value of inf indicates that that baseline was unable to achieve the target in the maximum allowed step budget.}
	\label{tab:baseline-performance-steps}
\end{table}

\begin{table}[!htp]
	\centering
	\begin{tabular}{llS[table-format=2.6]S[table-format=2.6]}
	\toprule 
	\textbf{Submission} & \textbf{Version} & \multicolumn{2}{c}{\textbf{Benchmark Score}}\\
	\cmidrule(lr){3-4}
	& & {Runtime} & {Steps} \\
	\midrule
	\adamw &\textsc{tuned} $\beta_1$ & 0.600141 &0.596116 \\
	&\textsc{fixed} $\beta_1$ & 0.596985 &0.593047 \\
	&\textsc{opt-list} & 0.725260 &0.721035 \\
	\distshampoo &\textsc{tuned} $\beta_1$ & {-} &\B 0.85421 \\
	\heavyball &\textsc{tuned} $\beta_1$ & 0 &0 \\
	&\textsc{fixed} $\beta_1$ & 0 &0 \\
	&\textsc{opt-list} & 0.230504 &0.22686 \\
	\lamb &\textsc{tuned} $\beta_1$ & 0.248618 &0.248494 \\
	\nadamw &\textsc{tuned} $\beta_1$ & \B 0.849960 &0.595478 \\
	&\textsc{fixed} $\beta_1$ & 0.599691 &0.830414 \\
	&\textsc{opt-list} & 0.835602 &0.813194 \\
	\nesterov &\textsc{tuned} $\beta_1$ & 0 &0 \\
	&\textsc{fixed} $\beta_1$ & 0 &0 \\
	&\textsc{opt-list} & 0.233373 &0.232048 \\
	\adafactor &\textsc{tuned} $\beta_1$ & 0.236111 &0.47576 \\
	\samadam &\textsc{tuned} $\beta_1$ & 0.120368 &0.731717 \\
	\bottomrule
\end{tabular}
 	\caption{\textbf{Benchmark scores for all baselines presented in the paper.}}
	\label{tab:baselines-benchmark-scores}
\end{table}

\begin{table}[!htp]
	\centering
	\scriptsize
	\setlength\tabcolsep{2pt} \begin{tabular}{@{}l@{\hskip-10pt}S[table-format=5]S[table-format=5]S[table-format=6]S[table-format=6]S[table-format=6]S[table-format=6]S[table-format=5]S[table-format=6]@{}}
	\toprule
	\textbf{Algorithm}  &\textbf{\criteo} &\textbf{\fastmri} &\multicolumn{2}{c}{\textbf{\imagenet}} &\multicolumn{2}{c}{\textbf{\librispeech}} &\textbf{\ogbg} &\textbf{\wmt} \\
	\cmidrule(lr){4-5} \cmidrule(lr){6-7}
	&\textbf{\dlrmsmall} &\textbf{\unet} &\textbf{\resnetfifty} &\textbf{\vit} &\textbf{\conformer} &\textbf{\deepspeech} &\textbf{\gnn} &\textbf{\transformer} \\
	\midrule
	Steps &10667 &36189 &186667 &186667 &133333 &80000 &80000 &133333 \\\midrule
	\adamw &7602 &8906 &62827 &77380 &101368 &92064 &18361 &48261 \\
	\heavyball &7042 &8962 &\B 63008 &76535 &105506 &91460 &18303 &47818 \\
	\lamb &7166 &8972 &67797 &85254 &101982 &91802 &19330 &49911 \\
	\nadamw &\B 7703 &8935 &62958 &\B 77520 &\B 101780 &\B 92509 &18438 &\B 48151 \\
	\nesterov &7097 &\B 8859 &63137 &78791 &105108 &90874 &\B 18477 &47855 \\
	\adafactor &7419 &9058 &67381 &93895 &122920 &93404 &20379 &50896 \\
	\samadam &7583 &9048 &117997 &146507 &196144 &180832 &20852 &95139 \\
	\bottomrule
\end{tabular}
 	\caption{\textbf{Runtime measurements (rounded to the nearest second) for different \tas used in baselines.} For each \wl and \ta we measure the runtime for the mentioned number of steps. The highlighted entry for each workload corresponds to the runtime for the target-setting algorithm for that workload and thus defines our maximum allowed runtime for each workload.}
	\label{tab:baselines-runtimes}
\end{table}

\begin{table}[!htp]
	\centering
	\scriptsize
	\begin{tabular}{S[table-format=1.6]S[table-format=1.6]S[table-format=1.6]S[table-format=1.6]S[table-format=2]S[table-format=1.1]S[table-format=1.1]S[table-format=1.1]}
	\toprule
	\textbf{Learning} & $\bm{\beta_1}$ &$\bm{\beta_2}$ &\textbf{Weight} &\textbf{Warmup} &\textbf{Dropout} &\textbf{Aux.~Dropout} &\textbf{Label} \\
	\textbf{Rate} &\textbf{} &\textbf{} &\textbf{Decay} &\textbf{} &\textbf{} &\textbf{} &\textbf{Smoothing} \\
	\midrule
	0.002995 &0.960644 &0.998968 &0.006842 &5 &0 &0 &0 \\
	0.003331 &0.948 &0.998793 &0.003578 &2 &0.1 &0.1 &0 \\
	0.007502 &0.86932 &0.989658  &0.000071 &10 &0 &0 &0 \\
	0.000644 &0.931783 &0.969068 &0.677756 &5 &0 &0 &0 \\
	0.002827 &0.885906 &0.820717 &0.185407 &2 &0 &0 &0 \\
	0.001308 &0.973133 &0.998123 &0.163753 &10 &0 &0 &0.1 \\
	0.018978 &0.966607 &0.996816 &0.015654 &5 &0.1 &0.1 &0.2 \\
	0.001949 &0.532796 &0.981278 &0.037642 &5 &0 &0 &0.1 \\
	0.000845 &0.889576 &0.99785 &0.081354 &5 &0 &0 &0.2 \\
	0.000987 &0.99139 &0.993211 &0.00835 &10 &0 &0 &0.2 \\
	0.002107 &0.823119 &0.877457 &0.275905 &2 &0.1 &0.1 &0 \\
	0.002234 &0.751323 &0.612946 &0.215092 &5 &0 &0 &0 \\
	0.003926 &0.813913 &0.987628 &0.028657 &2 &0 &0 &0 \\
	0.001949 &0.532796 &0.981278 &0.037642 &5 &0 &0 &0 \\
	0.004958 &0.863744 &0.629185 &0.114739 &2 &0 &0 &0 \\
	0.000584 &0.962501 &0.998687 &0.000148 &5 &0.1 &0.1 &0.1 \\
	0.000504 &0.923701 &0.99475  &0.000012 &5 &0.1 &0.1 &0.2 \\
	0.000388 &0.674831 &0.946874 &0.124668 &10 &0.1 &0.1 &0 \\
	0.001749 &0.932661 &0.995516 &0.081216 &2 &0.1 &0.1 &0 \\
	0.000275 &0.788345 &0.076547 &0.023545 &2 &0.1 &0.1 &0.2 \\
	\bottomrule
\end{tabular}
 	\caption{\textbf{\optlisttext for \adamw.} The learning rate schedule used is \wcd.}
	\label{tab:opt-list-adam}
\end{table}

\begin{table}[!htp]
	\centering
	\scriptsize
	\begin{tabular}{S[table-format=1.6]S[table-format=1.6]S[table-format=1.6]S[table-format=1.6]S[table-format=2]S[table-format=1.1]S[table-format=1.1]S[table-format=1.1]}
	\toprule
	\textbf{Learning} & $\bm{\beta_1}$ &$\bm{\beta_2}$ &\textbf{Weight} &\textbf{Warmup} &\textbf{Dropout} &\textbf{Aux.~Dropout} &\textbf{Label} \\
	\textbf{Rate} &\textbf{} &\textbf{} &\textbf{Decay} &\textbf{} &\textbf{} &\textbf{} &\textbf{Smoothing} \\
	\midrule
	0.003331 &0.948 &0.998793 &0.003578 &2 &0.1 &0.1 &0 \\
	0.001614 &0.959792 &0.998463 &0.000033 &5 &0 &0 &0 \\
	0.010937 &0.974179 &0.998111 &0.007607 &5 &0.1 &0.1 &0 \\
	0.005146 &0.994362 &0.994663 &0.246009 &10 &0 &0 &0 \\
	0.002827 &0.885906 &0.820717 &0.185407 &2 &0 &0 &0 \\
	0.001308 &0.973133 &0.998123 &0.163753 &10 &0 &0 &0.1 \\
	0.018978 &0.966607 &0.996816 &0.015654 &5 &0.1 &0.1 &0.2 \\
	0.000845 &0.889576 &0.99785 &0.081354 &5 &0 &0 &0.2 \\
	0.001949 &0.532796 &0.981278 &0.037642 &5 &0 &0 &0.1 \\
	0.000987 &0.99139 &0.993211 &0.00835 &10 &0 &0 &0.2 \\
	0.001308 &0.973133 &0.998123 &0.163753 &10 &0 &0 &0 \\
	0.000845 &0.889576 &0.99785 &0.081354 &5 &0.1 &0.1 &0 \\
	0.004958 &0.863744 &0.629185 &0.114739 &2 &0 &0 &0 \\
	0.003528 &0.819231 &0.495851 &0.043397 &10 &0 &0 &0 \\
	0.001308 &0.973133 &0.998123 &0.163753 &10 &0 &0 &0 \\
	0.003296 &0.996693 &0.998649 &0.003729 &10 &0.1 &0.1 &0.1 \\
	0.000584 &0.962501 &0.998687 &0.000148 &5 &0.1 &0.1 &0.1 \\
	0.000279 &0.991934 &0.997984 &0.000324 &10 &0.1 &0.1 &0.1 \\
	0.001749 &0.932661 &0.995516 &0.081216 &2 &0.1 &0.1 &0 \\
	0.001011 &0.712472 &0.966607 &0.000069 &5 &0.1 &0.1 &0.1 \\
	\bottomrule
\end{tabular}
 	\caption{\textbf{\optlisttext for \nadamw.} The learning rate schedule used is \wcd.}
	\label{tab:opt-list-nadam}
\end{table}

\begin{table}[!htp]
	\centering
	\scriptsize
	\setlength\tabcolsep{5pt} \begin{tabular}{S[table-format=1.6]S[table-format=1.6]rS[table-format=1]S[table-format=1.6]S[table-format=1.3]S[table-format=1.1]S[table-format=1.1]S[table-format=1.1]}
	\toprule
	\textbf{Learning} &$\bm{\beta_1}$ &\textbf{Weight} &\textbf{Warmup} &\textbf{Decay} &\textbf{End} &\textbf{Dropout} &\textbf{Aux.} &\textbf{Label} \\
	\textbf{Rate} &\textbf{} &\textbf{Decay} &\textbf{} &\textbf{Steps} &\textbf{Factor} &\textbf{} &\textbf{Dropout} &\textbf{Smoothing} \\
	\midrule
	0.333132 &0.948 &1.40e-7 &5 &0.942079 &0.01 &0.1 &0.1 &0 \\
	0.082037 &0.980735 &1.01e-6 &5 &0.891621 &0.01 &0.1 &0.1 &0 \\
	0.810523 &0.898228 &1.00e-7 &5 &0.842587 &0.01 &0.1 &0.1 &0 \\
	0.028609 &0.981543 &5.77e-4 &5 &0.984398 &0.01 &0 &0 &0 \\
	0.416058 &0.970426 &1.99e-5 &5 &0.936585 &0.01 &0 &0 &0 \\
	4.131896 &0.927476 &5.67e-6 &5 &0.900777 &0.001 &0 &0 &0.2 \\
	0.191165 &0.995978 &3.83e-6 &5 &0.871275 &0.01 &0.1 &0.1 &0.2 \\
	1.376742 &0.736477 &5.09e-6 &5 &0.977277 &0.01 &0 &0 &0.2 \\
	0.032559 &0.988578 &3.32e-6 &5 &0.876362 &0.001 &0 &0 &0.1 \\
	0.130821 &0.973133 &2.90e-7 &5 &0.816545 &0.001 &0 &0 &0.2 \\
	0.022941 &0.984057 &2.40e-7 &5 &0.924988 &0.01 &0.1 &0.1 &0 \\
	0.010036 &0.986308 &3.22e-5 &5 &0.994571 &0.01 &0 &0 &0 \\
	0.026287 &0.992389 &3.88e-4 &5 &0.945944 &0.01 &0.1 &0.1 &0 \\
	0.014244 &0.970264 &4.22e-4 &5 &0.940451 &0.01 &0 &0 &0 \\
	0.019827 &0.95789 &2.41e-4 &5 &0.80861 &0.001 &0.1 &0.1 &0 \\
	2.491773 &0.944937 &1.30e-7 &5 &0.861509 &0.001 &0.1 &0.1 &0 \\
	2.051309 &0.917965 &4.58e-6 &5 &0.82041 &0.001 &0.1 &0.1 &0.1 \\
	1.897755 &0.966607 &6.90e-7 &5 &0.987857 &0.01 &0.1 &0.1 &0.1 \\
	0.169804 &0.99636 &1.03e-6 &5 &0.998233 &0.001 &0.1 &0.1 &0.1 \\
	0.253647 &0.989819 &1.15e-6 &5 &0.932109 &0.01 &0 &0 &0.1 \\
	\bottomrule
\end{tabular}
 	\caption{\textbf{\optlisttext for \nesterov.} The learning rate schedule used is \wldc.}
	\label{tab:opt-list-nesterov}
\end{table}

\begin{table}[!htp]
	\centering
	\scriptsize
	\setlength\tabcolsep{5pt} \begin{tabular}{S[table-format=1.6]S[table-format=1.6]rS[table-format=1]S[table-format=1.6]S[table-format=1.3]S[table-format=1.1]S[table-format=1.1]S[table-format=1.1]}
	\toprule
	\textbf{Learning} &$\bm{\beta_1}$ &\textbf{Weight} &\textbf{Warmup} &\textbf{Decay} &\textbf{Decay} &\textbf{Dropout} &\textbf{Aux.} &\textbf{Label} \\
	\textbf{Rate} &\textbf{} &\textbf{Decay} &\textbf{} &\textbf{Steps} &\textbf{Factor} &\textbf{} &\textbf{Dropout} &\textbf{Smoothing} \\
	\midrule
	0.299534 &0.960644 &1.10e-7 &5 &0.839739 &0.01 &0.1 &0.1 &0 \\
	5.133865 &0.673928 &2.50e-7 &5 &0.98745 &0.01 &0 &0 &0 \\
	1.042065 &0.862619 &3.90e-7 &5 &0.920716 &0.01 &0.1 &0.1 &0 \\
	0.028609 &0.981543 &5.77e-4 &5 &0.984398 &0.01 &0 &0 &0 \\
	0.416058 &0.970426 &1.99e-5 &5 &0.936585 &0.01 &0 &0 &0 \\
	4.131896 &0.927476 &5.67e-6 &5 &0.900777 &0.001 &0 &0 &0.2 \\
	7.263293 &0.860749 &2.81e-6 &5 &0.889586 &0.01 &0.1 &0.1 &0.2 \\
	1.376742 &0.736477 &5.09e-6 &5 &0.977277 &0.01 &0 &0 &0.2 \\
	1.111547 &0.589943 &6.31e-6 &5 &0.911763 &0.01 &0 &0 &0.1 \\
	0.392622 &0.813913 &6.62e-6 &5 &0.854795 &0.01 &0 &0 &0.2 \\
	0.130821 &0.973133 &2.90e-7 &5 &0.816545 &0.001 &0.1 &0.1 &0 \\
	0.022941 &0.984057 &2.40e-7 &5 &0.924988 &0.01 &0.1 &0.1 &0 \\
	0.142788 &0.961398 &1.83e-6 &5 &0.84625 &0.01 &0 &0 &0 \\
	0.027867 &0.991934 &3.20e-7 &5 &0.923564 &0.001 &0 &0 &0 \\
	0.026287 &0.992389 &3.88e-4 &5 &0.945944 &0.01 &0.1 &0.1 &0 \\
	2.051309 &0.917965 &4.58e-6 &5 &0.82041 &0.001 &0.1 &0.1 &0.1 \\
	1.897755 &0.966607 &6.90e-7 &5 &0.987857 &0.01 &0.1 &0.1 &0.1 \\
	2.491773 &0.944937 &1.30e-7 &5 &0.861509 &0.001 &0.1 &0.1 &0 \\
	0.426111 &0.995127 &9.80e-7 &5 &0.934347 &0.001 &0 &0 &0.2 \\
	0.169804 &0.99636 &1.03e-6 &5 &0.998233 &0.001 &0.1 &0.1 &0.1 \\
	\bottomrule
\end{tabular}
 	\caption{\textbf{\optlisttext for \heavyball.} The learning rate schedule used is \wldc.}
	\label{tab:opt-list-heavyball}
\end{table}

\section{Workload Details}
\label{app:workload_details}

\subsection{\criteo}
\label{app:wl_details_criteo}

We train on the \criteo Click Logs \dataset \citep{Criteo2014} to train a standard ads recommender model, \dlrm \citep{Naumov2019} to predict the CTR. The \dataset contains examples with $13$ numerical features and $26$ categorical.
The numerical features are log-transformed and the $26$ categorical features are hashed into a single embedding table. The data is from 24 days of click data, split into one file per day. We use the first 23 days as the training split, resulting in $4,195,197,692$ training examples. We then use the first half of the 24th day as the test set, and the second half of the 24th day as a validation set, resulting in $89,137,319$ validation and $89,137,318$ test examples. See \href{https://github.com/mlcommons/algorithmic-efficiency/blob/main/algorithmic_efficiency/workloads/criteo1tb/input_pipeline.py}{here} for the implementation of our \criteo input pipeline. Unlike many other Criteo 1TB pipelines, instead of the AUC we use the sigmoid binary cross entropy loss as an evaluation metric. We do this to have a metric that decomposes elementwise, which avoids requiring submitters to run expensive AUC evaluations that would have required maintaining arrays the size of the roughly 89 million evaluation examples.

We split the training split into files with $5,000,000$ lines each and randomly shuffle them. However, the training time budget we set typically only allows for around 60\% of an epoch to be consumed at the batch sizes we can fit into memory. In our \href{\initcodebase}{paper experiment codebase}, we reshuffle the dataset each time we have a preemption, which could result in repeated examples, but in the \href{\mlccodebase}{benchmark codebase} we do not assume that  the code runs on preemptable instances so this is not a concern.

\subsubsection{\dlrmsmall Model}
\label{app:wl_details_criteo_dlrm}

The concatenated features form the input layer for the \dlrm model \citep{Naumov2019}.
The single embedding table is of size $4$M entries with an embedding dimension of $128$.
The dense features are fed into a three-layer fully-connected network with $512, 256, 128$ units per layer. The outputs of this layer are then concatenated to the embedding lookups of the categorical features, and fed into the cross-interaction layer. Finally, the cross-interaction output is passed into a five-layer fully-connected network with $1024, 1024, 512, 256, 1$ units per layer. All layers use a \relu activation, except for the final $1$d output. A dropout layer follows the $512$ dimensional layer in the second network, after the \relu activation. See \href{https://github.com/mlcommons/algorithmic-efficiency/blob/main/algorithmic_efficiency/workloads/criteo1tb/criteo1tb_jax/models.py}{here} for our \dlrm model in Jax and \href{https://github.com/mlcommons/algorithmic-efficiency/blob/main/algorithmic_efficiency/workloads/criteo1tb/criteo1tb_pytorch/models.py}{here} for our \dlrm model in PyTorch.

\subsubsection{\criteo \dlrmsmall Workload Variants}
\label{app:wl_details_criteo_variants}
The three workload variants for the \dlrmsmall model are:
\begin{itemize}
    \item \textbf{Embed Init Scale}: We changed the initialization of the embedding layer from $\mathcal{N}\left(0, \frac{1}{\sqrt{4194304}}\right) \approx \mathcal{N}\left(0, 0.00049\right)$ (where $4194304$ is the vocabulary size), to $\mathcal{N}\left(0, 1\right)$.
    \item \textbf{LayerNorm}: \Layernorm was added after the activations of each layer, except for the final $1$d output.
    \item \textbf{Residual}: For every layer after the first layer of each subnetwork, instead of being a simple fully-connected layer, it is the residual branch in a residual subnetwork.
\end{itemize}

\begin{table}[!htp]
	\centering
	\begin{tabular}{lS[table-format=1.6]S[table-format=1.6]S[table-format=2.0]S[table-format=1.6]S[table-format=3.0]}
\toprule
\textbf{Variant} & \textbf{Validation}  & \multicolumn{4}{c}{\textbf{\Hp Performance Transfer}}                                                              \\
& \textbf{Performance} & \multicolumn{2}{c}{\textsc{Base} $\rightarrow$ \textsc{Variant}} & \multicolumn{2}{c}{\textsc{Variant} $\rightarrow$ \textsc{Base}} \\ \cmidrule(l){3-6} 
&          & {Performance} & {Rank} & {Performance} & {Rank} \\ \midrule
\textsc{Base Workload}    & \B 0.123609 &                  &      &                  &      \\
\addlinespace
\textsc{Embed Init Scale} & 0.123800   & 0.123880          & 3    & 0.123920          & 16   \\
\textsc{LayerNorm}        & 0.123653 & 0.123797         & 14   & 0.127778         & 136  \\
\textsc{Residual}         & 0.123920  & 0.124010          & 5    & 0.123860          & 13   \\
\bottomrule
\end{tabular}
 	\caption{\textbf{\Hp transfer between the base \wl and the variants of \dlrmsmall on \criteo.} We show the validation performance of the base \wl compared to the performance achieved by the variants. Further, we show the performance (and \hp ranking) when using the optimal \hp point from the base \wl on the variants (\textsc{Base} $\rightarrow$ \textsc{Variant}) and vice-versa (\textsc{Variant} $\rightarrow$ \textsc{Base}). All runs are from the same search space for \nadamw. See \Cref{sec:exp_randomized_workloads_process} for a detailed description of our protocol for accepting \wl variants.}
	\label{tab:criteo-variant-comparison}
\end{table}

\subsection{\fastmri}
\label{app:wl_details_fastmri}

The \fastmri \citep{Zbontar2018} \dataset was released by NYU Langone Health as part of a challenge organized in collaboration with Facebook AI Research (FAIR, now Meta AI).
The challenge aims to reduce the time to acquire an MRI scan by up to a factor of ten without any loss in diagnostic quality.
In MRI acquisition, a ``pulse sequence'' of spatially- and temporally-varying magnetic fields induces the subject's body to emit electromagnetic response fields that are measured by one or more receiver coils placed near the area to be imaged.
Using these fine-grained Fourier-space measurements (\textit{k-space} in the medical literature), clinicians reconstruct volumetric images with high-quality soft tissue contrast.
However, while powerful, these images can take $30$ minutes or more to produce, causing logistical issues ranging from patient discomfort to low patient throughput.
Recent strides in machine learning seek to reduce this time by reconstructing the volumetric images from sub-sampled \textit{k-space} measurements.
The \fastmri task provides raw \textit{k-space} data, randomly masks this data to simulate sub-sampling, and asks a supplied algorithm to faithfully reconstruct the image as compared to the inverse Fourier transform of the complete (non-sub-sampled) \textit{k-space} data.

In this benchmark, we use \fastmri's single-coil knee data which is organized into volumes, with each volume coming from one patient and being composed of a collection of slices (i.e., images).
In this task, we treat each slice as an independent example.
The official data set contains $34,742$ training slices (from $973$ volumes), and $7,135$ validation slices (from $199$ volumes).
In the data pipeline, we use a fixed random sequence to mask certain columns (down-sampling by a factor of four) from each raw \textit{k-space} slice before applying an inverse Fourier transform and normalizing the resulting image.
The target ground-truth image is normalized using the same mean and standard deviation.
Since the official test set targets are not publicly available, we split the validation set roughly in half to obtain disjoint validation and test sets---the first $100$ validation HDF5 files are used for validation and contain $3,554$ slices, while the final $99$ validation HDF5 are used for test and contain $3,581$ slices. See \href{https://github.com/mlcommons/algorithmic-efficiency/blob/main/algorithmic_efficiency/workloads/fastmri/input_pipeline.py}{here} for the implementation of our \fastmri input pipeline.

\subsubsection{\unet Model}
\label{app:wl_details_fastmri_unet}

We train a \unet model similar to the one described in \citet{Ronneberger2015}.
Our \unet implementation has $32$ channels, four down-sampling convolutional blocks, and four up-sampling transpose convolutional/convolutional block pairs.
Each layer uses dropout with a rate that may be tuned by the submission; the default dropout rate is 0.0. See \href{https://github.com/mlcommons/algorithmic-efficiency/blob/main/algorithmic_efficiency/workloads/fastmri/fastmri_jax/models.py}{here} for the implementation of our \unet model in Jax and \href{https://github.com/mlcommons/algorithmic-efficiency/blob/main/algorithmic_efficiency/workloads/fastmri/fastmri_pytorch/models.py}{here} for the implementation of our \unet model in PyTorch.

\subsubsection{\fastmri \unet Workload Variants}
\label{app:wl_details_fastmri_variants}
The three workload variants for the \unet model are:
\begin{itemize}
    \item \textbf{Channels \& Pooling}: The base number of channels in each convolution block was increased to $64$ (which is multiplied by another factor of $2$ with each down or up sample level), and the number of down and up sample layers was decreased to $3$.
    \item \textbf{\Tanh}: Activation functions were swapped to \Tanh.
    \item \textbf{LayerNorm}: Instance normalization was swapped to \layernorm (which has learnable parameters).
\end{itemize}
\begin{table}[!htp]
	\centering
	\begin{tabular}{lS[table-format=1.6]S[table-format=1.6]S[table-format=2.0]S[table-format=1.6]S[table-format=1.0]}
\toprule
\textbf{Variant} & \textbf{Validation}  & \multicolumn{4}{c}{\textbf{\Hp Performance Transfer}}                                                              \\
& \textbf{Performance} & \multicolumn{2}{c}{\textsc{Base} $\rightarrow$ \textsc{Variant}} & \multicolumn{2}{c}{\textsc{Variant} $\rightarrow$ \textsc{Base}} \\ \cmidrule(l){3-6} 
&          & {Performance} & {Rank} & {Performance} & {Rank} \\ \midrule
\textsc{Base Workload}    & 0.734523 &                  &      &                  &      \\
\addlinespace
\textsc{Channels \& Pooling} &0.734603 &0.734172 &12 &0.734277 &8 \\
\textsc{\Tanh}         &0.729743 &0.728203 &6 &0.734438 &3 \\
\textsc{LayerNorm}        & \B 0.734968 &0.733304 &26 &0.734364 &6 \\
\bottomrule
\end{tabular}
 	\caption{\textbf{\Hp transfer between the base \wl and the variants of \unet on \fastmri.} We show the validation performance of the base \wl compared to the performance achieved by the variants. Further, we show the performance (and \hp ranking) when using the optimal \hp point from the base \wl on the variants (\textsc{Base} $\rightarrow$ \textsc{Variant}) and vice-versa (\textsc{Variant} $\rightarrow$ \textsc{Base}). All runs are from the same search space for \nadamw. See \Cref{sec:exp_randomized_workloads_process} for a detailed description of our protocol for accepting \wl variants.}
	\label{tab:fastmri-variant-comparison}
\end{table}

\subsection{\imagenet}
\label{app:wl_details_imagenet}

For our \imagenet \wl we used the ILSVRC 2012 training and validation sets as the training and validation splits \citep{Deng2009}, and \imagenet-v2 as the test split \citep{recht2019imagenet}. For training preprocessing we take a random crop and randomly flip the image, whereas for the validation and test sets we just take a center crop (colloquially referred to as "ResNet preprocessing"). Images are fed into the model normalized to $\left[0, 1\right]$. We used RandAugment \citep{Cubuk2019} and mixup \citep{Zhang2017} for our \vit \wl but not for \resnet. Code for our \imagenet input pipelines can be found \href{https://github.com/mlcommons/algorithmic-efficiency/blob/main/algorithmic_efficiency/workloads/imagenet_resnet/imagenet_jax/input_pipeline.py}{here} in tf.data, \href{https://github.com/mlcommons/algorithmic-efficiency/blob/main/algorithmic_efficiency/workloads/imagenet_resnet/imagenet_pytorch/workload.py#L54}{here} in PyTorch, and \href{https://github.com/mlcommons/algorithmic-efficiency/blob/main/algorithmic_efficiency/workloads/imagenet_resnet/imagenet_v2.py}{here} for \imagenet-v2.

\subsubsection{\resnetfifty Model}
\label{app:wl_details_imagenet_resnet}

With the exception of \Cref{sec:chal_model_changes}, all experiments use the \resnetfifty defined in \citet[Section 4.1]{He2016}.
We use ghost \batchnorm with a virtual batch size of $64$ \citep{Hoffer2017}.
To further improve optimization stability, the scales in the final \batchnorm layer in each residual block are initialized to all zeros.
This has the effect of initializing each residual block to be the identity function. See \href{https://github.com/mlcommons/algorithmic-efficiency/blob/main/algorithmic_efficiency/workloads/imagenet_resnet/imagenet_jax/models.py}{here} for our implementation of \resnetfifty in \jax and  \href{https://github.com/mlcommons/algorithmic-efficiency/blob/main/algorithmic_efficiency/workloads/imagenet_resnet/imagenet_pytorch/models.py}{here} for our implementation of \resnetfifty in \pytorch.

In \Cref{sec:chal_model_changes} we ran experiments on an unstable $200$-layer \resnetvtwo architecture \citep{He2016a}.
These experiments also used a virtual batch size of $64$.
In contrast to the above \resnetfifty, we initialize all \batchnorm layers in the default way (with scalar factor all initialized to $1$).
This was necessary to produce the training instability at initialization.
The extra \batchnorm layer was added right after every residual connection, so each residual block returns $BN(x + F(x))$ instead of the standard $x + F(x)$.

\subsubsection{\imagenet \resnetfifty Workload Variants}
\label{app:wl_details_imagenet_resnet_variants}
The three workload variants for the \unet model are:
\begin{itemize}
    \item \textbf{SiLU}: Changed activation functions to \silu \citep{elfwing2018sigmoid}.
    \item \textbf{GELU}: Changed activation functions to \gelu \citep{hendrycks2016gaussian}.
    \item \textbf{BN Init Scale}: Changed the initialization of the scale variable in the final \batchnorm layer in the residual branches from $0.0$ to $8.0$.
\end{itemize}

In \Cref{tab:resnet50-variant-comparison} we present the results of our protocol to test our variants (described in \Cref{sec:exp_randomized_workloads_process}). We observe that the ranks of hyperparameter transfers were not as high as variants in other models. In such cases we tested these models along an alternative protocol to ensure that the optimum learning rate is indeed different for these models. 

\begin{table}[!htp]
	\centering
	\begin{tabular}{lS[table-format=1.5]S[table-format=1.5]S[table-format=1.0]S[table-format=1.5]S[table-format=1.0]}
\toprule
\textbf{Variant} & \textbf{Validation}  & \multicolumn{4}{c}{\textbf{\Hp Performance Transfer}}                                                              \\
& \textbf{Performance} & \multicolumn{2}{c}{\textsc{Base} $\rightarrow$ \textsc{Variant}} & \multicolumn{2}{c}{\textsc{Variant} $\rightarrow$ \textsc{Base}} \\ \cmidrule(l){3-6} 
&          & {Performance} & {Rank} & {Performance} & {Rank} \\ \midrule
\textsc{Base Workload}    & 0.22702 &                  &      &                  &      \\
\addlinespace
\textsc{SiLU} & \B 0.21996 &0.24234 &9 &0.22980 &2 \\
\textsc{GELU} & 0.22148 &0.22858 &3 &0.22980 &2 \\
\textsc{BN Init Scale} &0.23502 &0.23502 &1 &0.22702 &1 \\
\bottomrule
\end{tabular}
 	\caption{\textbf{\Hp transfer between the base \wl and the variants of \resnetfifty on \imagenet.} We show the validation performance of the base \wl compared to the performance achieved by the variants. Further, we show the performance (and \hp ranking) when using the optimal \hp point from the base \wl on the variants (\textsc{Base} $\rightarrow$ \textsc{Variant}) and vice-versa (\textsc{Variant} $\rightarrow$ \textsc{Base}). All runs are from the same search space for \nadamw. See \Cref{sec:exp_randomized_workloads_process} for a detailed description of our protocol for accepting \wl variants.}
	\label{tab:resnet50-variant-comparison}
\end{table}

\subsubsection{\vitfull Model}
\label{app:wl_details_imagenet_vit}

For all experiments, we use the \texttt{S/16} variant of the \vitfull (\vit) \citep{Dosovitskiy2021} as enumerated in \citet{Steiner2021} with no regularization and \texttt{light2} data augmentation.
We chose the \texttt{S/16} variant given its relatively small size (width of $384$, depth of $12$, MLP dimension of $1536$, with $6$ heads and a $16 \times 16$ image patch size).
\citet{Steiner2021} define \texttt{light2} data augmentation as using \texttt{mixup} \citep{Zhang2017} per batch, with $\alpha=0.2$ and \texttt{RandAugment} \citep{Cubuk2019} per image with two layers and magnitude $15$.
See \href{https://github.com/mlcommons/algorithmic-efficiency/blob/main/algorithmic_efficiency/workloads/imagenet_vit/imagenet_jax/models.py}{here} for our implementation of \vitfull in Jax and  \href{https://github.com/mlcommons/algorithmic-efficiency/blob/main/algorithmic_efficiency/workloads/imagenet_vit/imagenet_pytorch/models.py}{here} for our implementation of \vitfull in PyTorch.

\subsubsection{\imagenet \vitfull Workload Variants}
\label{app:wl_details_imagenet_vit_variants}
The three workload variants for the \unet model are:
\begin{itemize}
    \item \textbf{\postln}: \Layernorm was applied after the residual branch was added back into the trunk. 
    \item \textbf{MAP}: The pooling layer type was changed from global average pooling to multihead attention pooling.
    \item \textbf{GLU}: We included gated linear units in the \textsc{MLPBlock}.
\end{itemize}
In \Cref{tab:vit-variant-comparison} we present the results of our protocol to test our variants (described in \Cref{sec:exp_randomized_workloads_process}). We observe that the ranks of hyperparameter transfers were 0 for two variants. For these variants we tested these models along an alternative protocol to ensure that the optimum learning rate is indeed different for these models. 
\begin{table}[!htp]
	\centering
	\begin{tabular}{lS[table-format=1.5]S[table-format=1.5]S[table-format=2.0]S[table-format=1.5]S[table-format=3.0]}
\toprule
\textbf{Variant} & \textbf{Validation}  & \multicolumn{4}{c}{\textbf{\Hp Performance Transfer}}                                                              \\
& \textbf{Performance} & \multicolumn{2}{c}{\textsc{Base} $\rightarrow$ \textsc{Variant}} & \multicolumn{2}{c}{\textsc{Variant} $\rightarrow$ \textsc{Base}} \\ \cmidrule(l){3-6} 
&          & {Performance} & {Rank} & {Performance} & {Rank} \\ \midrule
\textsc{Base Workload}     &      0.22530            &      &                  &      \\
\addlinespace
\postln & 0.24794   & 0.24794          & 14    & 0.28478          & 5   \\
\textsc{MAP}        & 0.23004 & 0.23004         & 0   & 0.22530        & 0  \\
\textsc{GLU}   &      \B 0.22258   & 0.22258          & 0    & 0.22530         & 0   \\
\bottomrule
\end{tabular}
 \caption{\textbf{\Hp transfer between the base \wl and the variants of \vit on \imagenet.} We show the validation performance of the base \wl compared to the performance achieved by the variants. Further, we show the performance (and \hp ranking) when using the optimal \hp point from the base \wl on the variants (\textsc{Base} $\rightarrow$ \textsc{Variant}) and vice-versa (\textsc{Variant} $\rightarrow$ \textsc{Base}). All runs are from the same search space for \nadamw. See \Cref{sec:exp_randomized_workloads_process} for a detailed description of our protocol for accepting \wl variants.}
	\label{tab:vit-variant-comparison}
\end{table}

\subsection{\librispeech}
\label{app:wl_details_librispeech}

The \librispeech \dataset \citep{Panayotov2015} is a corpus of read English speech with sampling rate of $16$ kHz.
We train our speech recognition models on the combination of \texttt{train-clean-100}, \texttt{train-clean-360}, and \texttt{train-other-500} splits from the \librispeech \dataset giving us $960$ hours of raw audio data.
For validation we use a combination of \texttt{dev-clean} and \texttt{dev-other} splits resulting in $5567$ examples in the validation set.
We report the word error rates on the \texttt{test-clean} split as our test set performance.

We preprocess the data by eliminating any examples with audio length greater than $320$k and a target sentence length greater than $256$.
We then compute logmel spectrogram features for the raw audio input and use the \texttt{SentencePiece} \citep{kudo2018spm} tokenizer with a vocab size of $1024$ to tokenize the target sentences.
We pad the sequences to bring all examples to same length and handle paddings inside the model and while computing metrics like loss and word error rate.
See \href{https://github.com/mlcommons/algorithmic-efficiency/blob/main/algorithmic_efficiency/workloads/librispeech_conformer/input_pipeline.py}{here} for the implementation of our \librispeech input pipeline.

\subsubsection{\conformer Model}
\label{app:wl_details_librispeech_conformer}

\conformer \citep{Gulati2020} is an architecture combing attention and convolution layers to capture both global and local relationships in input audio.
We use a $4$-layer deep \conformer model with model encoder dimension of $512$.
Model weights are initialized using Xavier uniform initialization.
CTC loss \citep{Graves2006} is used to train the model.
See \href{https://github.com/mlcommons/algorithmic-efficiency/blob/main/algorithmic_efficiency/workloads/librispeech_conformer/librispeech_jax/models.py}{here} for our implementation of \conformer in Jax and  \href{https://github.com/mlcommons/algorithmic-efficiency/blob/main/algorithmic_efficiency/workloads/librispeech_conformer/librispeech_pytorch/models.py}{here} for our implementation of \conformer in PyTorch.

\paragraph{Inference} We use a greedy decoding procedure to generate decoded logits which are then transformed back into sentences using \texttt{SentencePiece} \citep{kudo2018spm} tokenizer to compute word error rates.

\subsubsection{\librispeech \conformer Workload Variants}
\label{app:wl_details_librispeech_conformer_variants}
The three workload variants for the \conformer model are:
\begin{itemize}
    \item \textbf{GELU}: Activations functions were changed to \gelu.
    \item \textbf{LayerNorm change} The \layernorm before the final readout layer was removed. 
    \item \textbf{Attention Temp} The \emph{output} of the attention softmax was multiplied by a temperature constant of $1.6$. Note that this is different than other attention temperature setups where the temperature is multiplied in before the softmax.
\end{itemize}

In \Cref{tab:conformer-variant-comparison} we present the results of our protocol to test our variants (described in \Cref{sec:exp_randomized_workloads_process}). We observe that the ranks of hyperparameter transfers were not as high as variants in other models. In such cases we tested these models along an alternative protocol to ensure that the optimum learning rate is indeed different for these models.

\begin{table}[!htp]
	\centering
	\begin{tabular}{lS[table-format=1.6]S[table-format=1.6]S[table-format=1.0]S[table-format=1.6]S[table-format=1.0]}
\toprule
\textbf{Variant} & \textbf{Validation}  & \multicolumn{4}{c}{\textbf{\Hp Performance Transfer}}                                                              \\
& \textbf{Performance} & \multicolumn{2}{c}{\textsc{Base} $\rightarrow$ \textsc{Variant}} & \multicolumn{2}{c}{\textsc{Variant} $\rightarrow$ \textsc{Base}} \\ \cmidrule(l){3-6} 
&          & {Performance} & {Rank} & {Performance} & {Rank} \\ \midrule
\textsc{Base Workload}    & \B 0.077790 &                  &      &                  &      \\
\addlinespace
\textsc{GELU} &0.079354 &0.079391 &2 &0.085503 &3 \\
\textsc{LayerNorm Change} &0.084175 &0.084175 &1 &0.229800 &1 \\
\textsc{Attention Temp} &0.083092 &0.083092 &1 &0.077790 &1 \\
\bottomrule
\end{tabular}
 	\caption{\textbf{\Hp transfer between the base \wl and the variants of \conformer on \librispeech.} We show the validation performance of the base \wl compared to the performance achieved by the variants. Further, we show the performance (and \hp ranking) when using the optimal \hp point from the base \wl on the variants (\textsc{Base} $\rightarrow$ \textsc{Variant}) and vice-versa (\textsc{Variant} $\rightarrow$ \textsc{Base}). All runs are from the same search space for \nadamw. See \Cref{sec:exp_randomized_workloads_process} for a detailed description of our protocol for accepting \wl variants.}
	\label{tab:conformer-variant-comparison}
\end{table}

\subsubsection{\deepspeech Model}
\label{app:wl_details_librispeech_deepspeech}

We use a variant of the \deepspeech \citep{Amodei2015} model with residual connections, \dropout \citep{dropout}, \layernorm, and \specaug \citep{Park2019} to improve performance.
We use a convolution subsampling layer to reduce input dimensions by a factor of $4$, which is further passed through $6$ bi-directional LSTM layers and $3$ feed-forward layers with a model internal dimension of $512$ across layers.
We use \batchnorm inside the LSTM and feed-forward layers as post normalization layers.
Model weights are initialized using Xavier uniform initialization.
The CTC loss \citep{Graves2006} is used to train the model.
See \href{https://github.com/mlcommons/algorithmic-efficiency/blob/main/algorithmic_efficiency/workloads/librispeech_deepspeech/librispeech_jax/models.py}{here} for our implementation of \deepspeech in Jax and  \href{https://github.com/mlcommons/algorithmic-efficiency/blob/main/algorithmic_efficiency/workloads/librispeech_deepspeech/librispeech_pytorch/models.py}{here} for our implementation of \deepspeech in PyTorch.
For experiments in this paper, we used a slightly different word piece tokenizer than \conformer, but we plan to update the tokenizer before the call for submissions to also use the same \texttt{SentencePiece} \citep{kudo2018spm} tokenizer. We expect the \deepspeech validation WER target to improve from 0.1162 to 0.111825. 

\subsubsection{\librispeech \deepspeech Workload Variants}
\label{app:wl_details_librispeech_deepspeech_variants}
The three workload variants for the \deepspeech model are:
\begin{itemize}
    \item \textbf{\Tanh}: Activation functions were changed to \Tanh.
    \item \textbf{No residual}: We removed the residual connections in the model. Interestingly this improved the overall performance of the model. 
    \item \textbf{Norm \& SpecAugment}: We removed the decoder \layernorm layer. We replaced all other \batchnorm layers with \layernorm layers. We changed \specaug specifications, specifically the \textsc{Frequency Mask} from $2$ to $4$ and the \textsc{Time Mask} from $10$ to $15$.
\end{itemize}

\begin{table}[!htp]
	\centering
	\begin{tabular}{lS[table-format=1.6]S[table-format=1.6]S[table-format=2.0]S[table-format=1.6]S[table-format=1.0]}
\toprule
\textbf{Variant} & \textbf{Validation}  & \multicolumn{4}{c}{\textbf{\Hp Performance Transfer}}                                                              \\
& \textbf{Performance} & \multicolumn{2}{c}{\textsc{Base} $\rightarrow$ \textsc{Variant}} & \multicolumn{2}{c}{\textsc{Variant} $\rightarrow$ \textsc{Base}} \\ \cmidrule(l){3-6} 
&          & {Performance} & {Rank} & {Performance} & {Rank} \\ \midrule
\textsc{Base Workload}    & 0.113950 &                  &      &                  &      \\
\addlinespace
\textsc{\Tanh} &0.131300 &0.177558 &32 &0.122959 &9 \\
\textsc{No Residual} & \B 0.105063 &0.135881 &23 &0.116508 &2 \\
\textsc{Norm \& SpecAugment} &0.130967 &0.141442 &6 &0.117388 &3 \\
\bottomrule
\end{tabular}
 	\caption{\textbf{\Hp transfer between the base \wl and the variants of \deepspeech on \librispeech.} We show the validation performance of the base \wl compared to the performance achieved by the variants. Further, we show the performance (and \hp ranking) when using the optimal \hp point from the base \wl on the variants (\textsc{Base} $\rightarrow$ \textsc{Variant}) and vice-versa (\textsc{Variant} $\rightarrow$ \textsc{Base}). All runs are from the same search space for \nadamw. See \Cref{sec:exp_randomized_workloads_process} for a detailed description of our protocol for accepting \wl variants.}
	\label{tab:deepspeech-variant-comparison}
\end{table}

\subsection{\ogbg}
\label{app:wl_details_ogbg}

We use the \ogbg-MOLPCBA \dataset \citep{Hu2020} containing molecular graphs and $128$ molecular properties.
The goal is to predict the properties given the graphs.
The graphs contain on average $26$ nodes and $28$ edges.
The train split contains $350,343$ examples, and the train and validation splits both contain $43,793$ examples.

Because the graphs are of varying sizes, we construct the batches using dynamic batching.
We specify a maximum number of nodes and edges per batch and take graphs from the \dataset into the batch until one of these thresholds is exceeded.
We set these thresholds to $\mathrm{BatchSize} \cdot \mathrm{AvgNumNodes}$ and $2 \cdot \mathrm{BatchSize} \cdot \mathrm{AvgNumEdges}$ (the factor of $2$ accounts for the bidirectional nature of the edges), respectively.
See \href{https://github.com/mlcommons/algorithmic-efficiency/blob/main/algorithmic_efficiency/workloads/ogbg/input_pipeline.py}{here} for the implementation of our \ogbg input pipeline.

\subsubsection{\gnn Model}
\label{app:wl_details_ogbg_gnn}

The model is defined as a graph neural network (\gnn) \citep{graphnetwork}, which is a generalization of graph architectures such as GIN \citep{Xu2019}.
We use the implementation provided by the \jraph library \citep{jraph2020github}.
The model first performs an embedding step which transforms the node and edge features with a linear embedding of size $256$, and creates global features of size $128$ initialized to zeros.
Then the model performs $5$ message passing steps following \citet[Algorithm 1]{graphnetwork} which update the node, edge, and global features.
Each of the update functions is a $1$-layer fully-connected network with a dense layer of size $256$, \layernorm, \relu, and \dropout.
The \dropout rate can be tuned by the submissions but it defaults to 0.1.
The weights for the model are initialized using LeCun normal initialization \citep{Klambauer2017}.
The final output is read from the global features after the last step.
See \href{https://github.com/mlcommons/algorithmic-efficiency/blob/main/algorithmic_efficiency/workloads/ogbg/ogbg_jax/models.py}{here} for our implementation of \gnn in Jax and  \href{https://github.com/mlcommons/algorithmic-efficiency/blob/main/algorithmic_efficiency/workloads/ogbg/ogbg_pytorch/models.py}{here} for our implementation of \gnn in PyTorch.

\subsubsection{\ogbg \gnn Workload Variants}
\label{app:wl_details_ogbg_variants}
The three workload variants for the \gnn model are:
\begin{itemize}
    \item \textbf{GELU}: The activation was changed to \gelu.
    \item \textbf{SiLU}: The activation was changed to \silu.
    \item \textbf{Altered Layers}: An additional hidden layer of width $256$ was added to each fully connected network, the latent dim was reduced to $128$, and the number of message passing steps was reduced to $3$. Also, the \layernorm layers were swapped for \batchnorm.
\end{itemize}

\begin{table}[!htp]
	\centering
	\begin{tabular}{lS[table-format=1.6]S[table-format=1.6]S[table-format=2.0]S[table-format=1.6]S[table-format=2.0]}
\toprule
\textbf{Variant} & \textbf{Validation}  & \multicolumn{4}{c}{\textbf{\Hp Performance Transfer}}                                                              \\
& \textbf{Performance} & \multicolumn{2}{c}{\textsc{Base} $\rightarrow$ \textsc{Variant}} & \multicolumn{2}{c}{\textsc{Variant} $\rightarrow$ \textsc{Base}} \\ \cmidrule(l){3-6} 
&          & {Performance} & {Rank} & {Performance} & {Rank} \\ \midrule
\textsc{Base Workload}    & 0.280012 &                  &      &                  &      \\
\addlinespace
\textsc{GELU} &0.284251 &0.261344 &25 &0.267038 &17 \\
\textsc{SiLU} & \B 0.287569 &0.258375 &28 &0.275818 &3 \\
\textsc{Altered Layers} & 0.269345 &0.252153 &15 &0.244109 &44 \\
\bottomrule
\end{tabular}
 	\caption{\textbf{\Hp transfer between the base \wl and the variants of \gnn on \ogbg.} We show the validation performance of the base \wl compared to the performance achieved by the variants. Further, we show the performance (and \hp ranking) when using the optimal \hp point from the base \wl on the variants (\textsc{Base} $\rightarrow$ \textsc{Variant}) and vice-versa (\textsc{Variant} $\rightarrow$ \textsc{Base}). All runs are from the same search space for \nadamw. See \Cref{sec:exp_randomized_workloads_process} for a detailed description of our protocol for accepting \wl variants.}
	\label{tab:gnn-variant-comparison}
\end{table}

\subsection{\wmt}
\label{app:wl_details_wmt}

The models are trained on the \wmt 2017 German$\rightarrow$English (De$\rightarrow$En) training \dataset \citep{Bojar2017} which consists of about $5.9$ million parallel sentence pairs.
We don't use any monolingual data or data augmentation methods.
The models are evaluated on the "validation" and "test" splits of the \wmt 2014 De$\rightarrow$En \dataset \citep{Bojar2014} which consists of $3000$ and $3003$ sentence pairs respectively.
We use \texttt{SentencePiece} \citep{kudo2018spm} with $32$k vocabulary size which is shared on source and target side.
Sentences longer than $256$ tokens on either the source or the target side are removed from the training data.
See \href{https://github.com/mlcommons/algorithmic-efficiency/blob/main/algorithmic_efficiency/workloads/wmt/input_pipeline.py}{here} for the implementation of our \wmt input pipeline.

\subsubsection{\transformer Model}
\label{app:wl_details_wmt_transformer}

We use the \transformer-big architecture from \citet{Vaswani2017} with some modifications.
More specifically, the model has $6$ encoder and decoder layers each, model dimension $1024$, hidden dimension $4096$ and $16$ attention heads.
The embedding parameters are shared on the encoder and the decoder side.
The same embedding matrix is also used for linear readout on the decoder (softmax).
For regularization, the model uses label smoothing and \dropout where the rates may be tuned by the submissions; the default label smoothing is $0.1$ and the default \dropout rate is $0.1$.
A batch size of $128$ corresponds to about $280$K tokens each on source and target side in one training batch.
See \href{https://github.com/mlcommons/algorithmic-efficiency/blob/main/algorithmic_efficiency/workloads/wmt/wmt_jax/models.py}{here} for our implementation of \transformer in Jax and  \href{https://github.com/mlcommons/algorithmic-efficiency/blob/main/algorithmic_efficiency/workloads/wmt/wmt_pytorch/models.py}{here} for our implementation of \transformer in PyTorch.

\paragraph{Decoding} To generate samples from the model, we use beam search decoding with beam size $4$, and a length penalty of $0.6$.
For evaluation, we use standard de-tokenized \bleu scores \citep{papineni-2002-bleu} using the \sacrebleu library \citep{post-2018-call}.\footnote{case.mixed + numrefs.1 + smooth.exp + tok.13a + sacreBLEU 1.3.1}

\subsubsection{\wmt \transformer Workload Variants}
\label{app:wl_details_wmt_variants}
The three workload variants for the \wmt model are:
\begin{itemize}
    \item \textbf{\postln}: \Layernorm was applied after the residual branch was added back into the trunk.
    \item \textbf{Attention Temp}: The attention logits were multiplied by a temperature of $4.0$, before applying the softmax.
    \item \textbf{GLU \& \Tanh}: Gated Linear Units (GLU) \citep{dauphin2017language} were used in the MLP blocks, and activation functions were changed to \Tanh.
\end{itemize}

\begin{table}[!htp]
	\centering
	\begin{tabular}{lS[table-format=2.4]S[table-format=2.4]S[table-format=2.0]S[table-format=2.4]S[table-format=2.0]}
\toprule
\textbf{Variant} & \textbf{Validation}  & \multicolumn{4}{c}{\textbf{\Hp Performance Transfer}}                                                              \\
& \textbf{Performance} & \multicolumn{2}{c}{\textsc{Base} $\rightarrow$ \textsc{Variant}} & \multicolumn{2}{c}{\textsc{Variant} $\rightarrow$ \textsc{Base}} \\ \cmidrule(l){3-6} 
&          & {Performance} & {Rank} & {Performance} & {Rank} \\ \midrule
\textsc{Base Workload}    & \B 30.8534 &                  &      &                  &      \\
\addlinespace
\textsc{Post-LN} &30.2011 &11.1946 &76 &29.7474 &23 \\
\textsc{Attention Temp} &30.2643 &0 &74 &29.4278 &32 \\
\textsc{GLU \& \Tanh} &30.1595 &0 &82 &29.7135 &25 \\
\bottomrule
\end{tabular}
 	\caption{\textbf{\Hp transfer between the base \wl and the variants of \transformer on \wmt.} We show the validation performance of the base \wl compared to the performance achieved by the variants. Further, we show the performance (and \hp ranking) when using the optimal \hp point from the base \wl on the variants (\textsc{Base} $\rightarrow$ \textsc{Variant}) and vice-versa (\textsc{Variant} $\rightarrow$ \textsc{Base}). All runs are from the same search space for \nadamw. See \Cref{sec:exp_randomized_workloads_process} for a detailed description of our protocol for accepting \wl variants.}
	\label{tab:wmt-variant-comparison}
\end{table}

\section{Preliminary Experiments for Randomized Workloads}
\label{app:ogbg_randomized}

In order to create randomized \wls based on one of the fixed \wls, we needed to write down a distribution over \wl modifications that has the properties we want. Our initial plan was to design a relatively broad distribution over natural changes to the base \wl, then draw \heldout \wl samples, rejecting ones that failed a ``trainability'' test, and then rejecting ones that were too close to the base \wl. Then we would revise the distribution to hopefully reduce the rejection rates and produce a final, official randomized \wl. We started this process for the \gnn \wl on \ogbg before eventually abandoning it.

The distribution we started with is defined over the \gnn's hidden dimensions, latent dimensions (node and edge embedding dimension), normalization layer type, activation function, number of message passing steps, and dropout probability\footnote{Our initial experiments on randomized \wls for \ogbg included the dropout probability as part of the \wl definition. However, in the competition, submitters are free to determine the dropout probability. Therefore, a final, official randomized \wl distribution for our benchmark could not include the dropout probability.}, see \Cref{tab:heldout-distribution}.
All \wl samples from this distribution fit into competition hardware GPU memory with a batch size of $512$. The batch size is not part of the \wl definition, so in principle we might generate \heldout \wls that require adjusting the batch size to remain within memory constraints. In general, we also need to find the best-performing batch size for target setting in order to reach competitive targets.
\begin{table}[!htp]
	\centering
	\begin{tabular}{@{}lc@{}}
	\toprule
	\textbf{\gnn Parameters} & \textbf{Distribution} \\
	\midrule
	Hidden dimension 1 &  $\{128, 256, 512, 1024, 2048\}$ \\
	Hidden dimension 2 &  $\{\text{None}, 128, 256, 512, 1024, 2048\}$ \\
	Latent dimension & $\{128, 256, 512, 1024, 2048\}$  \\
	Normalization layer & \{\batchnorm, layer normalization\}\\
	Activation function & \{\relu, \silu, Leaky ReLU\} \\
	$N$ message passing steps & $\{2,3,4,5,6,7,8\}$ \\ 
	Dropout probability & $[0.0, 0.5]$ \\ 
	\bottomrule
\end{tabular}
 	\caption{\textbf{Distribution used in a preliminary experiment for a potential randomized \gnn \ogbg \wl.}}
	\label{tab:heldout-distribution}
\end{table}

After sampling $60$ \heldout \wl candidates from the distribution described by \Cref{tab:heldout-distribution}, we ran a target-setting procedure on each sample. In other words, for each sample we tuned \nesterov using $100$ \hp points to find the \hps that achieve the best mAP. We considered a sample able to achieve an acceptable result in the \runtime budget if, after tuning, it reached an mAP no more than $10\,\%$ worse than the original base \wl mAP target. We found that roughly $30\,\%$ of the samples were trainable by this definition.  

To quickly check whether the surviving $16$ trainable \wl candidates were sufficiently different from the base \wl, we ran the best-performing \hp settings from the base \wl on each candidate. Unfortunately, in pretty much every case except for one activation function change, the best-performing \hp settings from the base \wl achieved about the same (within re-training noise) mAP as the best we could find after tuning just on the \wl sample.

We decided to abandon this process for constructing randomized \wl distributions (and not extend it to the other base \wls) for a variety of reasons. First, it was far too costly. Every time we revised the distribution we were considering, we had to run our entire target-setting procedure on every single sample from the distribution (in the experiment described above, this amounted to running $6000$ trials for just one version of the distribution). Our procedure above also ignored some of the technicalities around selecting batch sizes for target setting, and we had to be very careful to filter out samples (or write down distributions) that didn't violate the memory constraints of the competition hardware. We also did not have the understanding of what kinds of changes would cause the training problem to actually change that would be necessary to quickly come up with interesting distributions over \wls. Instead, we decided to adopt a procedure of manually constructing variants one at a time (see \Cref{sec:exp_randomized_workloads_process}).
 
\end{appendices}
 
\newpage
\bibliography{bibliography}

\end{document}